\documentclass[11pt]{article}

\usepackage[a4paper,margin=1in]{geometry}
\usepackage{newtxtext,newtxmath}

\usepackage{graphicx}
\usepackage{booktabs}
\usepackage{multirow}
\usepackage{caption}
\usepackage{subcaption}       
\usepackage{placeins}         
\usepackage{url}

\usepackage{algorithm}
\usepackage[noend]{algpseudocode}

\usepackage[title]{appendix}

\providecommand{\shorttitle}{Cross-Language Speaker Attribute Prediction}

\makeatletter
\renewcommand{\@oddhead}{\hfil \shorttitle \hfil}
\renewcommand{\@evenhead}{\hfil \shorttitle \hfil}
\makeatother

\usepackage[square,numbers,sort&compress]{natbib}
\usepackage[hidelinks]{hyperref} 

\begin{document}

\begin{center}
{\LARGE Cross-Language Speaker Attribute Prediction Using MIL and RL\par}

{\large
Sunny Shu$^{1}$,\,
Seyed Sahand Mohammadi Ziabari$^{1,2*}$,\,
Ali Mohammed Mansoor Alsahag$^{1}$
\par}
\vspace{6pt}

{\small
$^{1}$Informatics Institute, University of Amsterdam, 1098 XH Amsterdam, The Netherlands\\
$^{2}$Department of Computer Science and Technology, SUNY Empire State University, Saratoga Springs, NY 12866, USA
\par}
\vspace{6pt}

{\small
E-mail: \href{mailto:sunny.shu@student.uva.nl}{sunny.shu@student.uva.nl};
\href{mailto:a.m.m.alsahag@uva.nl}{a.m.m.alsahag@uva.nl};
\href{mailto:sahand.ziabari@sunyempire.edu}{sahand.ziabari@sunyempire.edu}
\par}

{\small
\par}
\end{center}

\vspace{12pt}

\noindent\textbf{Abstract}\quad
Speaker attribute prediction from text remains difficult in multilingual settings due to linguistic variation, domain mismatch, and data imbalance across languages. We present \emph{RLMIL-DAT}, a multilingual extension of the RL--MIL framework that couples reinforcement-learning-based instance selection with Domain Adversarial Training (DAT) via a gradient reversal layer to induce language-invariant utterance representations. We evaluate on a five-language Twitter corpus (few-shot) and a custom 40-language VoxCeleb2-derived corpus (zero-shot) for gender and age prediction. Across 27 model variants (3 encoders $\times$ 3 pooling heads $\times$ 3 training frameworks) and five random seeds, \emph{RLMIL-DAT} consistently improves Macro-F1 over both standard MIL and the original RL--MIL. Gains are largest and statistically significant on gender, with improvements up to +0.17 Macro-F1 ($p\le 0.01$) depending on encoder and pooling; age remains harder, yielding smaller but positive trends. Ablations show DAT is the primary driver of these gains, enabling effective transfer from high-resource English to lower-resource languages by discouraging language-specific cues in the shared encoder. In zero-shot evaluation on the smaller VoxCeleb2 subset, improvements are directionally positive but less often significant, reflecting limited statistical power and the challenge of generalizing to 40 languages without target-language supervision. Overall, results validate the extensibility of RL--MIL to multilingual scenarios and demonstrate that combining instance selection with adversarial domain adaptation is an effective and robust strategy for cross-lingual speaker attribute prediction.
\vspace{6pt}


\section{Introduction}

\label{sec:introduction}

In contemporary society, data science informs almost every aspect of daily life. As of 2024, the number of global social media users is over five billion, and projections indicate that this figure will exceed six billion by 2028~\cite{statista_social_network_users}. Social media platforms provide striking and valuable resource access to written language data, which can be leveraged for various analytical purposes~\cite{schwartz2013personality}. One area that has gained significant attention is speaker attribute prediction, a method that provides insights into demographic and psychological characteristics without the need for intrusive methods~\cite{ziabari2024reinforced}. Previous research examined 700 million linguistic instances, from words and phrases to topics, in Facebook messages collected from 75,000 volunteers~\cite{schwartz2013personality}. The findings revealed notable variations in personality, gender, and age as reflected in the language used on social media, emphasizing the potential power of speaker attribute prediction in informing our understanding of individual and group characteristics~\cite{schwartz2013personality}. 

This potential also holds significant commercial and societal implications. From a commercial perspective, speaker attribute prediction facilitates the application of user profiling for market analysis, enabling the creation of more effective and targeted advertising and content personalization~\cite{kabbur2010content}. Societal applications include utilizing an understanding of demographic language differences to develop better and fairer Natural Language Processing (NLP) tools, and creating more trustworthy information systems that support users of low-resource languages in accessing knowledge not sourced from their native language~\citep{johannsen2015cross, muller2023evaluating}.

Speaker attribute prediction commonly aims to determine characteristics such as gender, age, political ideology, and other personal traits from speech data~\cite{ziabari2024reinforced}. Gender generally refers to whether the speaker is identified as a male or female. The usage of language by the speakers could differ by gender, females typically produce more and longer conversational turns, while males have a higher type-token ratio and preferences of particles such as "eh", "uh" and "em"~\cite{liesenfeld2021predicting, argamon2007mining, ziabari2024reinforced}. Age is generally a period of time that a person has lived; it can also be classified into age groups, such as child, adult, and senior. A speaker’s age can influence their vocabulary and acoustic characteristics, for example, younger speakers often use swear words and laughter, whereas older speakers tend to use more truncated words in their speech ~\cite{liesenfeld2021predicting, torre2009age}.    

Reinforced Multiple Instance Learning (RL-MIL) framework has been recently proposed as an approach for speaker attribute prediction, it treats the prediction as a Multiple Instance Learning (MIL) problem, and it has succeeded in improving speaker attribute prediction by learning to select relevant speaker utterances using a Reinforcement Learning (RL) component~\cite{ziabari2024reinforced}. 

Transfer Learning refers to improving performance on a new task by using the knowledge gained from a related task that has been studied~\cite{torrey2010transfer}. Cross-language learning is an application of transfer learning used in NLP tasks, one study applied this approach to solve the problem of text classification when the source language differs from the target language~\cite{weiss2016survey}. There are several evaluation benchmarks to assess transfer learning performance, for classification tasks, accuracy and F1-score are commonly used~\cite{kabbur2010content}.  

Traditionally, speaker attribute prediction systems have relied on extensive, annotated monolingual datasets; however, real-world applications demand models that can process multilingual speech with varied grammatical structures and vocabularies~\cite{ziabari2024reinforced, aufrant2018training}. In today’s interconnected, multicultural society, speakers frequently switch between languages, and their utterances reflect diverse grammatical patterns, lexical choices, and cultural nuances~\cite{kenton2019bert, conneau-etal-2020-unsupervised}. Consequently, cross-lingual speaker attribute prediction will be more challenging than its monolingual counterpart, as models predominantly trained on English data may struggle to generalize to languages with distinct phonetic, tonal, and prosodic characteristics~\cite{wardah2025language}. The original RL-MIL implementation for speaker attribute prediction was carried out in a monolingual context (English), whether the RL-MIL framework can be effectively adapted to multilingual environments has not been intensively explored. 
\noindent The original RL\textendash MIL implementation for speaker attribute prediction was developed and evaluated in a monolingual (English) context, and its adaptation to multilingual environments remains underexplored. In this study we therefore ask to what extent modern multilingual embeddings and cross\textendash lingual transfer techniques improve the generalizability of the RL\textendash MIL framework for speaker attribute prediction in multilingual settings, as measured by Macro F1. To address this question, we (i) compare state\textendash of\textendash the\textendash art multilingual embedding approaches in their ability to capture features relevant for predicting speaker attributes; (ii) examine how multilingual encoders differ from monolingual encoders with respect to both prediction accuracy and fairness; and (iii) evaluate whether transfer learning strategies can effectively leverage knowledge from high\textendash resource languages to improve performance in low\textendash resource languages.

\section{Related Work}
\label{sec:related_work}


\subsection{Multiple Instance Learning and Reinforcement Learning}
MIL is a framework that assigns a single label to a bag of instances rather than to individual examples. MIL provides a way to handle situations where only a general label is available for a set of instances, a common occurrence in real-life data where each utterance may not individually convey the overall speaker attribute~\cite{ilse2018attention, dietterich1997solving}.

For many tasks, such as image and text analysis, additional feature extraction is required before MIL can be applied effectively, and attention mechanisms have proven useful in this regard~\cite{ilse2018attention, xu2015show, bahdanau2014neural, lin2017structured}. Recent study introduced the RL-MIL framework which combines MIL with RL to select the most informative utterances from a speaker’s bag for speaker attribute prediction~\cite{ziabari2024reinforced}. This approach not only addresses the computational challenge of processing large numbers of instances, but also ensures that only the most relevant utterances contribute to the final prediction. While reinforcement learning has been successfully applied to other NLP data selection tasks like self-training or active learning, those methods typically select entire documents or training examples~\cite{fang2017learning, chen2018learning}. In contrast, the RL-MIL framework applies a policy at a finer granularity, selecting the most informative instances within a single bag. This shifts the challenge from document-level selection to an instance-level selection. 
Subsequent work has extended this reinforcement-driven MIL paradigm toward multi-task speaker attribute prediction, demonstrating that policy-based instance selection remains effective when jointly modeling multiple speaker attributes \cite{lakeman2025reinforcementmil}.

Despite its success in monolingual settings, the original RL-MIL framework has not been explicitly extended to multilingual speech, where differences in syntax, lexicon, and phonetic realization can significantly affect the prediction of speaker attributes. This gap motivates our work, which seeks to integrate modern multilingual embeddings into the RL-MIL framework. 

\subsection{Multilingual Models}
The potential of multilingual models was first highlighted by BERT (Bidirectional Encoder Representations from Transformers) for Language Understanding, whose architecture has inspired a host of multilingual adaptations~\cite{kenton2019bert}. However, while mBERT (multilingual BERT) exhibits some cross-lingual capability, its performance on low-resource languages is limited by its design~\cite{pires2019multilingual}. To overcome these limitations, XLM-R (XLM-RoBERTa) as a transformer-based masked language model pre-trained on text in 100 languages and has achieved state-of-the-art results on cross-lingual classification, sequence labeling, and question answering tasks~\cite{conneau2019unsupervised}. This work demonstrates that scaling model capacity and training data can yield a single model that performs well across many languages, even those with limited resources. In addition, new approaches in multilingual representation learning, including a variant transformer-based model LaBSE (Language-agnostic BERT Sentence Embedding), have produced sentence embeddings that reflect both universal features and language-specific details, forming a unified representation that further enhances cross-lingual knowledge transfer~\cite{artetxe2019massively, feng2020language}. Another recent model mBART (Multilingual Bidirectional and Auto-Regressive Transformers), shows that a denoising pre-training framework originally designed for machine translation can be adapted to produce robust multilingual embeddings for speech tasks~\cite{liu2020mbart}. 

There also exist alternative multilingual models that target different goals, like the text-to-text framework exemplified by mT5 (massively multilingual Text-to-Text Transfer Transformer), which is trained on a denoising objective across 101 languages and is often oriented towards generative tasks~\cite{xue2020mt5}. Another model, like MuRIL (Multilingual Representations for Indian Languages), belongs to language-family-specific models and is optimized for Indian languages~\cite{khanuja2021muril}.
These embeddings give valuable insights to our proposed extension of RL-MIL, as they offer a unified feature space where both high-resource and low-resource languages can be effectively represented. For this study, mBERT and XLM-R were selected as the primary multilingual encoders. mBERT was chosen to serve as a strong, general-purpose baseline, while XLM-R represents the state-of-the-art for this type of model. Both are powerful encoders capable of producing the rich, contextualized utterance-level embeddings that are a direct architectural fit for the bag-of-instances approach used in the RL-MIL framework. 

\subsection{Cross-Lingual Transfer Techniques}
Cross-lingual transfer learning has been explored extensively to mitigate the scarcity of annotated data in low-resource languages. Transfer approaches can be categorized into data space and parameter space methods~\cite{aufrant2018training}. Data space transfer involves creating synthetic target data, often through annotation projection or data translation; while parameter space transfer involves borrowing model parameters from high-resource languages to improve performance on low-resource tasks~\cite{aufrant2018training, pikuliak2021cross}. 

The effectiveness of the powerful multilingual models can be limited where there is a significant domain mismatch between the pre-training corpora and the target application~\cite{pires2019multilingual}. The pre-training corpora are often composed of clean text from Wikipedia, like BERT, whereas social media text from platforms like Twitter is characterized by informal language, including slang, abbreviations, and non-standard grammar, which differs significantly from formal encyclopedic text~\cite{pires2019multilingual, huang2020multilingual, schwartz2013personality}. Yet another aspect, in multilingual social media environments, there is frequent code-switched data, where users mix languages within a single utterance, creating an additional difficulty for models not explicitly trained to handle such complex linguistic patterns~\cite{conneau2020unsupervised, schwartz2013personality}. The linguistic difference creates a domain gap that can hinder effective knowledge transfer; therefore, explicit adaptation techniques are required to create more robust, domain-invariant representations.

Recent approaches focus on learning language-invariant representations through Domain Adversarial Training (DAT), commonly implemented with a Gradient Reversal Layer (GRL)~\cite{joty2017cross, nateras2022cross, ganin2015unsupervised}. The effectiveness of this adversarial approach is well-documented across a variety of NLP tasks, including learning domain-invariant features for Named Entity Recognition (NER), creating language-invariant representations for cross-lingual sentiment classification, and adapting models to new languages in general text classification~\cite{li2020survey, chen2018adversarial, dong2020leveraging}. Building on these successes, this study investigates the application of DAT to the domain of multilingual speaker attribute prediction within the RL-MIL framework. The adversarial technique forces the encoder to produce features that cannot be distinguished by a language classifier, therefore effectively reducing the differences between source and target domains~\cite{joty2017cross}. 
Other adversarial and discrepancy-based methods also exist for domain adaptation.

Discrepancy-based methods, Deep Adaptation Networks (DAN) embed activations from several layers, directly minimise the Maximum Mean Discrepancy (MMD) between source and target domain feature distributions, and jointly optimise the standard classification loss to preserve task discrimination~\cite{long2015learning}. Another adversarial approach, Adversarial Dropout Regularization (ADR), applies two independent dropout masks within the classifier, adversarially maximises the resulting prediction discrepancy, and then trains the feature extractor to minimise it, encouraging the model to learn features that are not reliant on domain-specific cues~\cite{saito2017adversarial}. Another adversarial method, SpeakerGAN, utilizes a GAN-based objective to adapt speaker embeddings from a source language to a target language~\cite{CHEN2020211}. In this approach, a language discriminator guides the encoder to learn language-invariant representations, ensuring consistent performance even when processing unseen languages~\cite{CHEN2020211}. These studies provide a theoretical foundation for our approach, as we aim to leverage cross-lingual transfer for speaker attribute prediction in a multilingual setting. The DAT-GRL method was chosen as a strong and well-established foundation for extending the RL-MIL framework.

Building on the RL-MIL framework for speaker attribute prediction~\cite{ziabari2024reinforced}, this study aims to extend the approach by integrating state-of-the-art multilingual embeddings models, selected from mBERT~\cite{pires2019multilingual}, XLM-R~\cite{conneau2019unsupervised}, LaBSE~\cite{feng2020language}, and mBART~\cite{liu2020mbart} and combining with cross-lingual transfer techniques to enhance the prediction of speaker attributes across different languages, and to bridge the gap between high-resource and low-resource languages.

\section{Method}
\label{sec:Method}



The methodology for this project aims to extend the existing Reinforced Multiple Instance Learning (RL-MIL) framework proposed by Alireza S. Ziabari et al~\cite{ziabari2024reinforced}, to effectively handle multilingual text data for speaker attribute prediction, using modern multilingual embeddings and cross-lingual transfer techniques. This section details the datasets considered and the extensive preprocessing undertaken, outlines the baseline RL-MIL model, describes the proposed multilingual extension incorporating Domain Adversarial Training (DAT), specifies the experimental setup for replicability, and defines the evaluation metrics.

\subsection{Datasets}
Two different datasets were used to assess the model's performance in varied multilingual contexts. A detailed summary of the key characteristics of each dataset is provided in Appendix A (Table \ref{tab:dataset_overview}).

\subsubsection{Multilingual Twitter Corpus}
Initial experiments were conducted on the Multilingual Twitter Corpus for Hate Speech Detection~\cite{huang2020multilingual}. 
This dataset includes tweets in 
English, Italian, Polish, Portuguese, and Spanish. It comes with a range of metadata, including inferred demographic information like age, gender, and nationality. The experimental design makes up a few-shot, cross-lingual setting, as all five languages were present in the training, validation, and test sets. Thus, allowed the model to learn from a few examples in the low-resource languages while primarily leveraging the high-resource English data.

\subsubsection{VoxCeleb2 Datasets (Subset)}
To rigorously evaluate the model's cross-lingual generalization, a custom subset was developed using the VoxCeleb2~\cite{nagrani2020voxceleb} dataset and its corresponding metadata enriched extension~\cite{hechmi2021voxceleb}.
VoxCeleb2 is an audio-visual dataset consisting of short clips of human speech, extracted from interview videos uploaded to YouTube. 
It consists of more than 1 million utterances from 6,112 celebrities, and it comes with metadata, including the speaker's identity. The dataset is also multilingual, although the majority of recordings are in English. For this study, we utilized an extended version of the dataset's metadata, which incorporates speaker age and gender information. 

The experimental design creates a zero-shot transfer setting, as there are 40 languages in the selected subset, and some languages appeared only once. Consequently, these languages are not guaranteed to be in the training data and may appear only in the validation or test sets. Thus, forces the model to predict attributes for languages it has never encountered during training, serving as a reliable gauge of its ability to generalize.

\subsection{Data Preparation Pipeline}
The raw data from both the Multilingual Twitter Corpus and VoxCeleb2 subset required significant preprocessing to create a standardized format suitable for the model. The pipeline for each dataset is detailed below. 
\subsubsection{Multilingual Twitter Corpus}
The preparation for the Twitter corpus began with consolidating data from five separate TSV files, one for each language, into a single CSV file of approximately 108,000 tweets. During this process, a \texttt{source\_language} column was created to retain the origin of each entry. 
Subsequently, the unified dataset was cleaned by removing rows containing placeholder values (the string 'x') in important columns such as text, age, and gender. This reduced the dataset to approximately 60,000 valid entries. 
Afterwards, the text was processed through the standardized text preprocessing pipeline, with more details and rationale described in Section \ref{sec:preprocess}. 
Lastly, to allow domain adversaial training, the \texttt{source\_language} column was converted to a numerical \texttt{lang\_id}, providing a domain label for each instance.


\subsubsection{VoxCeleb2 Subset}
 Unlike the text-based Twitter corpus, the VoxCeleb2 dataset required transcription from audio. 
 This was done using the OpenAI Whisper ASR model. First, we loaded a master CSV file containing the speaker metadata. Then, systematically iterated through the directory structure, locating all individual audio clips for each speaker ID (VoxCeleb\_ID) present in the metadata file. Each audio clip was processed by the Whisper \texttt{large-v2} model, and the resulting transcript was extracted and merged back with its original speaker and clip metadata, where each row represents a single transcribed utterance. The transcription of the subset required approximately 14 hours. Due to the significant time investment, it was infeasible to process a larger portion or the full VoxCeleb2 dataset within the scope of this research.

Following transcription, A two-step age imputation was performed to fill in missing values, first by calculating age from birth and upload years, and then by propagating known ages within speaker groups. Any rows where age could not be inferred were dropped. Subsequently, the \texttt{langdetect} library was applied to each transcribed text entry to automatically determine its language, resulting in a new \texttt{lang\_code} column. The script successfully identified 40 unique languages in the subset. This process also served as a filter that checks for unqualified transcripts, rows with transcription errors, or where the language was undetectable were removed. 
The remaining transcribed text was then processed using the same standardized cleaning and normalization pipeline described in \ref{sec:preprocess}. To prepare the data for domain adversarial training, the inferred \texttt{lang\_code} was converted into the numerical \texttt{lang\_id} to serve as the domain label.  

\subsubsection{Text Preprocessing Pipeline}\label{sec:preprocess}
A standard text preprocessing pipeline was applied to all text entries from both the Twitter and VoxCeleb2 corpora. This includes normalizing characters, so converting all text to lowercase and stripping accents; removing URLs, user mentions, and hashtags; removing any non-alphabetic characters or a whitespace character, which removes emojis, special symbols, and punctuation; and collapsing multiple spaces into a single space.

An important and deliberate decision of the preprocessing pipeline was to retain stopwords and omit lemmatization. First, while stopwords removal is a standard procedure in many text classification tasks, preliminary A/B testing for this project revealed that removing them resulted in lower F1 scores for attribute prediction. This suggests that function words carry subtle but important stylistic signals that are indicative of a speaker's demographic attributes. Second, in a multilingual context, reducing words to their root lemma form can strip away valuable morphological cues. For instance, while a lemma might distinguish between "child" and "adult," it would likely erase the specific word endings that signal gender (masculine vs. feminine) or convey nuances through diminutives and augmentatives. These morphological features are considered strong cues for age and gender prediction in many languages. Therefore, to preserve these critical linguistic signals and maximize model performance, stopwords were kept, and no lemmatization was performed. 

\subsubsection{Adaptation to Ziabari's code}
Ziabari's \texttt{prepare\_data.py} script serves as the crucial link between the raw input data and the data format required by the training scripts, like \texttt{run\_mil.py}, and \texttt{run\_rlmil.py}. It is responsible for several key stages: 
\begin{enumerate}
    \item [Stage 1] parses command-line arguments to determine which dataset to process, which text embedding model to use, the desired bag size, and other parameters.
    \item [Stage 2] based on the dataset name, executes specific logic to load the raw data and perform initial preprocessing. Here, we added the logic to our specific multilingual datasets to the \texttt{create\_dataset} function: \\
    $\bullet$ User-Level Bag Creation: A critical adaptation involved transforming the data structure. The input CSV contained one row per tweet (or one row per transcribed utterance), but the MIL, or RL-MIL framework, requires one row per user, with all tweets (or utterances) from that user aggregated into a "bag" (list) of instances. 
    The text from the specified data embedded column name ('text') was aggregated into lists, forming the bags. Relevant user-level attributes ('gender', 'age', etc.) were retained by taking the 'first' observed value for each user, assuming these attributes are constant per user.\\
    $\bullet$ Age Binning Implementation: modeled directly after the \texttt{political\_data\_with\_age} example in the original script, was implemented. This involved: Converting the aggregated 'age' column from strings to numeric, and defining specific bins. For the Multilingual Twitter Corpus, which contained users as young as 11, a six-category binning strategy was used: \texttt{[float('-inf'), 18, 27, 40, 55, 70, float('inf')]} with corresponding labels \texttt{["Youth ", "Young Adult", "Adult", "Middle-aged", "Senior", "Elderly"]}. For the VoxCeleb2 Subset, a simpler three-category strategy was implemented to suit its data distribution: \texttt{[float('-inf'), 30, 55, float('inf')]} with corresponding labels \texttt{["Young", "Middle-aged", "Senior"]}.
    
    
    \item [Stage 3] Handling the text encoding using the specified transformer model. It loads the corresponding tokenizer and pre-trained model from Hugging Face using AutoTokenizer and AutoModel. For this project, in addition to the original monolingual roberta-base encoder, we will be using two multilingual encoders: bert-base-multilingual-cased (mBERT) and xlm-roberta-base (XLM-R). 

    \item [Stage 4] Iterates through each user row, applying the \texttt{create\_bag} function. This function takes the user's list of texts (the "bag"), truncates it based on \texttt{whole\_bag\_size}, and calls \texttt{get\_embeddings}. The \texttt{get\_embeddings} function then tokenizes these texts and feeds them through the loaded transformer model to generate numerical embeddings for each text using a defined pooling strategy. 

    \item [Stage 5] \texttt{create\_bag} pads these generated embeddings and creates corresponding attention masks (\texttt{bag\_embeddings}, \texttt{bag\_mask}) to ensure a consistent input size for all users. 

    \item [Stage 6] Then, it splits the fully processed user-level DataFrame into training, validation, and test sets. 
    The splitting strategy was adapted to the size of each dataset. In both cases, the splits were stratified by the 'age' column to preserve the class distribution across the sets. For the larger Multilingual Twitter Corpus, 
    80\% training, 10\% validation, and 10\% test split was used. For the much smaller VoxCeleb2 subset,
    a different split of 70\% training, 15\% validation, and 15\% test was used. This adjustment was necessary, as the RL-MIL framework uses a predefined pool size of 10. This requires the validation and test sets to contain a minimum number of samples, specifically a number greater than or equal to the pool size. Finally, each data split is saved as a seperate \texttt{.pickle} file. These pickle files are the direct input for the subsequent model training scripts.
\end{enumerate}

\subsection{The Approach}
\subsubsection{Baseline: RL-MIL}
The foundation RL-MIL framework proposed by Ziabari et al. addresses speaker attribute prediction under the MIL assumption, where each speaker is represented by a "bag" of utterances, and the prediction task is to assign a label (e.g., age group, gender) to the entire bag~\cite{ziabari2024reinforced}. The framework, consists of two main parts: an RL component for instance selection and an MIL component for prediction.
Key components:
\begin{enumerate}
    \item RL Selection: This part uses a policy network, trained via reinforcement learning, to analyze all utterances within a user's bag and select a smaller subset that is the most informative for the prediction task. The policy learns based on rewards derived from the final attribute prediction accuracy. 
    \item MIL: This part processes the subset of utterances selected by the RL component to make the final attribute prediction. It involves three stages: an initial feature transformation network, referred as autoencoder; a Multiple Instance Learning pooling layer that aggregates the transformed utterance representations into a single bag vector (using methods like Mean, Max, Attention pooling); and a MLP classification head that predicts the speaker attribute from the aggregated bag vector. The three above-mentioned pooling mechanisms aggregate the instance-level embeddings in conceptually different ways: 
    \begin{enumerate}
        \item Mean Pooling: calculates the mean of all utterance embeddings ($h_{ij}s$) in the selected subset to form a single bag-level representation; 
        \item Max pooling finds the maximum of ($h_{i,j}$) along each dimension to form the bag-level representation; 
        \item Attention pooling is more sophisticated, as it introduces learnable weights to create a weighted sum of ($h_{ij}s$), allowing it to dynamically learn the importance of each utterance and select multiple informative instances while down-weighting others. 
    \end{enumerate}
\end{enumerate}



The limitation of this RL-MIL framework was its development and evaluation primarily on monolingual data (English), and its effectiveness on diverse multilingual datasets was left unexplored, which motivates our project. 



\subsubsection{Proposed Extension: RL-MIL with DAT}

\begin{figure}[h]
\centering
  \centering
    \includegraphics[width=8cm]{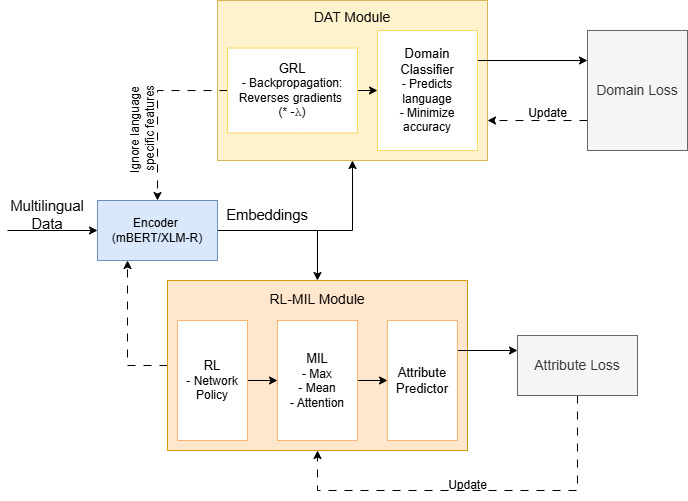}
\hfill
\caption{Methodology workflow: extended RL-MIL framework with parallel DAT module for cross-lingual speaker attribute prediction.}
\label{fig:fc}
\end{figure}

This thesis proposes extending the RL-MIL framework to effectively handle multilingual text data and improve cross-lingual generalization. The core modifications are illustrated in Figure \ref{fig:fc} and involve two main parts. The first part is the integration of multilingual encoders. The monolingual encoder used for generating initial utterance (tweet) embeddings will be replaced with a pre-trained multilingual transformer model. Modern multilingual models like mBERT, and particularly XLM-R are pre-trained on massive text corpora spanning numerous languages. They learn rich, contextualized representations that show significant cross-lingual alignment, meaning that similar concepts are often mapped to nearby points in the embedding space regardless of the source language. This feature is especially useful for building models that can generalize across different languages. Implementation would involve using the Hugging Face transformers library, the AutoModel and AutoTokenizer classes, allow for flexible integration. Implementation involves specifying the model \texttt{bert-base-multilingual-cased} and \texttt{xlm-roberta-base} via the \texttt{embedding\_model} argument during the \texttt{prepare\_data.py} execution, to experiment with different encoders. 

The second part is the DAT module for Language Invariance. While pre-trained multilingual models possess cross-lingual capabilities, explicitly encouraging the model to learn language-invariant features can specifically benefit the downstream attribute prediction task to enhance generalization, helping lower-resource languages within the dataset. 

Therefore, the initial RL-MIL setup forms a single branch in which an RL policy selects relevant speaker utterances, and the MIL module applies Max, Mean, or Attention pooling to aggregate these utterances into a final speaker embedding. Now, in addition to that primary branch, we introduce a parallel branch, the DAT module, that takes the same encoder output (embeddings) and feeds it to a domain classifier via a Gradient Reversal Layer (GRL). This adversarial objective forces the encoder to learn language-invariant features, thus reducing performance gaps across diverse linguistic contexts. 

\begin{algorithm}
\caption{GRL}
\label{alg:grl}
\begin{algorithmic}[1]
\Function{GRL\_Forward}{$x,\ \lambda$}            \Comment{$x \in \mathbb{R}^{d}$}
    \State $y \;\leftarrow\; x$                     \Comment{identity mapping}
    \State \textbf{store} $\lambda$                 \Comment{for backward step}
    \State \Return $y$
\EndFunction
\Function{GRL\_Backward}{$\nabla_{y}$}
    \State $\nabla_{x} \;\leftarrow\; -\,\lambda\,\nabla_{y}$  \Comment{flip and scale gradients}
    \State \Return $\nabla_{x}$
\EndFunction
\end{algorithmic}
\end{algorithm}

\begin{algorithm}
\caption{Domain Classifier}
\label{alg:dc}
\begin{algorithmic}[1]
\Function{DomainClassifier}{$h$} \Comment{$h \in \mathbb{R}^{d}$}
    \State $h_1 \gets \mathrm{ReLU}(W_1 h + b_1)$
    \State $\ell \gets W_2 h_1 + b_2$ \Comment{$\ell \in \mathbb{R}^{L}$}
    \State \Return $\ell$ \Comment{language logits}
\EndFunction  
\end{algorithmic}
\end{algorithm}

Algorithm~\ref{alg:grl} shows the pseudocode for the forward and backward pass behaviour of GRL. In the forward pass, it acts as an identity function; in the backward pass, it reverses the gradients and multiplies them by $\lambda$. Algorithm~\ref{alg:dc} shows the pseudocode of the domain classifier, which is initialised as a multilayer perceptron (MLP) that maps each utterance encoder embedding into vectorized language logits. The classifier is later optimised with cross-entropy against the language labels in the training loop, while its gradients flow to the shared encoder through the GRL. Finally, both branches receive feedback, one optimizing for speaker attribute prediction, the other enforcing language confusion, so that the entire model can robustly handle multilingual data while still ensuring the strengths of the original RL-MIL framework. The per-epoch training procedure is summarised, at a higher level of abstraction, in Algorithm~\ref{alg:dat}. This pseudocode includes the key part of Ziabari’s original RL-MIL loop and aggregates it with the DAT components we proposed. As a result, the total loss combines the RL policy loss, the original regularisation term, and the new domain classification loss, which enables the model to balance task performance from the original RL-MIL framework with cross-lingual generalisation. To programmatically realize this extended framework, a new training script, \texttt{run\_rlmil\_dat.py}, was developed as an extension of the original \texttt{run\_rlmil.py} script provided by Ziabari et al.
\begin{algorithm}[t]
\caption{DAT}
\label{alg:dat}
\begin{algorithmic}[1]
\Function{DAT\_forward}{}
  \For{each super-bag $B_i=\{x_{ij}\}_{j=1}^N$} \Comment{RL-MIL branch}
    \State $\mathbf h_{ij} \gets \mathrm{Encoder}(x_{ij};\,\theta_e)$
    \State $P_{ij} \gets \mathrm{Policy}(\mathbf h_{ij};\,\theta_p)$
    \State Select subset $b_i \subseteq B_i$ w.r.t.\ $P_{ij}$
    \State $\hat{y}_i \gets \mathrm{MIL}\bigl(\{\mathbf h_{ij}\}_{x_{ij}\in b_i};\,\phi\bigr)$
    \State $\mathcal{L}_{\text{task}} \gets \mathrm{CrossEntropy}(\hat{y}_i,\,y_i)$
    \State back-prop $\mathcal{L}_{\text{task}}$ \Comment{updates $\theta_e,\phi$}

    \ForAll{$x_{ij}\in b_i$} \Comment{DAT branch}
      \State $\tilde{\mathbf h}_{ij} \gets \mathrm{GRL}(\mathbf h_{ij},\,\lambda)$
      \State $\hat{\ell}_{ij} \gets \mathrm{DomainClassifier}(\tilde{\mathbf h}_{ij};\,\psi)$
    \EndFor
    \State $\displaystyle
           \mathcal{L}_{\text{domain}} \gets
           \frac{1}{|b_i|}\sum_{x_{ij}\in b_i}
           \mathrm{CrossEntropy}(\hat{\ell}_{ij},\,\ell_{ij})$

    \State Compute policy loss $\mathcal{L}_{p}$ and regulariser $\mathcal{L}_{\mathrm{reg}}$
    \State $\mathcal{L}_{\text{total}} \gets \mathcal{L}_{p} + \mathcal{L}_{\mathrm{reg}} + \mathcal{L}_{\text{domain}}$
    \State back-prop $\mathcal{L}_{\text{total}}$
    \State apply accumulated grads to $\theta_e,\phi$
    \State update $\psi,\theta_p$
    \State reset gradients
  \EndFor
\EndFunction
\end{algorithmic}
\end{algorithm}

To explain the interaction between the encoder and the DAT module, we consider a scenario, that the input is Spanish text. During the forward pass, the Spanish embeddings are passed unchanged from the encoder to the domain classifier, the domain classifier predicts the language, then computes the domain loss. During backward pass, domain classifier updates the weights to minimize domain loss, GRL reverses gradients (multiplies by hyperparameter $-\lambda$) before they reach the encoder, to force the encoder to adjust its weights to make future Spanish embeddings less Spanish-like. Over time, the encoder’s embeddings for Spanish inputs become harder to classify as Spanish, and bridge the gap between different languages. 

\begin{algorithm}
\caption{Gradient Reversal Layer}
\begin{algorithmic}[1]
\Function{GRL\_forward}{x, $\lambda$}
    \State y = x \Comment{In forward pass act as identity function}
    \State \textbf{store} $\lambda$ \Comment{Stored to be used in the backward pass}
    \State \textbf{return} $y$
\EndFunction    
\Function{GRL\_backward}{grad\_output}
    \State output = $-$ grad\_output $\times$ $\lambda$
    \Comment{Reverse the gradients and multiply by lambda}
    \State \textbf{return} output
\EndFunction
\end{algorithmic}
\end{algorithm}

\begin{algorithm}
\caption{GRL}
\label{alg:grl}
\begin{algorithmic}[1]
\Function{GRL\_Forward}{$x,\ \lambda$}            \Comment{$x \in \mathbb{R}^{d}$}
    \State $y \;\leftarrow\; x$                     \Comment{identity mapping}
    \State \textbf{store} $\lambda$                 \Comment{for backward step}
    \State \Return $y$
\EndFunction
\Function{GRL\_Backward}{$\nabla_{y}$}
    \State $\nabla_{x} \;\leftarrow\; -\,\lambda\,\nabla_{y}$  \Comment{flip and scale gradients}
    \State \Return $\nabla_{x}$
\EndFunction
\end{algorithmic}
\end{algorithm}

\begin{algorithm}
\caption{Domain Classifier}
\label{alg:dc}
\begin{algorithmic}[1]
\Function{DomainClassifier}{$h$} \Comment{$h \in \mathbb{R}^{d}$}
    \State $h_1 \gets \mathrm{ReLU}(W_1 h + b_1)$
    \State $\ell \gets W_2 h_1 + b_2$ \Comment{$\ell \in \mathbb{R}^{L}$}
    \State \Return $\ell$ \Comment{language logits}
\EndFunction  
\end{algorithmic}
\end{algorithm}

\begin{algorithm}
\caption{DAT}
\label{alg:dat}
Data: Encoder $\theta_e$, MIL head $\phi$, RL parameters $\theta_p$, Domain classifier $\psi$, GRL parameter $\lambda$, Training super-bags $\{B_i\}$ with language labels $\ell_{ij}$ 
\begin{algorithmic}[1] 
\Function{DAT\_forward}{} 
    \For{\textbf{each} super-bag $B_i=\{x_{ij}\}_{j=1}^N$} \Comment{RL-MIL branch}
    \State $\mathbf h_{ij} \leftarrow \text{Encoder}(x_{ij};\,\theta_e)$
    \State $P_{ij} \leftarrow \text{Policy}(\mathbf h_{ij};\,\theta_p)$
    \State Select subset $b_i \subseteq B_i$ w.r.t.\ $P_{ij}$
    \State $\hat{y}_i \leftarrow \text{MIL}\bigl(\{\mathbf h_{ij}\}_{x_{ij}\in b_i};\,\phi\bigr)$
    \State $\mathcal{L}_{\text{task}} \leftarrow \mathrm{CrossEntropy}(\hat{y}_i,\,y_i)$
    \State back-prop $\mathcal{L}_{\text{task}}$ \Comment{updates $\theta_e,\phi$}

    \ForAll{$x_{ij}\in b_i$} \Comment{DAT branch}
        \State $\tilde{\mathbf h}_{ij} \leftarrow \text{GRL}(\mathbf h_{ij},\,\lambda)$
        \State $\hat{\ell}_{ij}     \leftarrow \text{DomainClassifier}(\tilde{\mathbf h}_{ij};\,\psi)$
    \EndFor
    \State $\displaystyle
           \mathcal{L}_{\text{domain}} \leftarrow
           \frac{1}{|b_i|}\!\sum_{x_{ij}\in b_i}
           \mathrm{CrossEntropy}(\hat{\ell}_{ij},\,\ell_{ij})$
           
    \State Compute policy loss $\mathcal{L}_{p}$ and regulariser $\mathcal{L}_{\mathrm{reg}}$ 

    \State $\mathcal{L}_{\text{total}}
           \leftarrow
           \mathcal{L}_{p}\;+\;\mathcal{L}_{\mathrm{reg}}\;+\;
           \mathcal{L}_{\text{domain}}$
    \State back-prop $\mathcal{L}_{\text{total}}$
    \State applies accumulated grads to $\theta_e,\phi$
    \State updates $\psi,\theta_p$
    \State Reset gradients
\EndFor
\EndFunction  
\end{algorithmic}
\end{algorithm}
\subsection{Experimental Setup}
Our experimental setup is designed for a comprehensive evaluation of the proposed framework's generalization capabilities under both few-shot and zero-shot cross-lingual conditions. 
For each of the two datasets, we ran a series of experiments 
derived from the Cartesian product of three factors: the text encoder, the MIL pooling head, and the training model.

The specific setup for each factor is as follows:
\begin{itemize}
    \item Text encoders: 
    roberta-base (Monolingual),\\
    bert-base-multilingual-cased (Multilingual),\\ xlm-roberta-base (Multilingual)
    \item MIL Pooling Heads: 
    MeanMLP, MaxMLP, AttentionMLP
    \item Training frameworks: 
    MIL, RLMIL, RLMIL-DAT
    
\end{itemize}
This resulted in 27 different configurations tested on each dataset for a robust comparison of performance across various settings. While the primary research focus of this study was on the impact of the DAT module, this comprehensive setup also allows for a direct comparison of the effectiveness of the three MIL pooling heads in aggregating utterance embeddings under various monolingual and multilingual conditions. 

\subsubsection{Hyperparameters and Optimization}
Across the experiments, a consistent set of hyperparameters was used. The super bag $B_i$ of each speaker is capped at 100 instances (tweets/utterances) per bag. All runs use the same bag size $b_i$ = 20 ($b_i \in B_i$), and fixed autoencoder\_layer\_sizes = 768,256,768. 

The primary difference in setup between the two datasets lay in the DAT module's configuration. For the Multilingual Twitter Corpus, the \texttt{num\_languages} parameter for the domain classifier was set to 5, and for the VoxCeleb2 subset, the \texttt{num\_languages} parameter was set to 40, adapting to the difference in language distribution in the dataset. To find the optimal hyperparameters for each of the 27 configurations, searches are executed via Weights \& Biases Bayesian optimization. Each sweep is limited to 50 trials to find the best-performing learning rates and hidden layer dimensions, the resulting parameters for a representative run on the Twitter dataset (Seed - 42) are provided as an example in Appendix (Tables \ref{tab:best_mil}, \ref{tab:best_rlmil}, \ref{tab:best_rlmil_dat}). 
\subsection{Evaluation Metrics}

Model performance are assessed using standard metrics such as accuracy and F1-score. 

\begin{equation}
    Precision = \frac{TP}{TP + FP}
\end{equation}

\begin{equation}
    Recall = \frac{TP}{TP + FN}
\end{equation}

\begin{equation}
    F1-score = 2 \times \frac{Precision \times Recall}{Precision + Recall}
\end{equation}
For multi-class predictions (age attribute), the F1 score refers to the Macro-F1 score, which is calculated as the average of the F1 scores for each class, and each class is treated equally regardless of the number of instances that class exists in the dataset.




\section{Results}
Report tables and figures. 

\label{sec:results}

\begin{table*}[htbp]
\centering
\footnotesize
\setlength{\tabcolsep}{5pt}
\renewcommand{\arraystretch}{1.15}
\begin{tabular}{ll l rrr rrr rrr}
\toprule
\multirow{2}{*}{Encoder}
  & \multirow{2}{*}{Pooling}
  & \multirow{2}{*}{Label}
  & \multicolumn{3}{c}{\textbf{Framework}}
  & \multicolumn{3}{c}{\textbf{Statistics (vs. MIL)}}
  & \multicolumn{3}{c}{\textbf{Statistics (vs. RLMIL)}} \\
\cmidrule(lr){4-6}\cmidrule(lr){7-9}\cmidrule(lr){10-12}
 &  &  & MIL & RLMIL & RLMIL-DAT & $\Delta$ MIL & CI95 & p\_value & $\Delta$ RLMIL & CI95 & p\_value \\
\midrule
\multirow{6}{*}{0}
 & \multirow{2}{*}{Attention} & age    & 0.149 & 0.129 & 0.147 & -0.002 & 0.016 & 0.841 & 0.018 & 0.025 & 0.228 \\
 &                              & gender & 0.440 & 0.418 & 0.523 &  0.084 & 0.033 & 0.027 & 0.105 & 0.058 & 0.025 \\
 & \multirow{2}{*}{Max}        & age    & 0.144 & 0.127 & 0.137 & -0.007 & 0.032 & 0.704 & 0.011 & 0.015 & 0.235 \\
 &                              & gender & 0.379 & 0.397 & 0.534 &  \textbf{0.155} & 0.047 & \textbf{0.003} & \textbf{0.137} & 0.070 & \textbf{0.019} \\
 & \multirow{2}{*}{Mean}       & age    & 0.089 & 0.089 & 0.095 &  0.006 & 0.012 & 0.374 & 0.006 & 0.011 & 0.374 \\
 &                              & gender & 0.416 & 0.383 & 0.337 &  0.071 & 0.053 & 0.056 & 0.104 & 0.055 & 0.021 \\
\addlinespace
\multirow{6}{*}{1}
 & \multirow{2}{*}{Attention} & age    & 0.136 & 0.123 & 0.157 &  0.021 & 0.043 & 0.391 & 0.034 & 0.031 & 0.096 \\
 &                             & gender & 0.473 & 0.442 & 0.576 &  0.103 & 0.097 & 0.107 & \textbf{0.134} & 0.077 & \textbf{0.027} \\
 & \multirow{2}{*}{Max}       & age    & 0.114 & 0.108 & 0.154 &  0.040 & 0.038 & 0.111 & 0.046 & 0.036 & 0.064 \\
 &                             & gender & 0.477 & 0.459 & 0.551 &  0.074 & 0.134 & 0.342 & 0.092 & 0.105 & 0.161 \\
 & \multirow{2}{*}{Mean}      & age    & 0.134 & 0.098 & 0.140 &  0.006 & 0.038 & 0.764 & 0.042 & 0.024 & 0.028 \\
 &                             & gender & 0.508 & 0.437 & 0.584 &  0.076 & 0.093 & 0.185 & \textbf{0.147} & 0.078 & \textbf{0.021} \\
\addlinespace
\multirow{6}{*}{2}
 & \multirow{2}{*}{Attention} & age    & 0.101 & 0.099 & 0.127 &  0.026 & 0.023 & 0.087 & 0.028 & 0.019 & 0.047 \\
 &                             & gender & 0.370 & 0.373 & 0.389 &  0.019 & 0.038 & 0.374 & 0.017 & 0.033 & 0.374 \\
 & \multirow{2}{*}{Max}       & age    & 0.126 & 0.120 & 0.131 &  0.005 & 0.017 & 0.610 & 0.010 & 0.014 & 0.216 \\
 &                             & gender & 0.373 & 0.373 & 0.542 &  \textbf{0.169} & 0.032 & \textbf{0.001} & \textbf{0.169} & 0.032 & \textbf{0.001} \\
 & \multirow{2}{*}{Mean}      & age    & 0.089 & 0.089 & 0.089 &  0.000 & 0.000 & 0.070 & 0.000 & 0.000 & 1.000 \\
 &                             & gender & 0.373 & 0.373 & 0.373 &  0.000 & 0.000 & 0.178 & 0.000 & 0.000 & 1.000 \\
\bottomrule
\end{tabular}
\caption{Mean F1-scores and statistical significance on the Multilingual Twitter corpus (n=5 seeds). Encoder id: 0 = roberta-base, 1 = bert-base-multilingual-cased, 2 = xlm-roberta-base.}
\label{tab:gain}
\end{table*}

In this section, we will report the global macro-F1 (mean over 5 seedings) for comparability across 27 configurations and acknowledge that per-language scores would isolate the transfer effect more directly. 
Nevertheless, the consistent gains of RLMIL-DAT across multiple configurations provide strong evidence that the DAT module improves cross-lingual generalizability.

To assess the statistical significance of the performance differences between the frameworks, we employed the parametric paired t-test between the frameworks, after verifying the normality. First, we present the results from the experiments on the few-shot Multilingual Twitter Corpus, which was the larger and primary dataset for this study. Comparable evaluation challenges, including sensitivity to domain mismatch and variability introduced by model and data choices, have also been reported in comparative analyses of deep learning-based speaker modeling systems \cite{vanbeveren2025speakerverification}.


In Table \ref{tab:gain}, the column "$\Delta MIL$" denotes the gains in performance of RLMIL-DAT relative to MIL, while column "$\Delta RLMIL$" denotes the gains relative to RLMIL. Of the 36 encoder, pooling, attribute combinations, 30 exhibit positive gains, 
only 2 show small negative differences, 
and 4 remain unchanged. The most significant performance lift was for MaxMLP pooling on encoder 2 (xlm-roberta-base) for gender prediction, RLMIL-DAT boosts F1 by 0.169 over MIL and by 0.169 over RLMIL, both having a $p$\_value of 0.001, proving there is a statistically significant difference between these frameworks. Similarly, MaxMLP on encoder 0 (roberta-base) for gender achieves gains of 0.155 versus MIL and 0.137 versus RLMIL, both having a $p$\_value smaller than 0.05, demonstrating its significance. However, it incurs a minor drop (-0.007) in age prediction compared to MIL, which may be attributable to noise. In contrast, while the encoder 1 (bert-base-multilingual-cased) also shows a positive lift with MaxMLP, these improvements are not statistically significant (p > 0.05). Interestingly, for the same encoder (encoder 1), the DAT module provided the largest and most significant improvements over RLMIL when paired with MeanMLP ($\Delta=0.147, p=0.021$) and AttentionMLP ($\Delta=0.134, p=0.027$). This contrasts with the other encoders, where the gains from DAT were most pronounced with the MaxMLP pooling head.

\begin{figure}[t]
  \centering
  \begin{subfigure}[t]{0.48\textwidth}
    \centering
    \includegraphics[width=\textwidth]{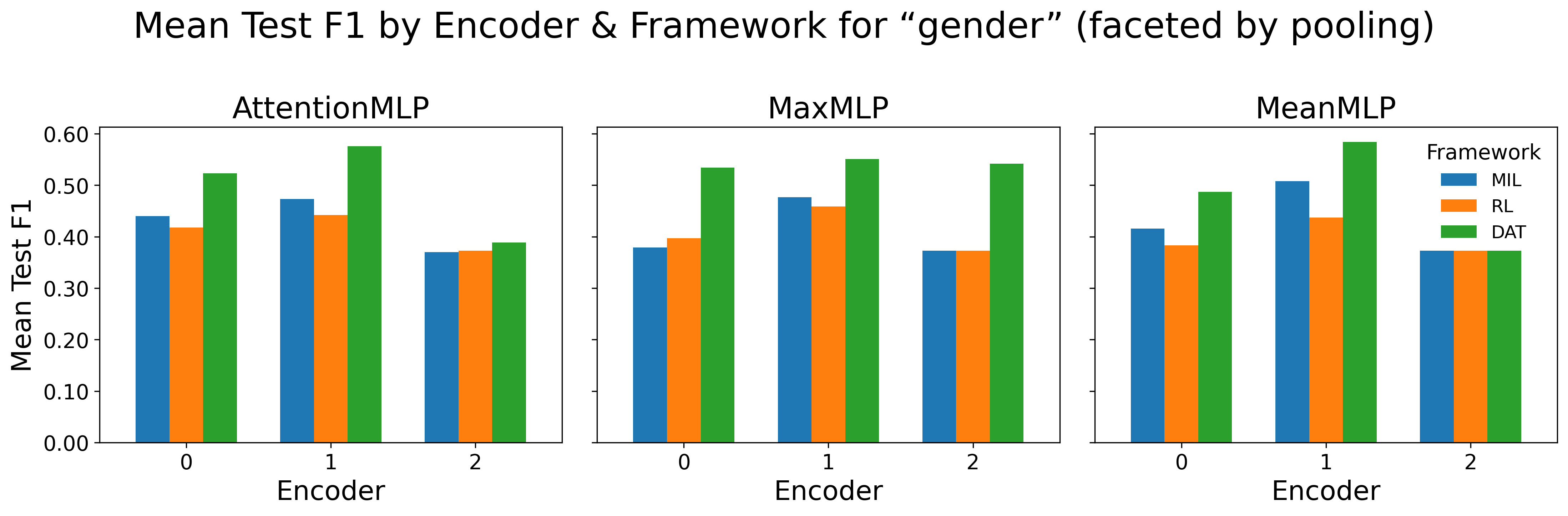}
    \caption{Gender (MeanMLP)}
    \label{fig:bar_gender}
  \end{subfigure}\hfill
  \begin{subfigure}[t]{0.48\textwidth}
    \centering
    \includegraphics[width=\textwidth]{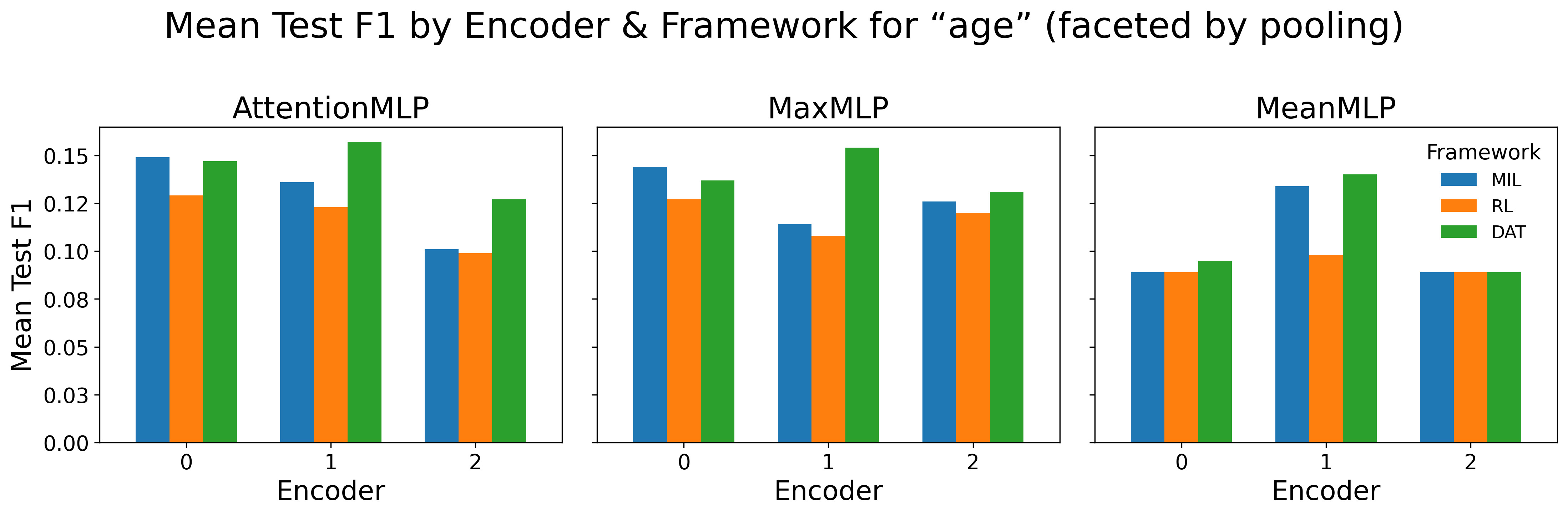}
    \caption{Age (MeanMLP)}
    \label{fig:bar_age}
  \end{subfigure}
  \caption{Test F1 by encoder and model, faceted by pooling (Twitter).}
  \label{fig:bar}
\end{figure}

A more granular view of these performance trends is shown in Figure \ref{fig:bar}, which visualizes the mean F1 scores from the 5-seed experiments. For gender prediction, RLMIL-DAT (green bars) consistently outperformed all three encoders and all three pooling methods. This demonstrates the robustness of the DAT module for this task, 
and signifying that the adversarial training is successfully forcing the model to learn language-invariant features even without multilingual pre-training (encoder 0). For age prediction, the results confirm that this is a more challenging task with less distinct patterns. While RLMIL-DAT still consistently outperforms RLMIL, the marginal improvements are smaller, and the performance of standard MIL is more competitive, in some cases outperforming both RLMIL and RLMIL-DAT.   

\begin{figure}[h!]
    \centering

    \begin{subfigure}[b]{0.33\textwidth}
        \centering
        \includegraphics[width=\textwidth]{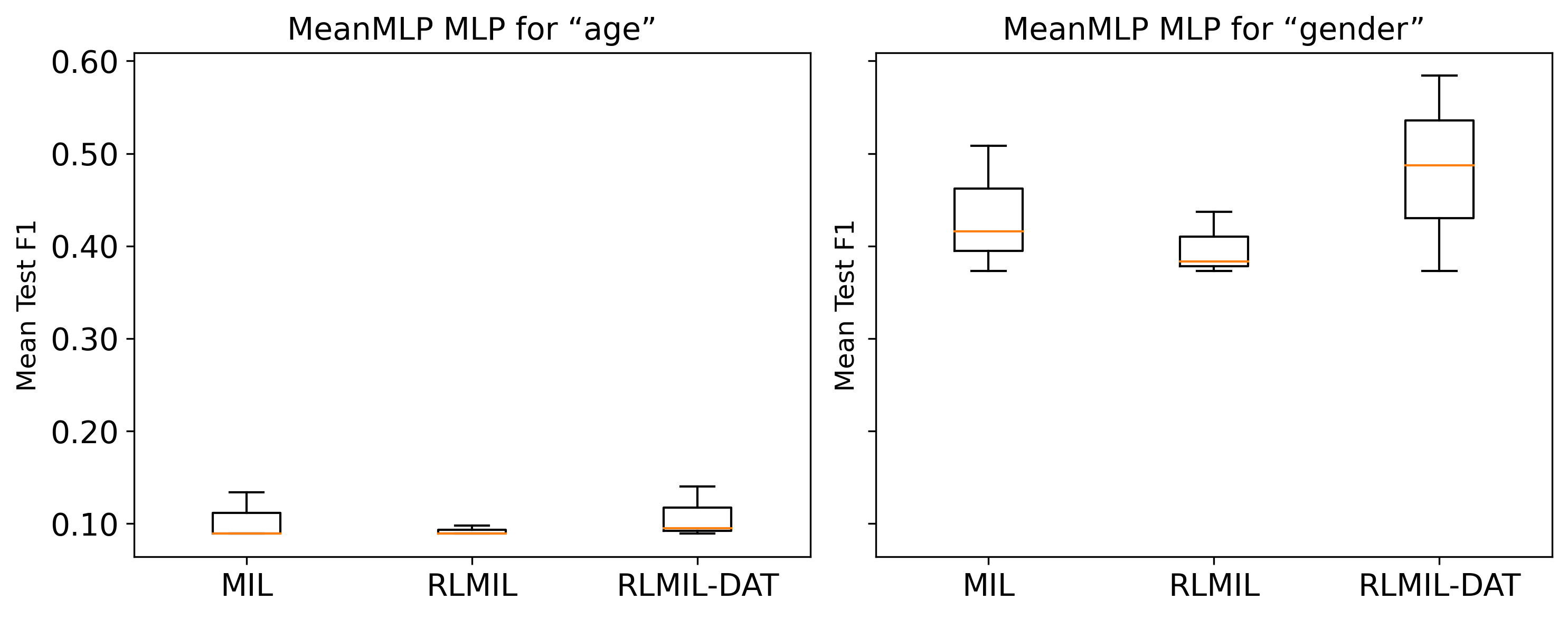}
        \label{fig:boxplot_mean}
    \end{subfigure}
    \hfill
    \begin{subfigure}[b]{0.33\textwidth}
        \centering
        \includegraphics[width=\textwidth]{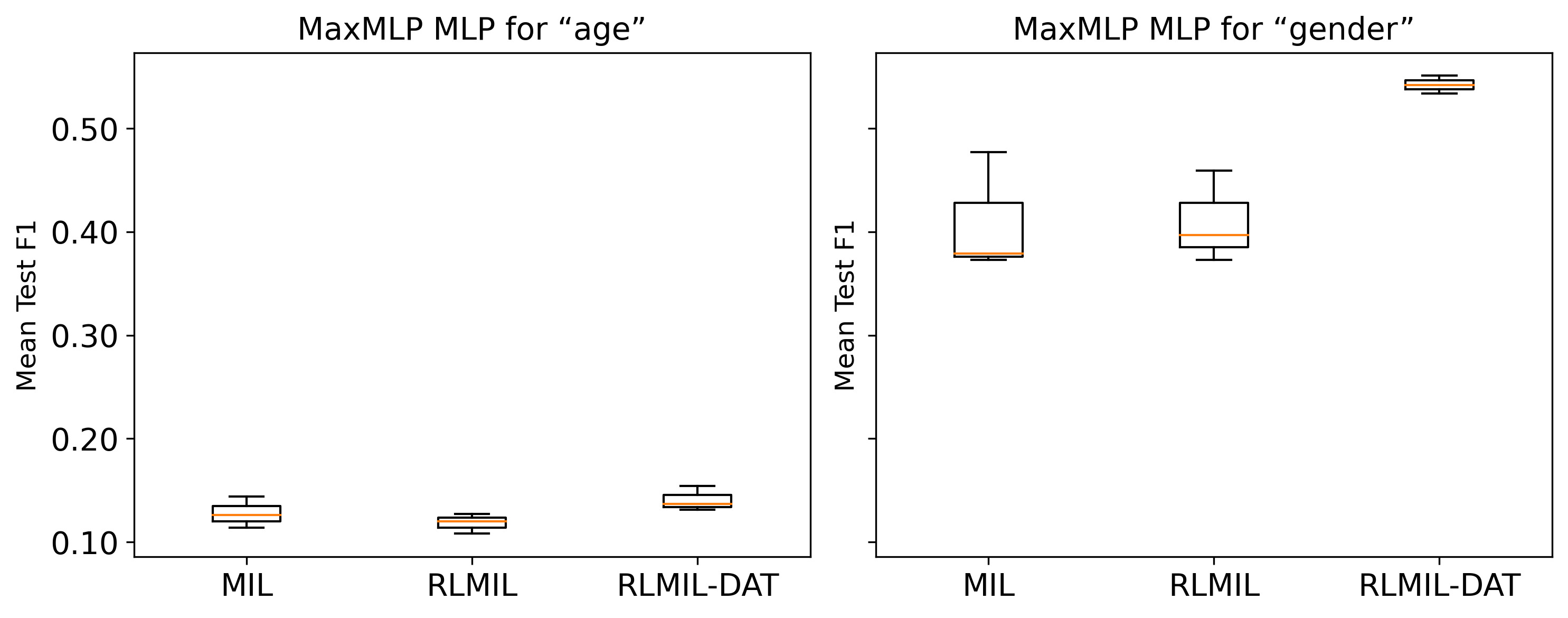}
        \label{fig:boxplot_max}
    \end{subfigure}
    \hfill
    \begin{subfigure}[b]{0.33\textwidth}
        \centering
        \includegraphics[width=\textwidth]{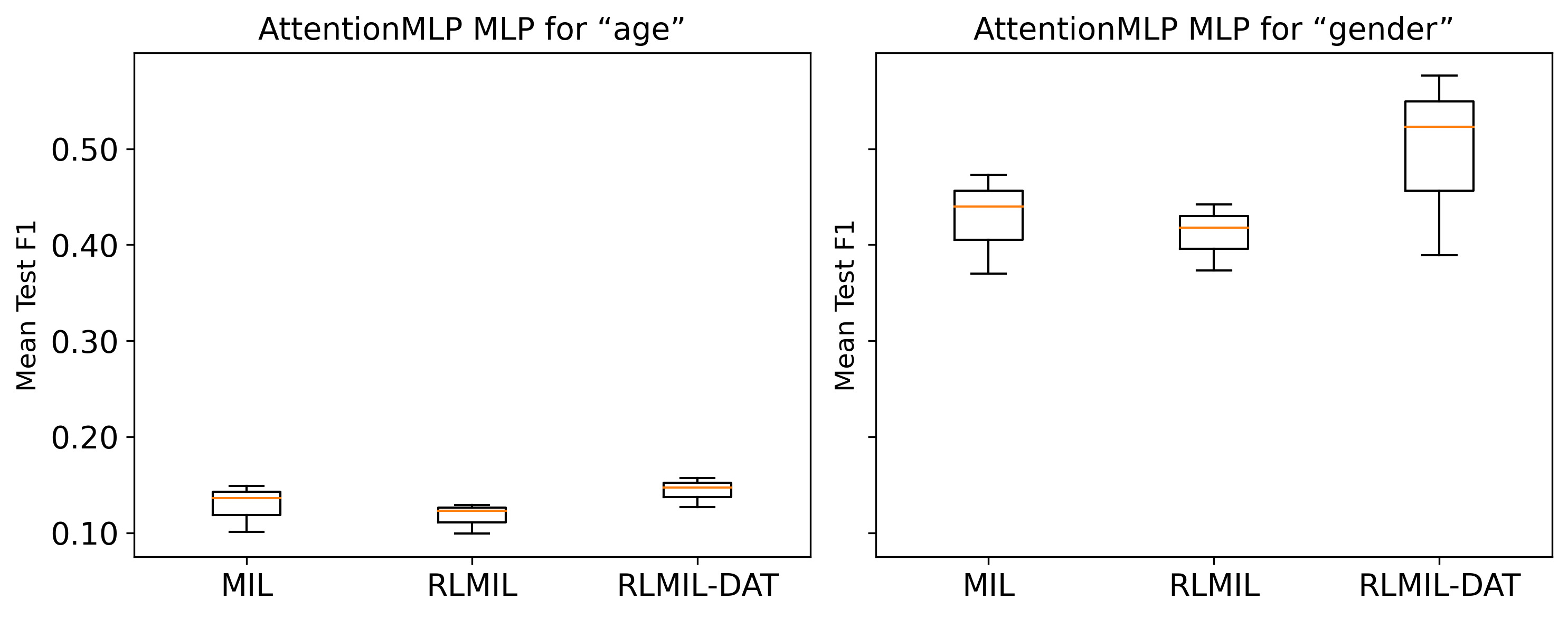}
        \label{fig:boxplot_attention}
    \end{subfigure}
    
    \caption{Mean Test F1 Distribution by pooling (Twitter).}
    \label{fig:boxplot}
\end{figure}

Figure \ref{fig:boxplot} further supports the observations by showing the distribution of the F1 test scores across the five seeds. For gender prediction, across all three pooling strategies, RLMIL-DAT models consistently achieve a higher median F1 score, indicating better average performance. However, the interquartile ranges for RLMIL-DAT are sometimes wider than the baselines, indicating greater variance, yet their overall position remains consistently higher. For age prediction, the situation is less distinct. While RLMIL-DAT still tends to yield a higher median performance, the overall F1 scores are much lower, and the distinction between the models is less pronounced.

While the aggregate results in Table \ref{tab:gain} and Figure \ref{fig:bar} are averaged across five seeds for statistical robustness, the following qualitative analysis of model behavior is based on a single, representative seed (Seed 42). A closer inspection of the train F1 versus test F1 scores reveals distinct patterns of overfitting and underfitting across the different frameworks (Table \ref{tab:pm_models}). The standard MIL framework, especially with pooling methods MeanMLP and MaxMLP, generally tends to overfit $train_{f1} > test_{f1}$, indicating that the model appears to memorize characteristics of the training bags rather than learning generalizable features. The more sophisticated AttentionMLP head mitigates this, showing a more balanced profile, for example, roberta-base for gender: train 0.614 vs. test 0.615. On the contrary, the RLMIL framework exhibits a consistent and severe underfitting pattern across all configurations ($train_{f1} < test_{f1}$), suggesting that the RL agent is struggling to select the representative instances, thus failing to train the classifier effectively.  The RLMIL-DAT models present a more balanced and desirable result. For gender prediction, the test F1 score significantly surpasses the train F1 score. For example, roberta-base with MeanMLP: train 0.574 vs. test 0.672, demonstrating a robust generalization. Although, it can also produce balanced or slightly overfit profiles in some configurations, this framework has constantly achieved higher test F1 scores than the other frameworks, thereby bridging the generalization gap.  

The results for the zero-shot experiments on the much smaller VoxCeleb2 subset are available in Appendix (Table \ref{tab:gain_voxceleb}). while the RLMIL-DAT framework often yields positive performance gains over the baselines, the improvements are generally not statistically significant ($p > 0.05$). This is likely due to the limited size of the dataset ($93$ bags) after preprocessing, which provides low statistical power for the tests. For this reason, our primary analysis has focused on the more statistically robust findings from the larger Twitter dataset.

\section{Discussion}
\label{sec:discussion}
This section revisits the main question and the revised sub-questions, then situates the findings against prior work, limitations, and future directions.

\medskip
\noindent\textit{Answering the main question.}
Across configurations and seeds, gains are substantial and are driven primarily by the cross-lingual transfer mechanism rather than by the choice of multilingual encoder. As measured by Macro F1, RLMIL--DAT consistently outperforms standard MIL and the original RL--MIL, with the adversarial component contributing most of the improvement (Table~\ref{tab:gain}, Figure~\ref{fig:boxplot}).

\medskip
\noindent\textit{Encoder effects.}
The impact of DAT depends on the starting cross-lingual capacity of the encoder. The monolingual \texttt{roberta-base} benefits the most: without prior multilingual pretraining, adversarial training compels the encoder to learn language-invariant features, yielding large and statistically significant gains. \texttt{bert-base-multilingual-cased} (mBERT) also improves reliably under DAT, typically with steadier learning dynamics, consistent with its partial but imperfect cross-lingual alignment. By contrast, \texttt{xlm-roberta-base} (XLM-R)---already strong on cross-lingual alignment---shows smaller or no gains in the simplest pooling regime. With Mean pooling, which averages instance representations, DAT often provides little additional benefit for XLM-R (Table~\ref{tab:gain}, Figure~\ref{fig:bar}). More expressive pooling (Max or Attention) preserves informative instance-level variation, re-exposing residual language cues for the domain classifier and restoring a clear DAT benefit even with XLM-R.

\medskip
\noindent\textit{Pooling and instance selection.}
Mean pooling can dilute subtle language-specific signals, limiting the leverage available to the domain classifier and thus the gradient-reversal signal. Max and Attention pooling better retain discriminative instance structure, which synergizes with DAT and yields the largest improvements in our experiments. This interaction helps explain why RLMIL--DAT most strongly lifts gender prediction, where salient lexical and stylistic cues are more concentrated at the utterance level (Figures~\ref{fig:bar}, \ref{fig:boxplot}).

\medskip
\noindent\textit{Transfer regimes.}
In the few-shot setting on the larger multilingual Twitter corpus, DAT consistently and significantly improves performance, evidencing effective transfer from the high-resource source language to lower-resource languages. In the zero-shot setting on the VoxCeleb2 subset, improvements are mostly positive but not consistently significant, likely due to the very small number of bags (93), reduced statistical power, and the greater difficulty of generalizing across 40 languages. These results suggest that RLMIL--DAT is promising for zero-shot transfer, but larger and more balanced datasets are needed for conclusive significance analyses.

\medskip
\noindent Overall, the data indicate that combining selective instance aggregation with adversarial domain confusion is a practical recipe for multilingual robustness: DAT provides the bulk of the cross-lingual gains, while encoder and pooling choices modulate how much of that potential is realized.

\subsection{Comparison with the Existing Studies}
Across all experiments, F1 scores for gender prediction were substantially higher than for age prediction. Our age prediction performance (Twitter 0.09 to 0.16, VoxCeleb2 0.28) was notably lower than that reported in the baseline study by Ziabari et al. (0.31 to 0.45). The discrepancy can be attributed to a convergence of many factors, including selected data, task complexity, and the impact of the linguistic context. First, the nature of the text differs dramatically. Our multilingual Twitter corpus consists of short, colloquial, topic-driven posts, whereas the original study used a dataset of long, formal political speeches. Age-related cues, like specific vocabulary, generational slang, are likely much weaker, more scattered, and less frequent in short tweets than in long speeches. Second, we have used a finer age binning strategy with six categories to accommodate our dataset's wider user age range (11 years to 96 years), compared to the four categories used by Ziabari et al. for speakers aged 27 and older. This increases the number of classes and also exacerbates class imbalance. Lastly, a multilingual setting also introduces cross-lingual noise, a challenge that does not exist in a monolingual environment. These combinations appear to have negatively impacted the RL selection mechanism. For age prediction, the RL agent may have struggled to learn an optimal selection policy. This difficulty likely derives from common RL challenges, including reward sparsity, where the signal from the final prediction is not frequent enough to guide the agent effectively; and a difficult exploration-exploitation trade-off, where the agent fails to explore the instance space sufficiently to discover the most informative utterances. 

To address the RL-related failures and create a more robust system, several improvements could be explored. As detailed in Section \ref{sec:future_work}, augmenting the RL reward function to be domain-aware could provide a denser and more informative signal for the agent. Additionally, experimenting with different policy network's capacity may allow the agent to learn a more complex and effective selection strategy, helping to resolve the underfitting issue discovered in Section \ref{sec:results}.


\subsection{Limitations}


\subsubsection{Reproducibility and Scalability} 
Running all 27 experiment configurations required a total of 
554.12 GPU hours for the Twitter dataset and 250.84 for the VoxCeleb dataset on a dual A100 setup. We ran these experiments in parallel to reduce the real-world timeline, which is equivalent to over 33 days of sequential processing. This high computational demand may limit the reproducibility and scalability of the framework, as the overhead from the DAT module would increase exponentially with more languages or attributes.

To run all 27 experimental configurations required 554.12 GPU hours for the Twitter dataset, and 250.84 GPU hours for the VoxCeleb dataset on a dual A100 setup. Although we ran these experiments in parallel to reduce the real-world timeline, the total compute time is equivalent to over 33 days of sequential processing. This high computational demand may limit the reproducibility. Furthermore, the scalability of the framework is an additional concern, as the computational overhead introduced by the DAT module would increase exponentially if the model were extended to hundreds of languages or a larger set of attributes.

\subsubsection{Encoder Optimization} \label{sec:future_work}
Another limitation 
is that the pre-trained multilingual encoders were utilized as fixed feature extractors rather than being fine-tuned on our specific task. This was a considered decision to isolate and evaluate the performance impact of the DAT module, 
ensure the observed performance gains can be directly attributed to the effectiveness of the adversarial training strategy, and not the combination of DAT and task-specific fine-tuning. However, we acknowledged that this approach may not yield the best possible performance, and fine-tuning could improve the prediction accuracy.

\section{Conclusions}
\label{sec:conclusion}

This paper addresses the challenge of multilingual speaker attribute prediction by extending the monolingual RL-MIL framework. We proposed and evaluated the extended RLMIL-DAT framework, which integrates a DAT module to foster the learning of language-invariant features. Experiments on the multilingual tweet corpus in a few-shot cross-lingual setting, show that RLMIL-DAT consistently outperformed both standard MIL and the RLMIL architectures. This performance gain was most substantial for gender prediction, suggesting that the framework is particularly effective for attributes with strong and transferable linguistic signals. The results confirm that the DAT module is the key to this improvement; it is more critical than simply switching between multilingual embeddings. This work successfully validates the extensibility of the RL-MIL architecture. However, we acknowledge that our findings on textual data may not directly apply to spoken language, and that a more fine-grained, per-language analysis was not performed.

For future researchers, this work articulates a clear trade-off: standard RL-MIL is a simpler and less computationally expensive choice for monolingual tasks; RLMIL-DAT, despite its increased implementation complexity and computational overhead, is the recommended choice for building robust and fair multilingual systems due to its superior ability to generalize across languages.

\subsection{Funding}
No funding was received for this work.

\subsection{Consent to Publish declaration}
Not applicable.

\subsection{Ethics and Consent to Participate declarations}
Not applicable.

\subsection*{Data Availability}

All source data are publicly available: the Multilingual Twitter Corpus for Hate Speech Detection \cite{huang2020multilingual} and VoxCeleb2 \cite{nagrani2020voxceleb}. 

\bibliography{refs}   


\onecolumn

\appendix
\begin{appendices}

\section{Supplemental Tables}
\label{sec:apx:first_appendix}

\begin{table}[ht]
    \centering
    \begin{tabular}{@{}lrrrrl@{}}
        \toprule
        \textbf{Dataset Name}              &
        \textbf{\# Bags (Users)}           &
        \textbf{\# Instances (Utterances)}              &
        \textbf{\# Languages}              &
        \textbf{Setting}                   &
        \textbf{Labels}                    \\ \midrule
        Multilingual Twitter Corpus & 527 & 59{,}054 & 5  & Few-shot  & Age, Gender \\
        VoxCeleb2 Subset            & 93  & 26{,}765 & 40 & Zero-shot & Age, Gender \\ \bottomrule
    \end{tabular}
    \caption{Overview of datasets used in the study.}
    \label{tab:dataset_overview}
\end{table}

\begin{table*}[ht]
\centering
\setlength{\tabcolsep}{5pt}
\renewcommand{\arraystretch}{1.15}
\begin{tabular}{ll l rrr rrr rrr}
\toprule
\multirow{2}{*}{Encoder}
  & \multirow{2}{*}{Pooling}
  & \multirow{2}{*}{Label}
  & \multicolumn{3}{c}{\textbf{Framework}}
  & \multicolumn{3}{c}{\textbf{Statistics}}
  & \multicolumn{3}{c}{\textbf{Statistics}} \\
\cmidrule(lr){4-6}\cmidrule(lr){7-9}\cmidrule(lr){10-12}
 &  &  & MIL & RL & DAT & $\Delta MIL$ & ci95\_M & p\_value & $\Delta RLMIL$ & ci95 & p\_value \\
\midrule
\multirow{6}{*}{0}
 & \multirow{2}{*}{Attention} & age    & 0.284 & 0.284 & 0.284 &  0.000 & 0.000 & 1.000 &  0.000 & 0.000 & 1.000 \\
 &                                & gender & 0.431 & 0.431 & 0.438 &  0.007 & 0.056 & 0.822 &  0.007 & 0.056 & 0.822 \\
 & \multirow{2}{*}{Max}       & age    & 0.284 & 0.284 & 0.284 &  0.000 & 0.000 & 1.000 &  0.000 & 0.000 & 1.000 \\
 &                                & gender & 0.431 & 0.431 & 0.457 &  0.027 & 0.080 & 0.546 &  0.027 & 0.080 & 0.546 \\
 & \multirow{2}{*}{Mean}      & age    & 0.284 & 0.284 & 0.284 &  0.000 & 0.000 & 1.000 &  0.000 & 0.000 & 1.000 \\
 &                                & gender & 0.431 & 0.431 & 0.432 &  0.001 & 0.011 & 0.864 &  0.001 & 0.011 & 0.864 \\
\addlinespace
\multirow{6}{*}{1}
 & \multirow{2}{*}{Attention} & age    & 0.281 & 0.288 & 0.286 &  0.005 & 0.010 & 0.374 & -0.003 & 0.006 & 0.374 \\
 &                                & gender & 0.418 & 0.411 & 0.415 & -0.003 & 0.006 & 0.374 &  0.004 & 0.007 & 0.374 \\
 & \multirow{2}{*}{Max}       & age    & 0.284 & 0.284 & 0.283 & -0.001 & 0.032 & 0.952 & -0.001 & 0.032 & 0.952 \\
 &                                & gender & 0.400 & 0.367 & 0.459 &  0.059 & 0.110 & 0.353 &  0.093 & 0.134 & 0.246 \\
 & \multirow{2}{*}{Mean}      & age    & 0.281 & 0.281 & 0.271 & -0.010 & 0.058 & 0.755 & -0.010 & 0.058 & 0.757 \\
 &                                & gender & 0.426 & 0.421 & 0.417 & -0.009 & 0.102 & 0.874 & -0.003 & 0.103 & 0.952 \\
\addlinespace
\multirow{6}{*}{2}
 & \multirow{2}{*}{Attention} & age    & 0.284 & 0.284 & 0.284 &  0.000 & 0.000 & 1.000 &  0.000 & 0.000 & 1.000 \\
 &                                & gender & 0.431 & 0.431 & 0.431 &  0.000 & 0.000 & 0.070 &  0.000 & 0.000 & 1.000 \\
 & \multirow{2}{*}{Max}       & age    & 0.284 & 0.284 & 0.284 &  0.000 & 0.000 & 1.000 &  0.000 & 0.000 & 1.000 \\
 &                                & gender & 0.431 & 0.431 & 0.429 & -0.001 & 0.003 & 0.374 & -0.001 & 0.003 & 0.374 \\
 & \multirow{2}{*}{Mean}      & age    & 0.284 & 0.284 & 0.284 &  0.000 & 0.000 & 1.000 &  0.000 & 0.000 & 1.000 \\
 &                                & gender & 0.431 & 0.431 & 0.474 &  0.043 & 0.085 & 0.374 &  0.043 & 0.085 & 0.374 \\
\bottomrule
\end{tabular}
\caption{Mean F1-Scores and Statistical Significance on the VoxCeleb2 Subset (n=5 Seeds). Encoder id: 0 = roberta-base, 1 = bert-base-multilingual-cased, 2 = xlm-roberta-base.}
    \label{tab:gain_voxceleb}
\end{table*}

\begin{table}[ht]
  \centering
  \small           
  \begin{tabular}{llclrrr}
    \toprule
    \textbf{Model} & \textbf{Pooling} & \textbf{Enc.} &
    \textbf{Attribute} & \textbf{best\_eval\_f1} & \textbf{train\_f1} & \textbf{test\_f1} \\
    \midrule
    \multicolumn{7}{@{}l}{}\\
    MIL    & Attention & 0 & age    & 0.205 & 0.157 & 0.181 \\
           &           &   & gender & 0.552 & 0.614 & 0.615 \\
           & Max       &   & age    & 0.252 & 0.162 & 0.153 \\
           &           &   & gender & 0.388 & 0.451 & 0.398 \\
           & Mean      &   & age    & 0.209 & 0.175 & 0.082 \\
           &           &   & gender & 0.508 & 0.465 & 0.398 \\
    \addlinespace
    RLMIL  & Attention & 0 & age    & 0.200 & 0.078 & 0.133 \\
           &           &   & gender & 0.521 & 0.408 & 0.482 \\
           & Max       &   & age    & 0.199 & 0.109 & 0.157 \\
           &           &   & gender & 0.369 & 0.337 & 0.403 \\
           & Mean      &   & age    & 0.091 & 0.046 & 0.088 \\
           &           &   & gender & 0.513 & 0.415 & 0.470 \\
    \addlinespace
    RLMIL--DAT & Attention & 0 & age    & 0.340 & 0.196 & 0.162 \\
               &           &   & gender & 0.638 & 0.577 & 0.628 \\
               & Max       &   & age    & 0.204 & 0.131 & 0.144 \\
               &           &   & gender & 0.662 & 0.564 & 0.623 \\
               & Mean      &   & age    & 0.198 & 0.104 & 0.118 \\
               &           &   & gender & 0.662 & 0.574 & 0.672 \\
    \midrule
    \multicolumn{7}{@{}l}{}\\
    MIL    & Attention & 1 & age    & 0.233 & 0.234 & 0.144 \\
           &           &   & gender & 0.654 & 0.671 & 0.650 \\
           & Max       &   & age    & 0.233 & 0.217 & 0.143 \\
           &           &   & gender & 0.633 & 0.696 & 0.599 \\
           & Mean      &   & age    & 0.220 & 0.187 & 0.126 \\
           &           &   & gender & 0.628 & 0.653 & 0.616 \\
    \addlinespace
    RLMIL  & Attention & 1 & age    & 0.180 & 0.082 & 0.133 \\
           &           &   & gender & 0.570 & 0.476 & 0.542 \\
           & Max       &   & age    & 0.182 & 0.090 & 0.127 \\
           &           &   & gender & 0.640 & 0.550 & 0.593 \\
           & Mean      &   & age    & 0.136 & 0.074 & 0.085 \\
           &           &   & gender & 0.546 & 0.451 & 0.531 \\
    \addlinespace
    RLMIL--DAT & Attention & 1 & age    & 0.276 & 0.427 & 0.193 \\
               &           &   & gender & 0.650 & 0.601 & 0.599 \\
               & Max       &   & age    & 0.228 & 0.171 & 0.137 \\
               &           &   & gender & 0.647 & 0.571 & 0.631 \\
               & Mean      &   & age    & 0.356 & 0.182 & 0.178 \\
               &           &   & gender & 0.650 & 0.617 & 0.611 \\
    \midrule
    \multicolumn{7}{@{}l}{}\\
    MIL    & Attention & 2 & age    & 0.165 & 0.176 & 0.144 \\
           &           &   & gender & 0.515 & 0.408 & 0.398 \\
           & Max       &   & age    & 0.193 & 0.148 & 0.140 \\
           &           &   & gender & 0.369 & 0.352 & 0.398 \\
           & Mean      &   & age    & 0.091 & 0.088 & 0.088 \\
           &           &   & gender & 0.515 & 0.424 & 0.398 \\
    \addlinespace
    RLMIL  & Attention & 2 & age    & 0.192 & 0.075 & 0.119 \\
           &           &   & gender & 0.369 & 0.366 & 0.398 \\
           & Max       &   & age    & 0.178 & 0.078 & 0.129 \\
           &           &   & gender & 0.369 & 0.336 & 0.398 \\
           & Mean      &   & age    & 0.091 & 0.048 & 0.088 \\
           &           &   & gender & 0.369 & 0.336 & 0.398 \\
    \addlinespace
    RLMIL--DAT & Attention & 2 & age    & 0.309 & 0.235 & 0.178 \\
               &           &   & gender & 0.567 & 0.370 & 0.525 \\
               & Max       &   & age    & 0.228 & 0.159 & 0.191 \\
               &           &   & gender & 0.645 & 0.545 & 0.560 \\
               & Mean      &   & age    & 0.091 & 0.046 & 0.088 \\
               &           &   & gender & 0.369 & 0.335 & 0.398 \\
    \bottomrule
  \end{tabular}
  \caption{Performance metrics for each \textbf{model}, \textbf{pooling} strategy and \textbf{encoder} (Twitter - Seed 42). (0: roberta-base, 1 :
bert-base-multilingual-cased, 2 : xlm-roberta-base).}
  \label{tab:pm_models}
\end{table}

\begin{table*}[ht]
\centering
\small
\begin{tabular}{lllrrrrr}
\toprule
\multirow{2}{*}{Encoder}
  & \multirow{2}{*}{Pooling}
  & \multirow{2}{*}{Label} &
\multicolumn{5}{c}{\textbf{Hyper-parameters}} \\
\cmidrule(lr){4-8}
\multicolumn{3}{c}{} &
LR & Batch & Epochs & Hdim & Early stopping \\
\midrule
0 & Attention & age    & 0.000137 & 64 & 239 & 256 & 10 \\
0 & Attention & gender & 0.000179 & 64 & 189 & 256 & 10 \\
0 & Max       & age    & 0.000277 & 64 & 302 & 256 & 10 \\
0 & Max       & gender & 0.000118 & 64 &  53 & 512 & 10 \\
0 & Mean      & age    & 0.000691 & 64 & 284 & 256 & 10 \\
0 & Mean      & gender & 0.000478 & 32 & 344 &  64 & 10 \\
1 & Attention & age    & 0.000759 & 64 & 365 &  64 & 10 \\
1 & Attention & gender & 0.000495 & 64 & 155 & 256 & 10 \\
1 & Max       & age    & 0.005760 & 64 & 356 &  64 & 10 \\
1 & Max       & gender & 0.002309 & 64 &  61 & 512 & 10 \\
1 & Mean      & age    & 0.000121 & 16 & 134 &  64 & 10 \\
1 & Mean      & gender & 0.000142 &  8 & 301 & 512 & 10 \\
2 & Attention & age    & 0.000227 & 16 & 158 & 512 & 10 \\
2 & Attention & gender & 0.000241 & 64 & 349 & 512 & 10 \\
2 & Max       & age    & 0.000562 & 64 & 220 & 512 & 10 \\
2 & Max       & gender & 0.000315 & 16 & 319 & 512 & 10 \\
2 & Mean      & age    & 0.000590 &  8 & 190 & 256 & 10 \\
2 & Mean      & gender & 0.000109 & 32 & 322 & 256 & 10 \\
\bottomrule
\end{tabular}
\caption{Best hyper-parameters for the \textbf{MIL} baseline}
\label{tab:best_mil}
\end{table*}

\begin{table*}[ht]
\centering
\footnotesize
\begin{tabular}{lllrrrrrr}
\toprule
\multirow{2}{*}{Encoder}
  & \multirow{2}{*}{Pooling}
  & \multirow{2}{*}{Label} &
\multicolumn{6}{c}{\textbf{Hyper-parameters}} \\
\cmidrule(lr){4-9}
\multicolumn{3}{c}{} &
LR & Batch & Epochs & Hdim & Actor LR & Early stopping \\
\midrule
0 & Attention & age    & 0.000001 & 128 & 800 & 8 & 0.001023 & 100 \\
0 & Attention & gender & 0.000001 & 128 & 800 & 8 & 0.000125 & 100 \\
0 & Max       & age    & 0.000001 & 128 & 800 & 8 & 0.000093 & 100 \\
0 & Max       & gender & 0.000001 & 128 & 800 & 8 & 0.000235 & 100 \\
0 & Mean      & age    & 0.000001 & 128 & 800 & 8 & 0.000295 & 100 \\
0 & Mean      & gender & 0.000001 & 128 & 800 & 8 & 0.000016 & 100 \\
1 & Attention & age    & 0.000001 & 128 & 800 & 8 & 0.007486 & 100 \\
1 & Attention & gender & 0.000001 & 128 & 800 & 8 & 0.000015 & 100 \\
1 & Max       & age    & 0.000001 & 128 & 800 & 8 & 0.003406 & 100 \\
1 & Max       & gender & 0.000001 & 128 & 800 & 8 & 0.001760 & 100 \\
1 & Mean      & age    & 0.000001 & 128 & 800 & 8 & 0.000010 & 100 \\
1 & Mean      & gender & 0.000001 & 128 & 800 & 8 & 0.000872 & 100 \\
2 & Attention & age    & 0.000001 & 128 & 800 & 8 & 0.000181 & 100 \\
2 & Attention & gender & 0.000001 & 128 & 800 & 8 & 0.000219 & 100 \\
2 & Max       & age    & 0.000001 & 128 & 800 & 8 & 0.001192 & 100 \\
2 & Max       & gender & 0.000001 & 128 & 800 & 8 & 0.000097 & 100 \\
2 & Mean      & age    & 0.000001 & 128 & 800 & 8 & 0.001494 & 100 \\
2 & Mean      & gender & 0.000001 & 128 & 800 & 8 & 0.000089 & 100 \\
\bottomrule
\end{tabular}
\caption{Best hyper-parameters for \textbf{RL-MIL}}
\label{tab:best_rlmil}
\end{table*}

\begin{table}[ht]
\centering
\resizebox{\textwidth}{!}{%
\begin{tabular}{lllrrrrrrrr}
\toprule
\multirow{2}{*}{Encoder}
  & \multirow{2}{*}{Pooling}
  & \multirow{2}{*}{Label} &
\multicolumn{8}{c}{\textbf{Hyper-parameters}} \\
\cmidrule(lr){4-11}
\multicolumn{3}{c}{} &
LR & Batch & Epochs & Hdim &
Actor LR & Encoder LR & Domain LR & Early stopping \\
\midrule
0 & Attention & age    & 0.000059 & 128 & 639 & 32 & 0.000016 & 0.000001 & 0.000222 & 100 \\
0 & Attention & gender & 0.000016 & 128 & 990 & 64 & 0.000053 & 0.000051 & 0.000016 & 100 \\
0 & Max       & age    & 0.000022 & 128 & 669 & 32 & 0.000022 & 0.000004 & 0.000010 & 100 \\
0 & Max       & gender & 0.000013 & 128 & 893 & 16 & 0.000010 & 0.000066 & 0.000172 & 100 \\
0 & Mean      & age    & 0.000080 & 128 & 567 & 64 & 0.000044 & 0.000003 & 0.000019 & 100 \\
0 & Mean      & gender & 0.000017 & 128 & 520 & 64 & 0.000091 & 0.000026 & 0.000026 & 100 \\
1 & Attention & age    & 0.000089 & 128 & 683 & 16 & 0.000027 & 0.000003 & 0.000649 & 100 \\
1 & Attention & gender & 0.000065 & 128 & 778 & 64 & 0.000072 & 0.000005 & 0.000205 & 100 \\
1 & Max       & age    & 0.000092 & 128 & 846 & 32 & 0.000010 & 0.000030 & 0.000533 & 100 \\
1 & Max       & gender & 0.000037 & 128 & 746 & 32 & 0.000099 & 0.000038 & 0.000017 & 100 \\
1 & Mean      & age    & 0.000070 & 128 & 800 & 32 & 0.000073 & 0.000029 & 0.000530 & 100 \\
1 & Mean      & gender & 0.000083 & 128 & 531 & 32 & 0.000023 & 0.000015 & 0.000584 & 100 \\
2 & Attention & age    & 0.000077 & 128 & 596 & 32 & 0.000021 & 0.000002 & 0.000031 & 100 \\
2 & Attention & gender & 0.000073 & 128 & 742 & 32 & 0.000013 & 0.000010 & 0.000034 & 100 \\
2 & Max       & age    & 0.000044 & 128 & 744 & 64 & 0.000040 & 0.000004 & 0.000745 & 100 \\
2 & Max       & gender & 0.000067 & 128 & 869 & 64 & 0.000014 & 0.000016 & 0.000016 & 100 \\
2 & Mean      & age    & 0.000039 & 128 & 580 & 16 & 0.000050 & 0.000016 & 0.000227 & 100 \\
2 & Mean      & gender & 0.000026 & 128 & 879 & 32 & 0.000085 & 0.000004 & 0.000031 & 100 \\
\bottomrule
\end{tabular}}
\caption{Best hyper-parameters for \textbf{RLMIL-DAT}}
\label{tab:best_rlmil_dat}
\end{table}
\FloatBarrier

\newpage
\section{Illustrative Learning Curves (Twitter - Seed 42)}
\label{sec:apx:second_appendix}

\begin{figure}[h!]
    \centering

    \begin{subfigure}[b]{0.3\textwidth}
        \centering
        \includegraphics[width=\textwidth]{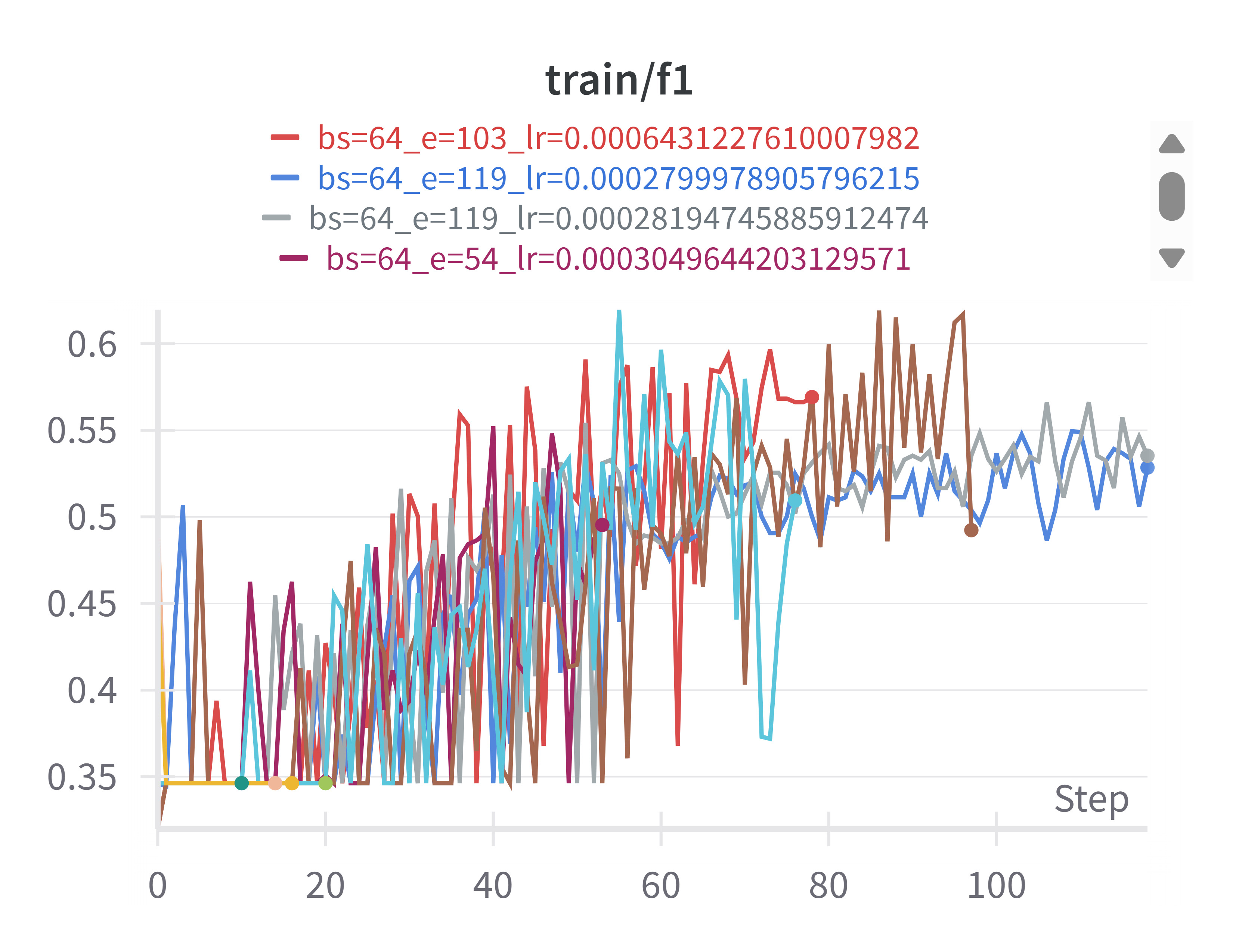}
        \caption{mil\_roberta\_MeanMLP\_gender}
        \label{fig:img1}
    \end{subfigure}
    \hfill
    \begin{subfigure}[b]{0.3\textwidth}
        \centering
        \includegraphics[width=\textwidth]{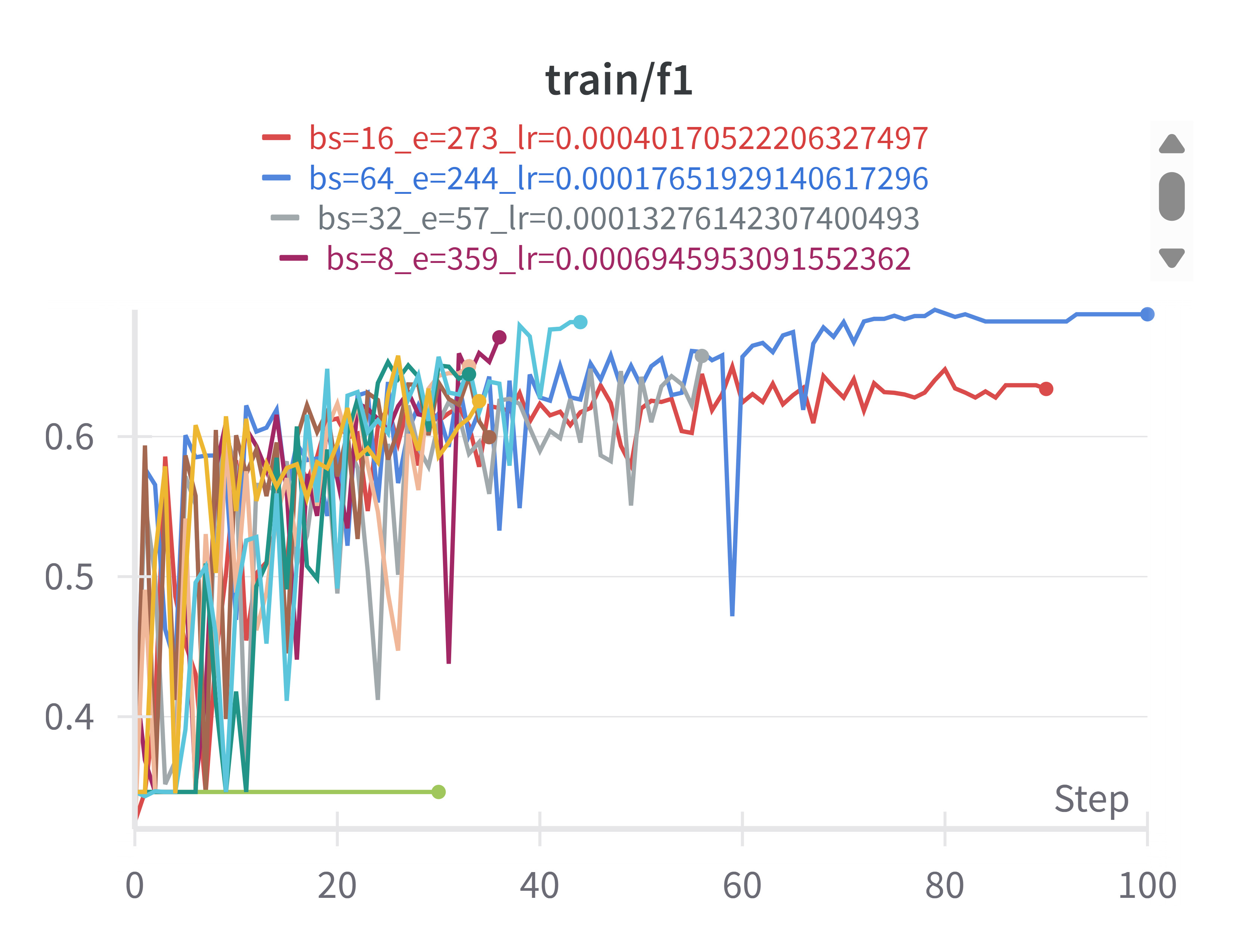}
        \caption{mil\_mbert\_MeanMLP\_gender}
        \label{fig:img2}
    \end{subfigure}
    \hfill
    \begin{subfigure}[b]{0.3\textwidth}
        \centering
        \includegraphics[width=\textwidth]{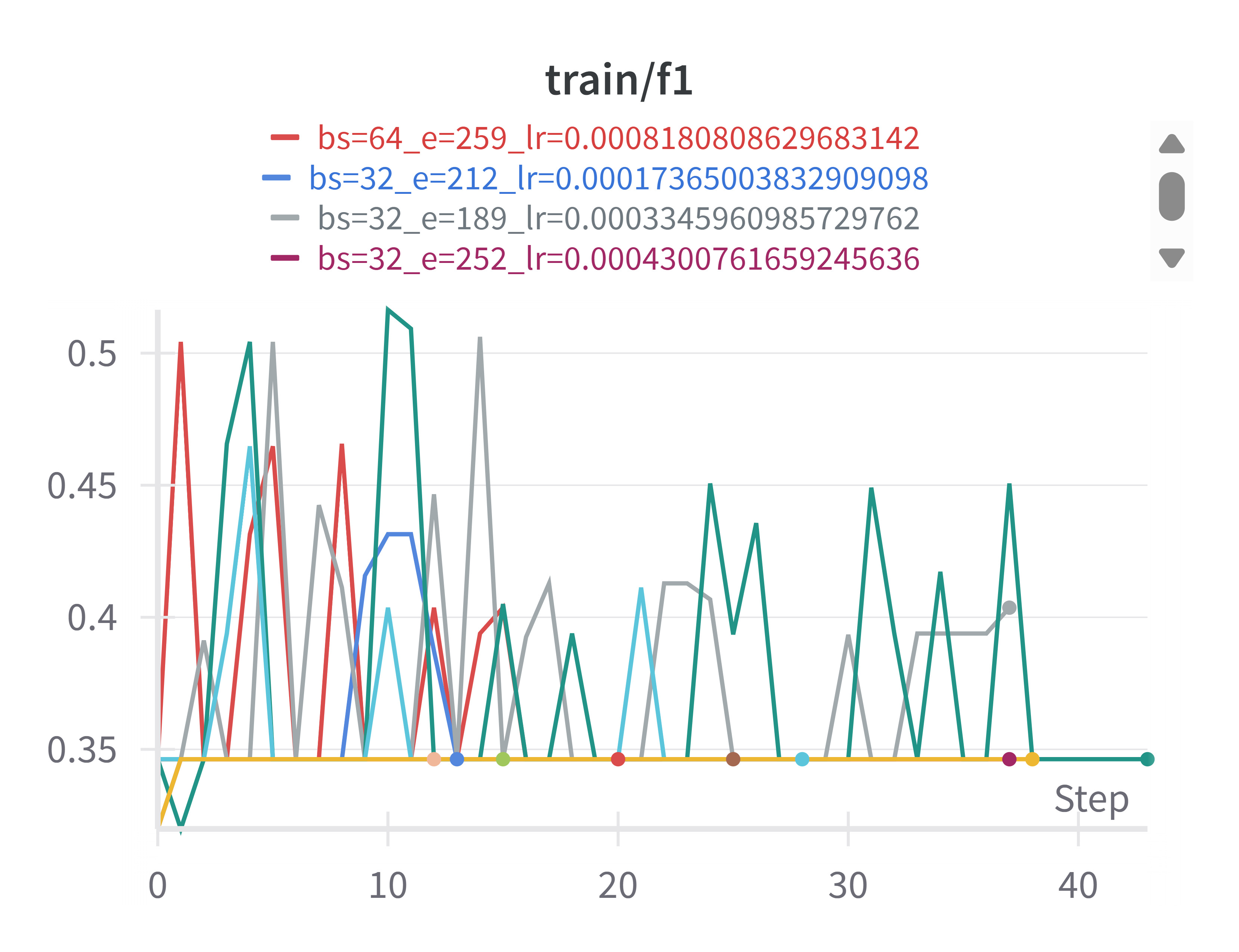}
        \caption{mil\_xlmr\_MeanMLP\_gender}
        \label{fig:img3}
    \end{subfigure}
    
    \vspace{0.5cm} 

    \begin{subfigure}[b]{0.3\textwidth}
        \centering
        \includegraphics[width=\textwidth]{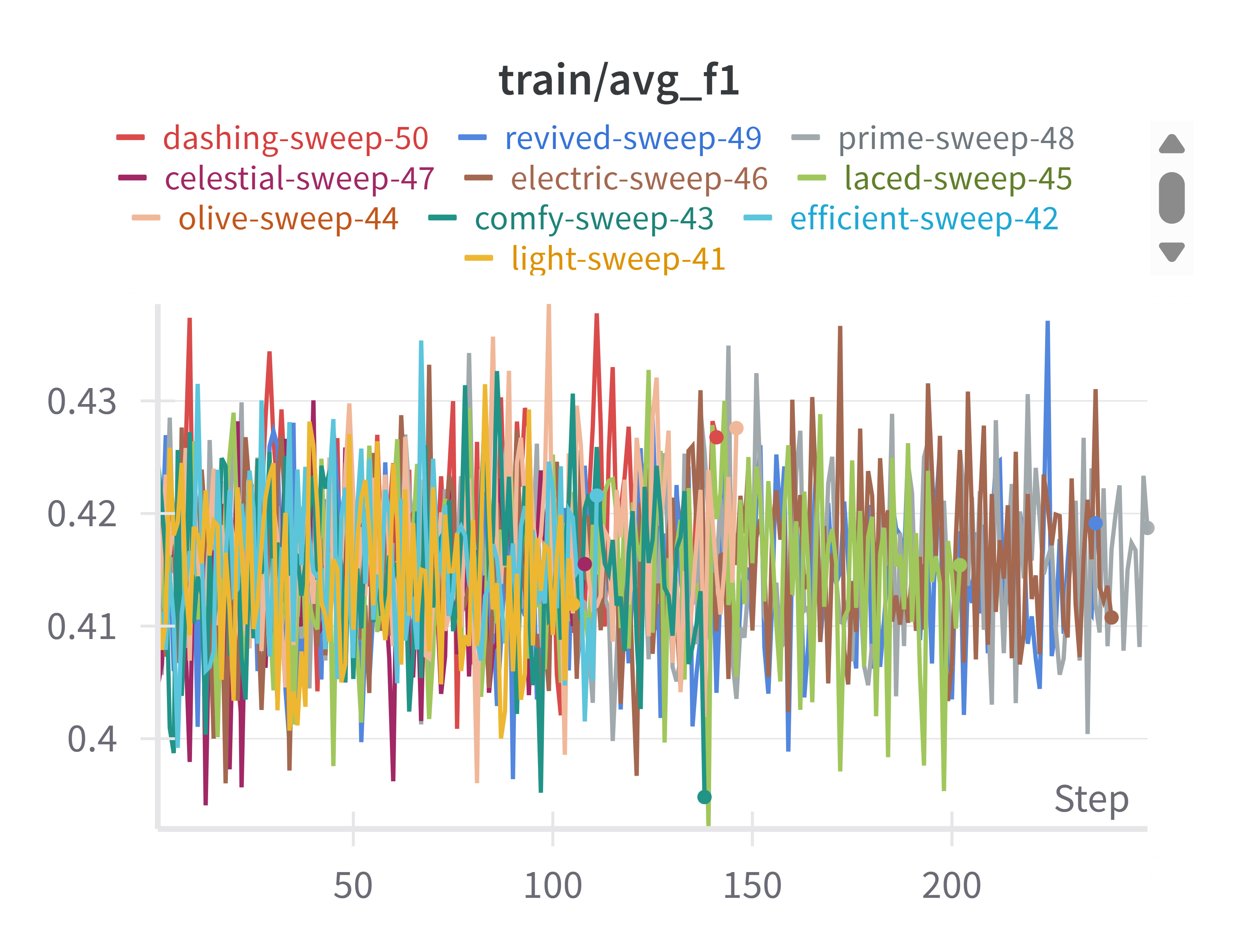}
        \caption{rlmil\_roberta\_MeanMLP\_gender}
        \label{fig:img4}
    \end{subfigure}
    \hfill
    \begin{subfigure}[b]{0.3\textwidth}
        \centering
        \includegraphics[width=\textwidth]{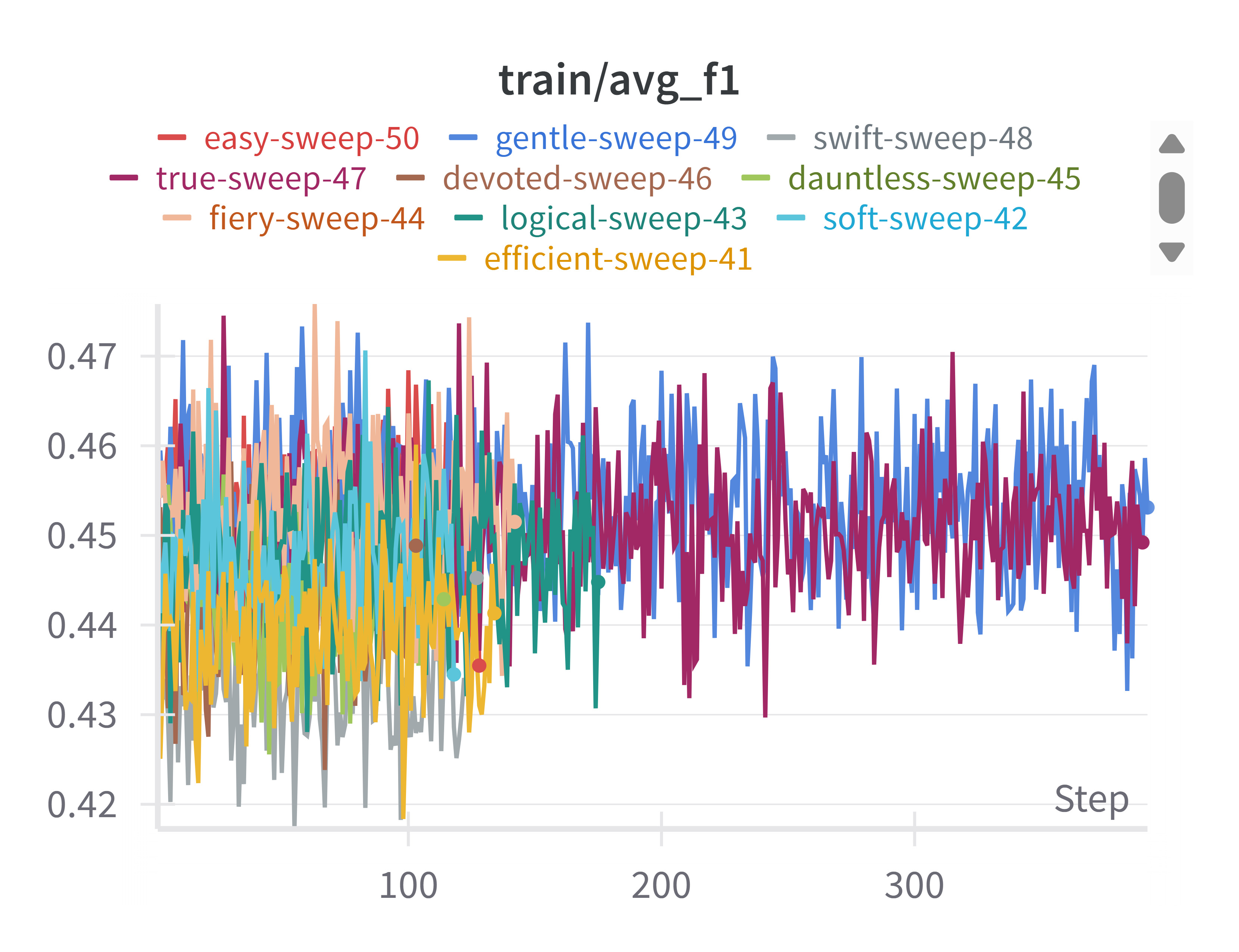}
        \caption{rlmil\_mbert\_MeanMLP\_gender}
        \label{fig:img5}
    \end{subfigure}
    \hfill
    \begin{subfigure}[b]{0.3\textwidth}
        \centering
        \includegraphics[width=\textwidth]{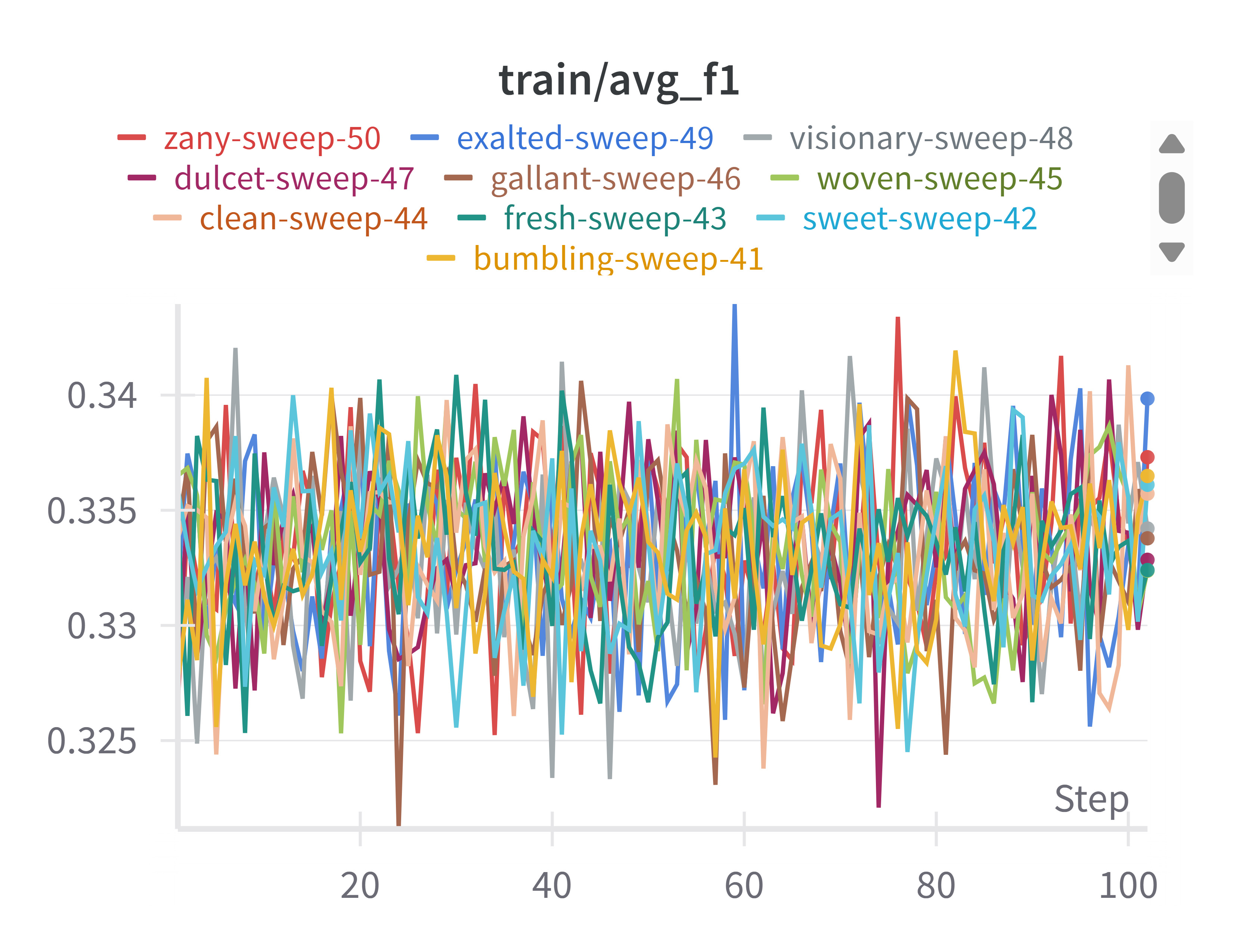}
        \caption{rlmil\_xlmr\_MeanMLP\_gender}
        \label{fig:img6}
    \end{subfigure}

    \vspace{0.5cm} 

    \begin{subfigure}[b]{0.3\textwidth}
        \centering
        \includegraphics[width=\textwidth]{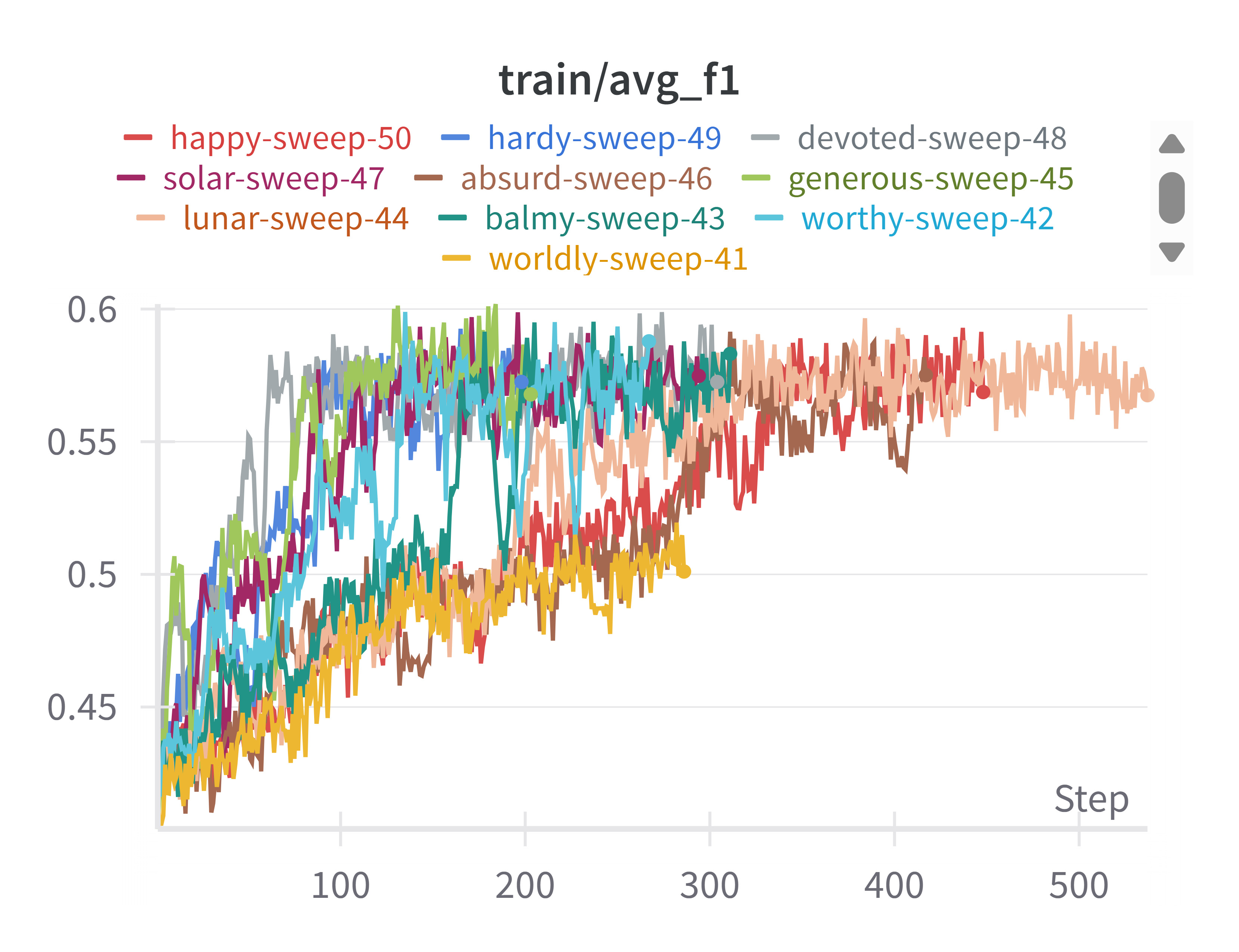}
        \caption{rlmil\_dat\_roberta\_MeanMLP\_gender}
        \label{fig:rlmil_dat_roberta_MeanMLP_gender}
    \end{subfigure}
    \hfill
    \begin{subfigure}[b]{0.3\textwidth}
        \centering
        \includegraphics[width=\textwidth]{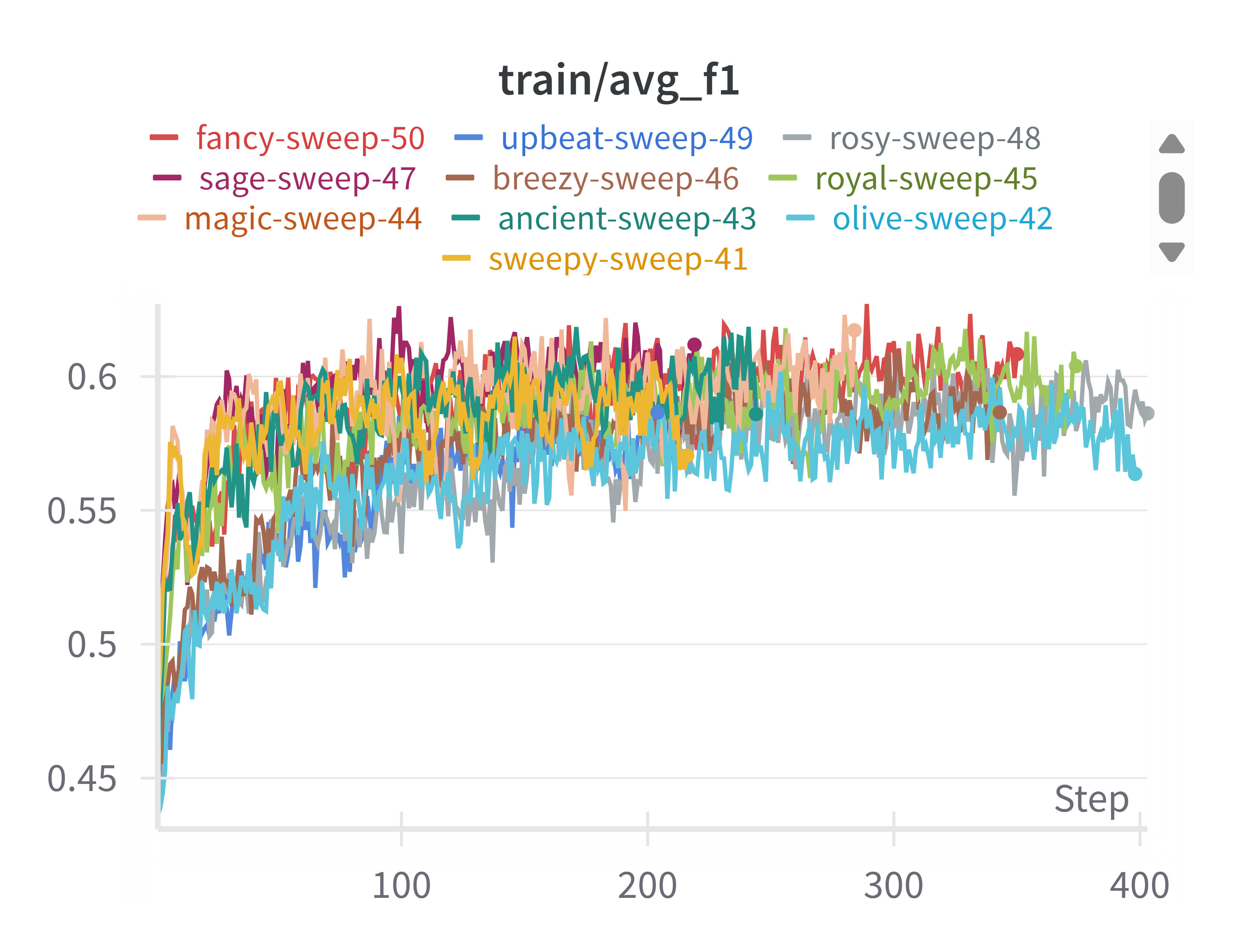}
        \caption{rlmil\_dat\_mbert\_MeanMLP\_gender}
        \label{fig:rlmil_dat_mbert_MeanMLP_gender}
    \end{subfigure}
    \hfill
    \begin{subfigure}[b]{0.3\textwidth}
        \centering
        \includegraphics[width=\textwidth]{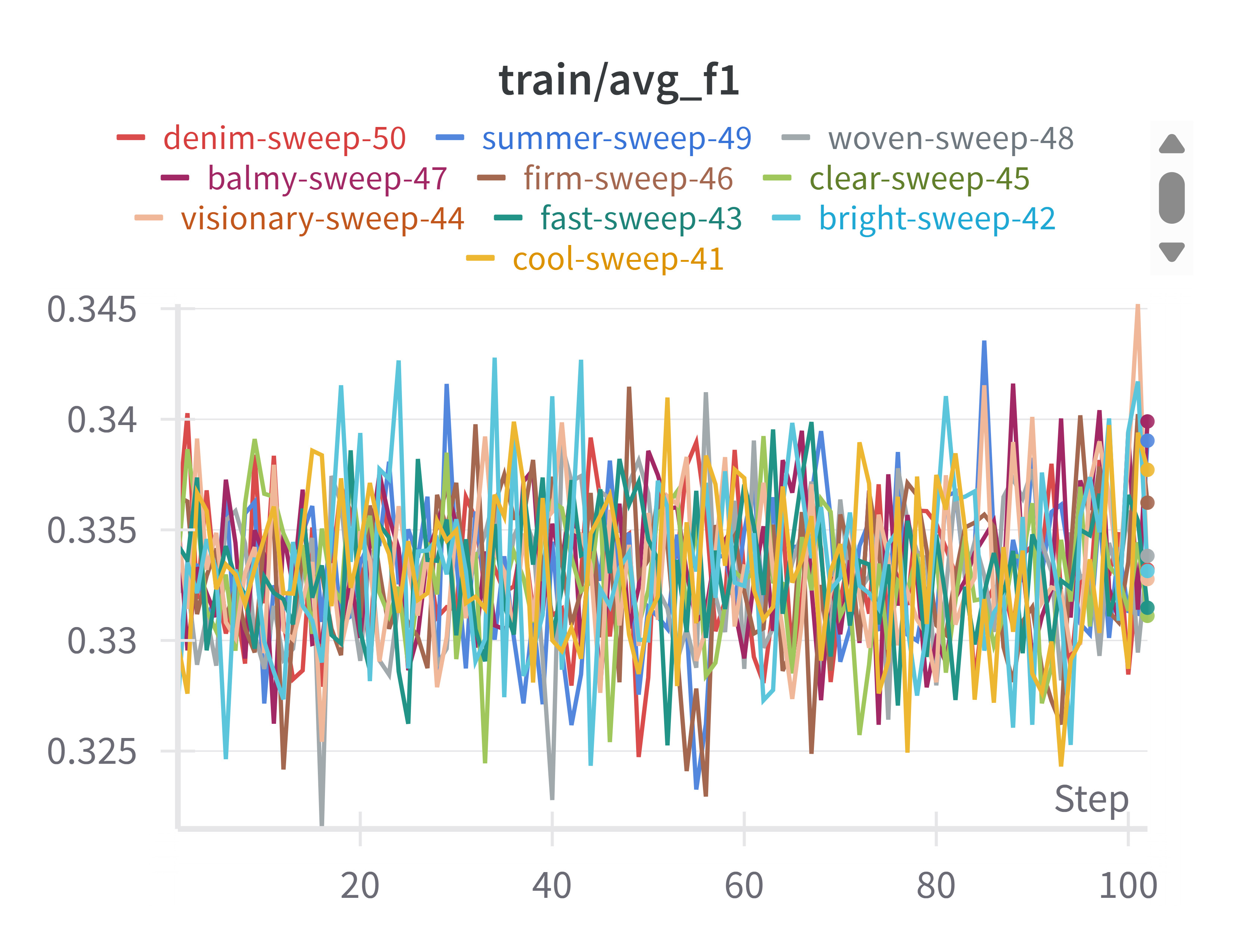}
        \caption{rlmil\_dat\_xlmr\_MeanMLP\_gender}
        \label{fig:rlmil_dat_xlmr_MeanMLP_gender}
    \end{subfigure}

    \caption{Training F1 Score Learning Curves for MeanMLP on Gender Prediction (Twitter - Seed 42). This figure displays the training F1 learning curves from Weights \& Biases for various model configurations predicting the gender attribute using the MeanMLP pooling head.}
    \label{fig:main_grid}
\end{figure}


\begin{figure}[h!]
    \centering

    \begin{subfigure}[b]{0.3\textwidth}
        \centering
        \includegraphics[width=\textwidth]{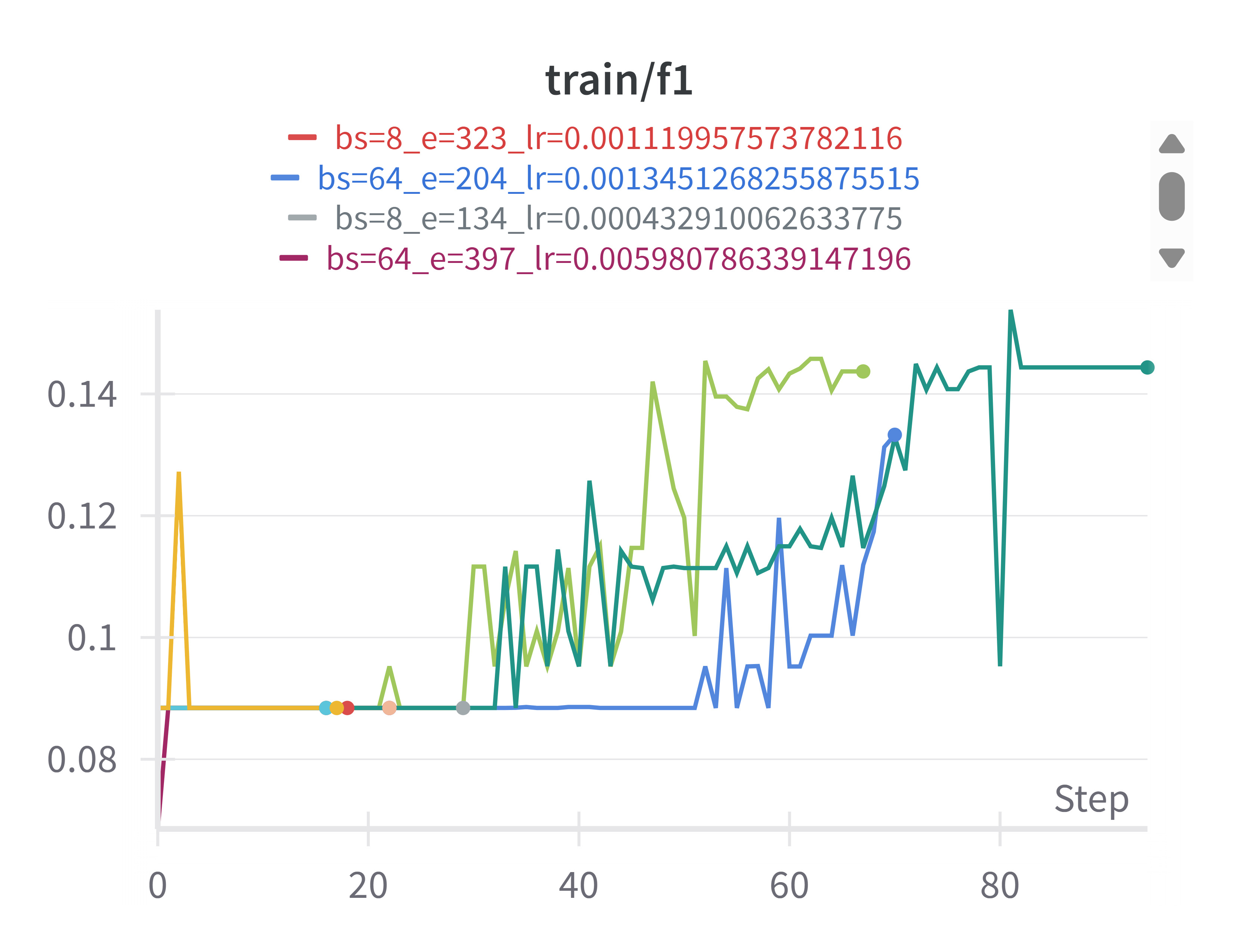}
        \caption{mil\_roberta\_MeanMLP\_age}
        \label{fig:img1}
    \end{subfigure}
    \hfill
    \begin{subfigure}[b]{0.3\textwidth}
        \centering
        \includegraphics[width=\textwidth]{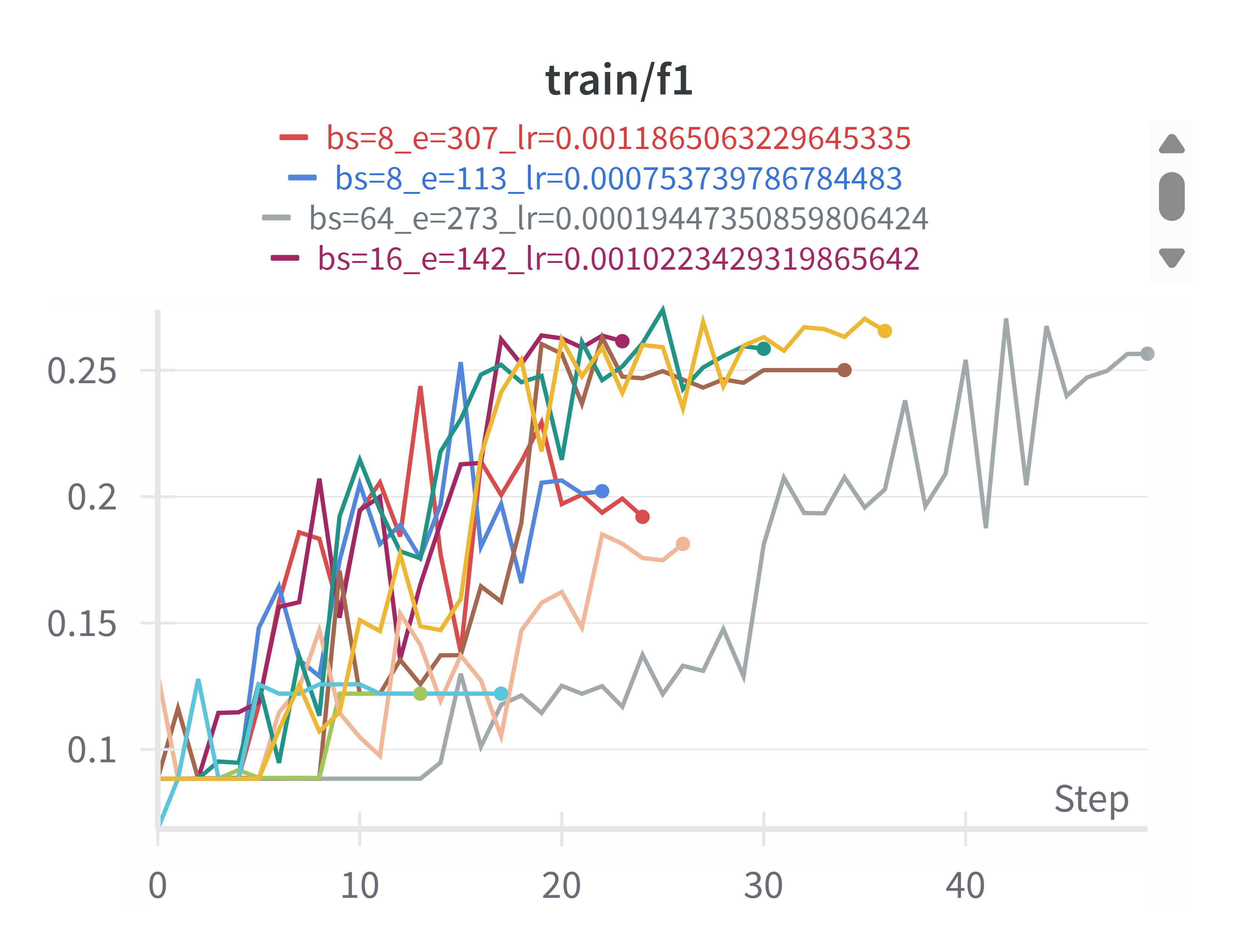}
        \caption{mil\_mbert\_MeanMLP\_age}
        \label{fig:img2}
    \end{subfigure}
    \hfill
    \begin{subfigure}[b]{0.3\textwidth}
        \centering
        \includegraphics[width=\textwidth]{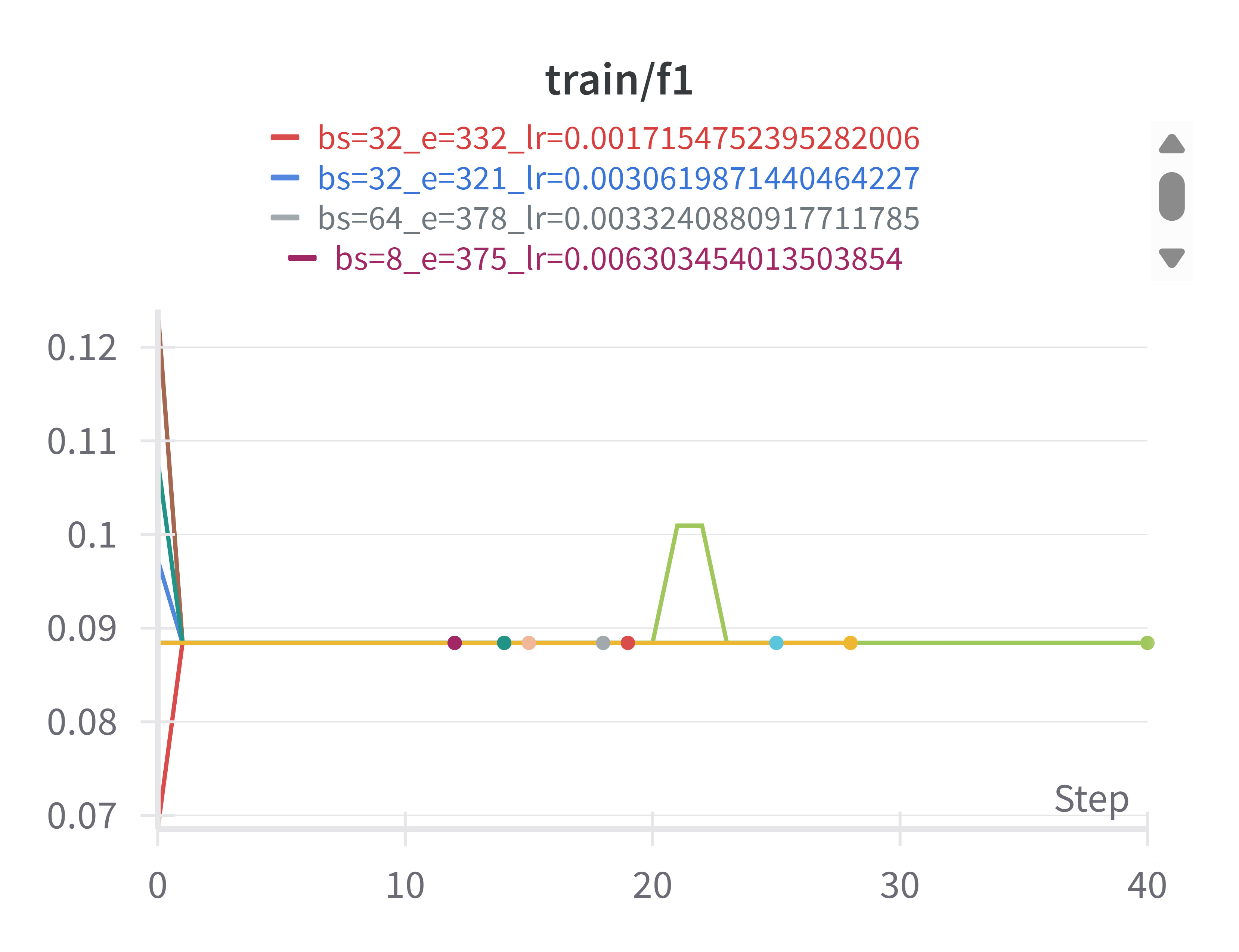}
        \caption{mil\_xlmr\_MeanMLP\_age}
        \label{fig:img3}
    \end{subfigure}
    
    \vspace{0.5cm} 

    \begin{subfigure}[b]{0.3\textwidth}
        \centering
        \includegraphics[width=\textwidth]{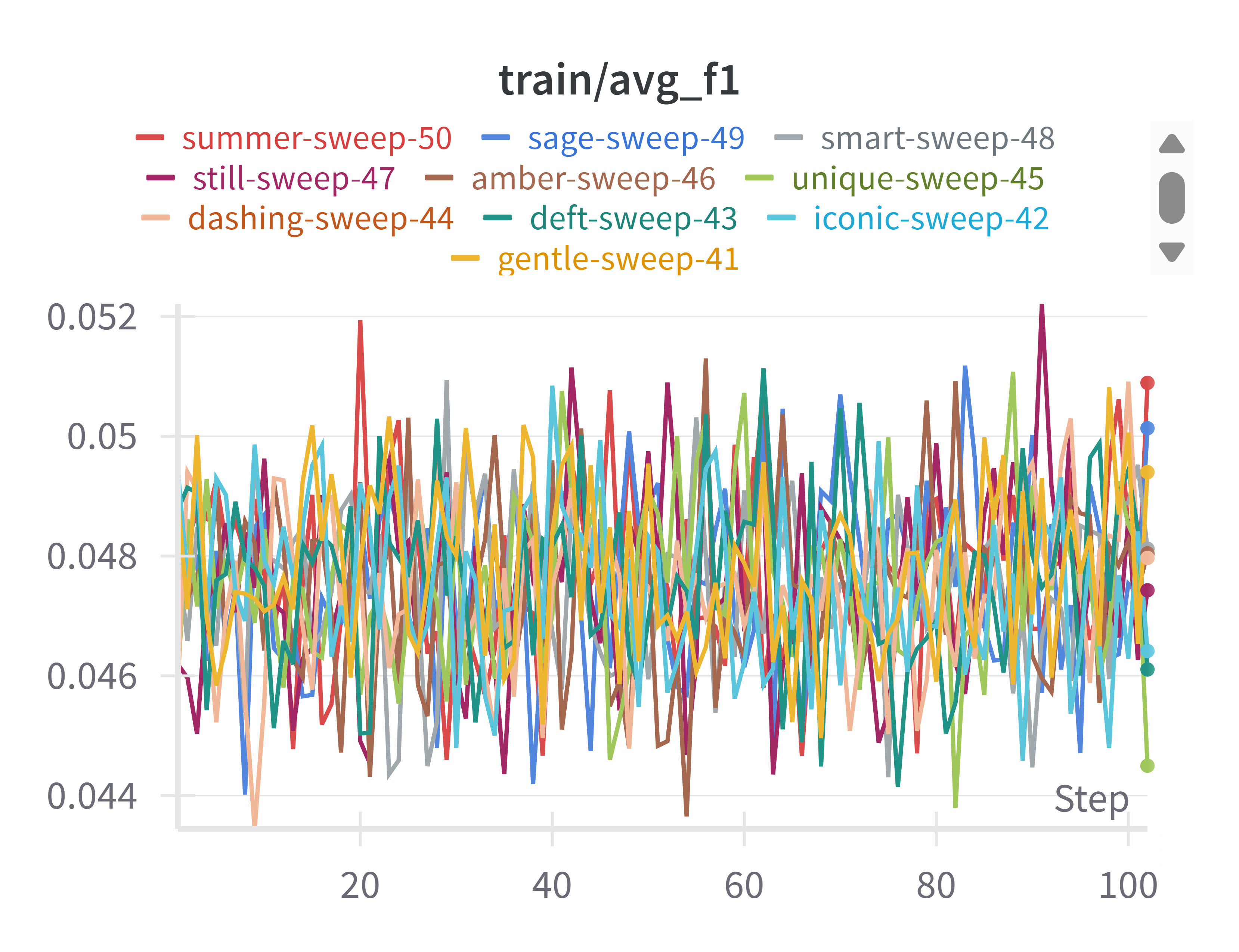}
        \caption{rlmil\_roberta\_MeanMLP\_age}
        \label{fig:img4}
    \end{subfigure}
    \hfill
    \begin{subfigure}[b]{0.3\textwidth}
        \centering
        \includegraphics[width=\textwidth]{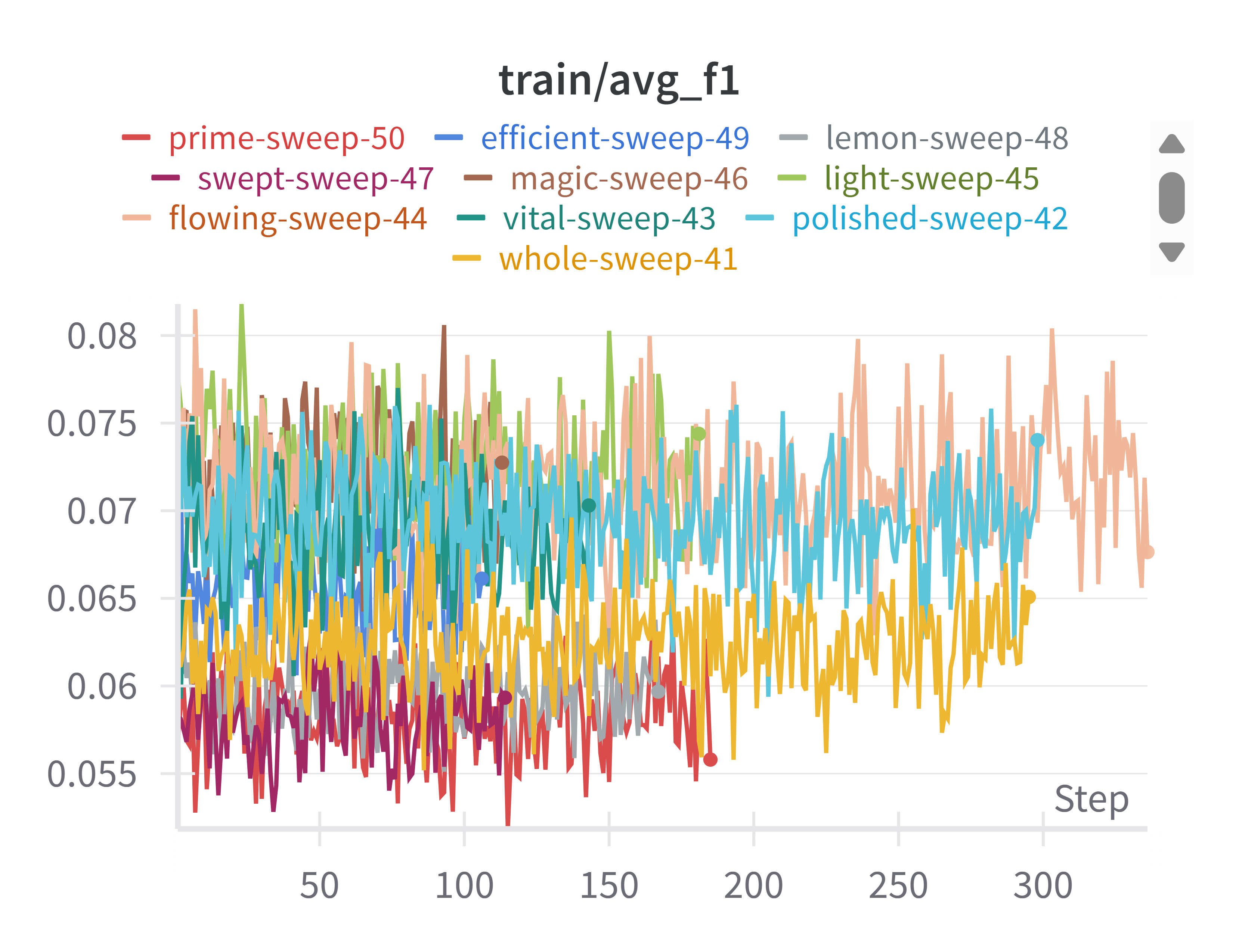}
        \caption{rlmil\_mbert\_MeanMLP\_age}
        \label{fig:img5}
    \end{subfigure}
    \hfill
    \begin{subfigure}[b]{0.3\textwidth}
        \centering
        \includegraphics[width=\textwidth]{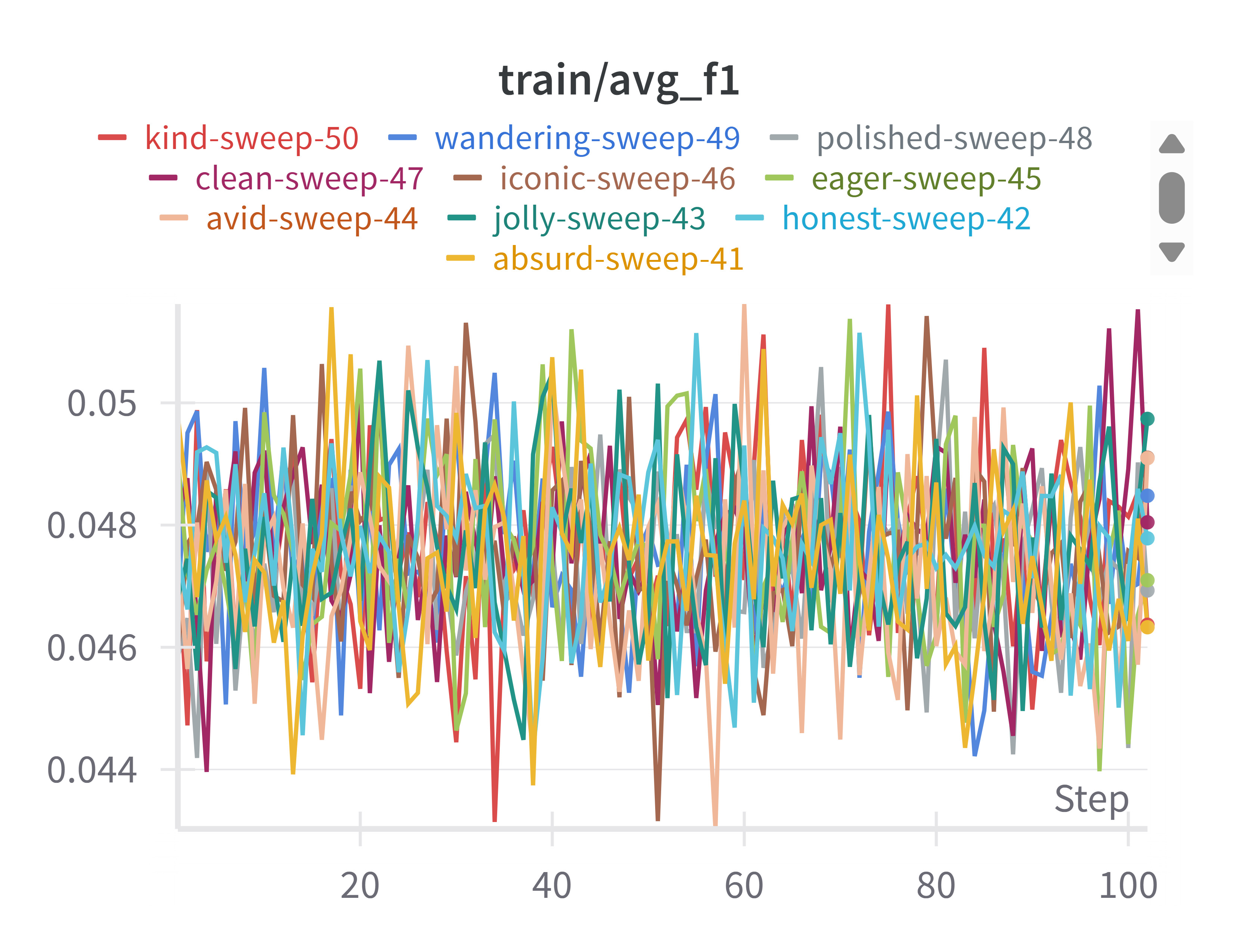}
        \caption{rlmil\_xlmr\_MeanMLP\_age}
        \label{fig:img6}
    \end{subfigure}

    \vspace{0.5cm} 

    \begin{subfigure}[b]{0.3\textwidth}
        \centering
        \includegraphics[width=\textwidth]{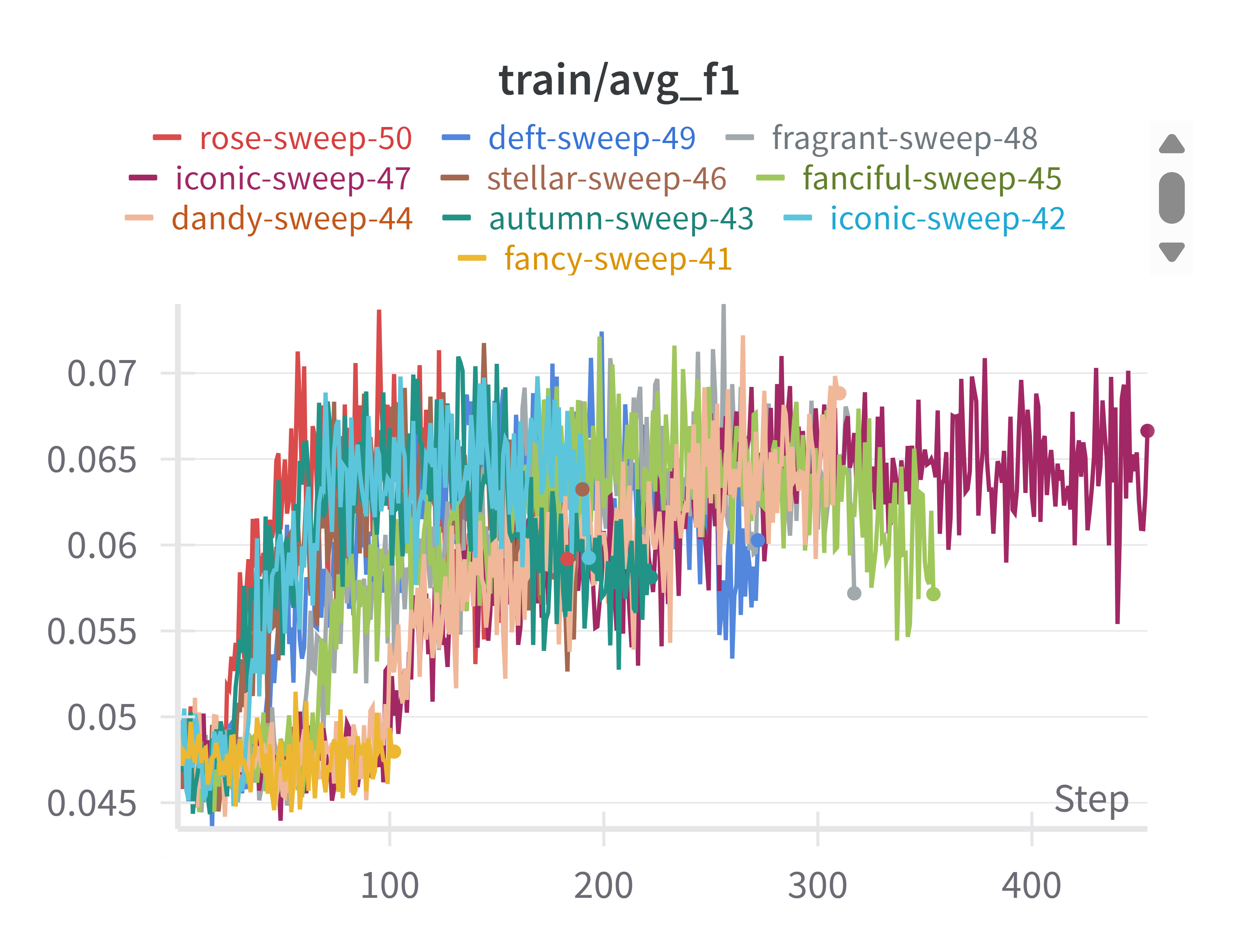}
        \caption{rlmil\_dat\_roberta\_MeanMLP\_age}
        \label{fig:img7}
    \end{subfigure}
    \hfill
    \begin{subfigure}[b]{0.3\textwidth}
        \centering
        \includegraphics[width=\textwidth]{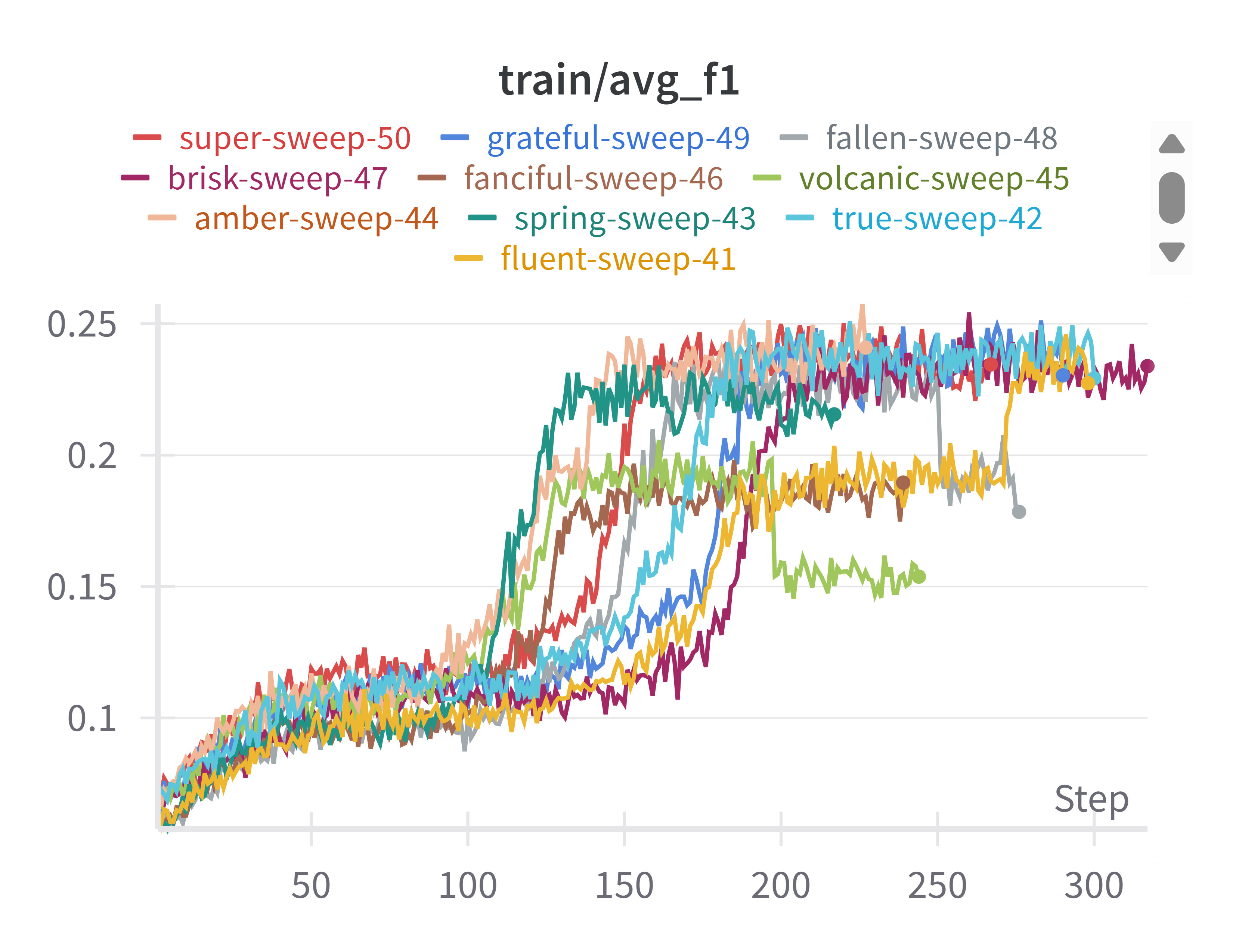}
        \caption{rlmil\_dat\_mbert\_MeanMLP\_age}
        \label{fig:img8}
    \end{subfigure}
    \hfill
    \begin{subfigure}[b]{0.3\textwidth}
        \centering
        \includegraphics[width=\textwidth]{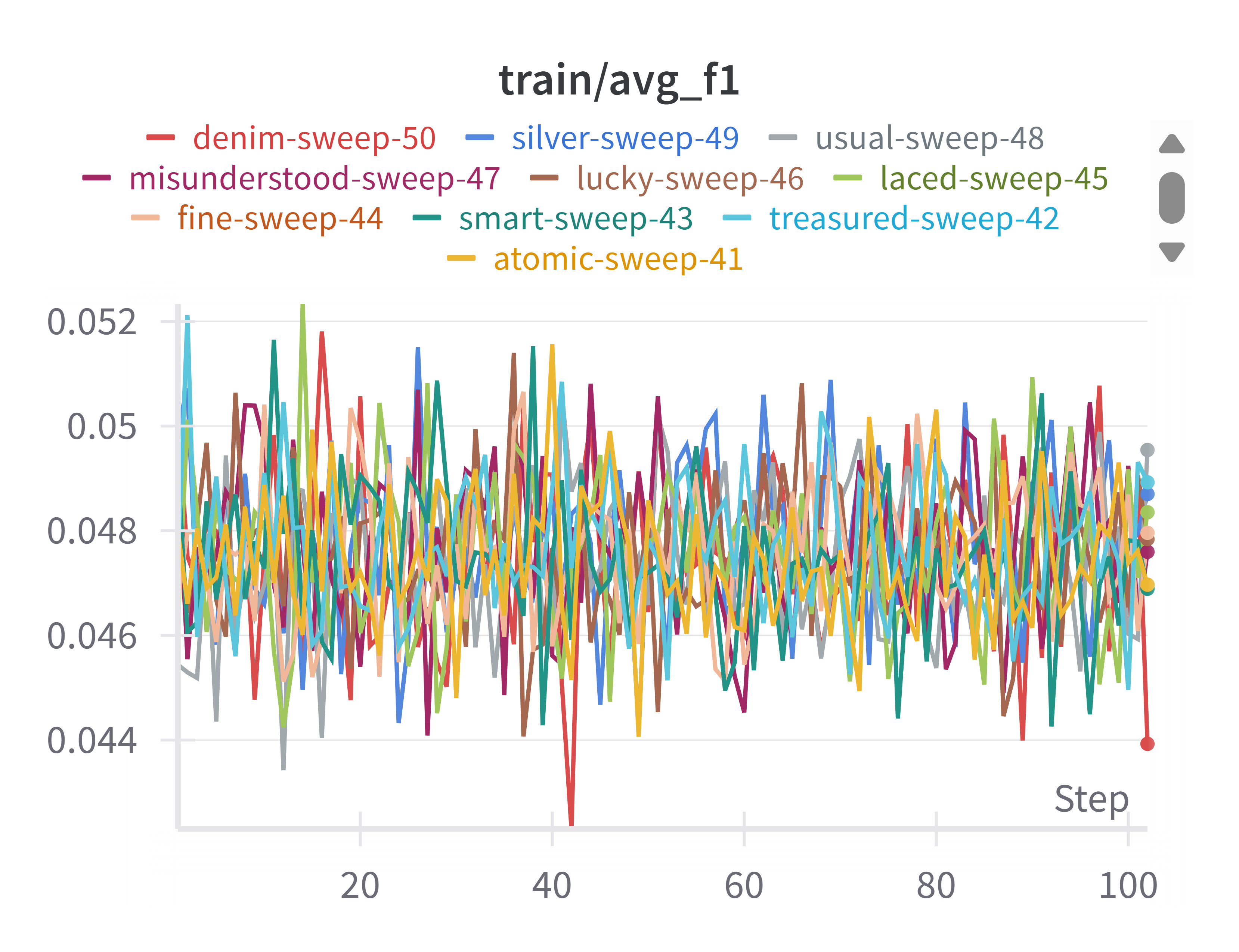}
        \caption{rlmil\_dat\_xlmr\_MeanMLP\_age}
        \label{fig:img9}
    \end{subfigure}

    \caption{Training F1 Score Learning Curves for MeanMLP on Age Prediction (Twitter - Seed 42). This figure displays the training F1 learning curves from Weights \& Biases for various model configurations predicting the age attribute using the MeanMLP pooling head. }
    \label{fig:main_grid}
\end{figure}




\begin{figure}[h!]
    \centering

    \begin{subfigure}[b]{0.3\textwidth}
        \centering
        \includegraphics[width=\textwidth]{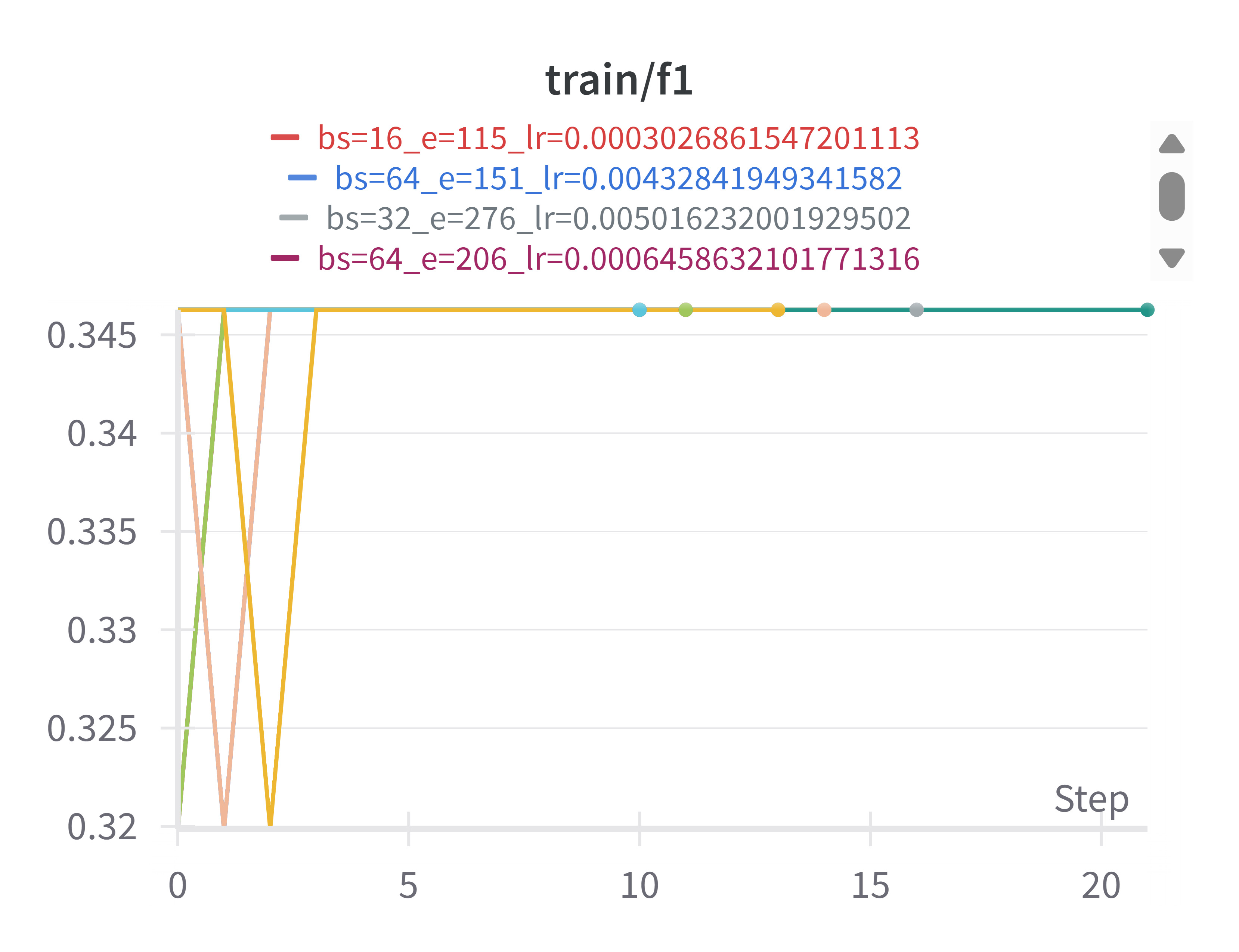}
        \caption{mil\_roberta\_MaxMLP\_gender}
        \label{fig:img1}
    \end{subfigure}
    \hfill
    \begin{subfigure}[b]{0.3\textwidth}
        \centering
        \includegraphics[width=\textwidth]{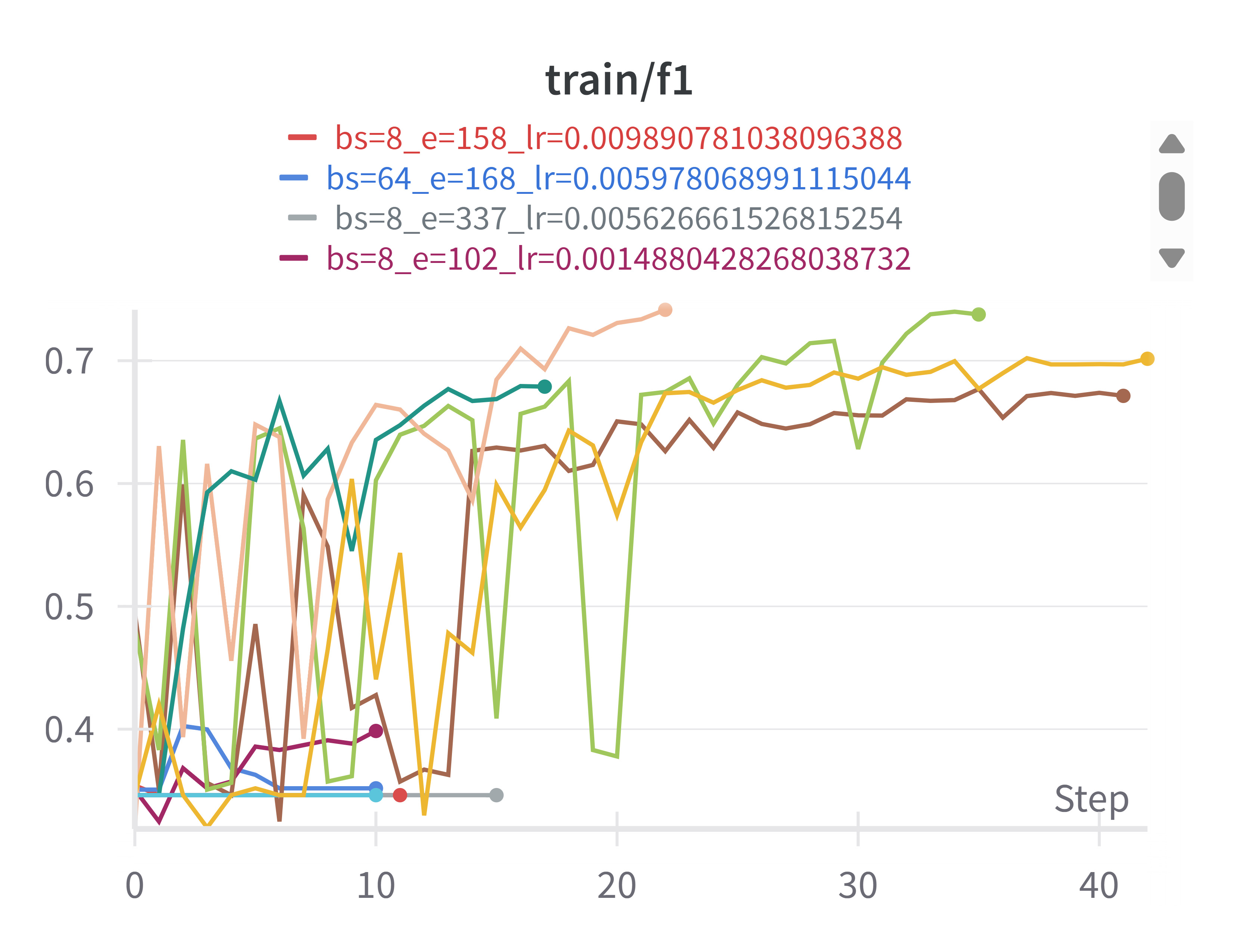}
        \caption{mil\_mbert\_MaxMLP\_gender}
        \label{fig:img2}
    \end{subfigure}
    \hfill
    \begin{subfigure}[b]{0.3\textwidth}
        \centering
        \includegraphics[width=\textwidth]{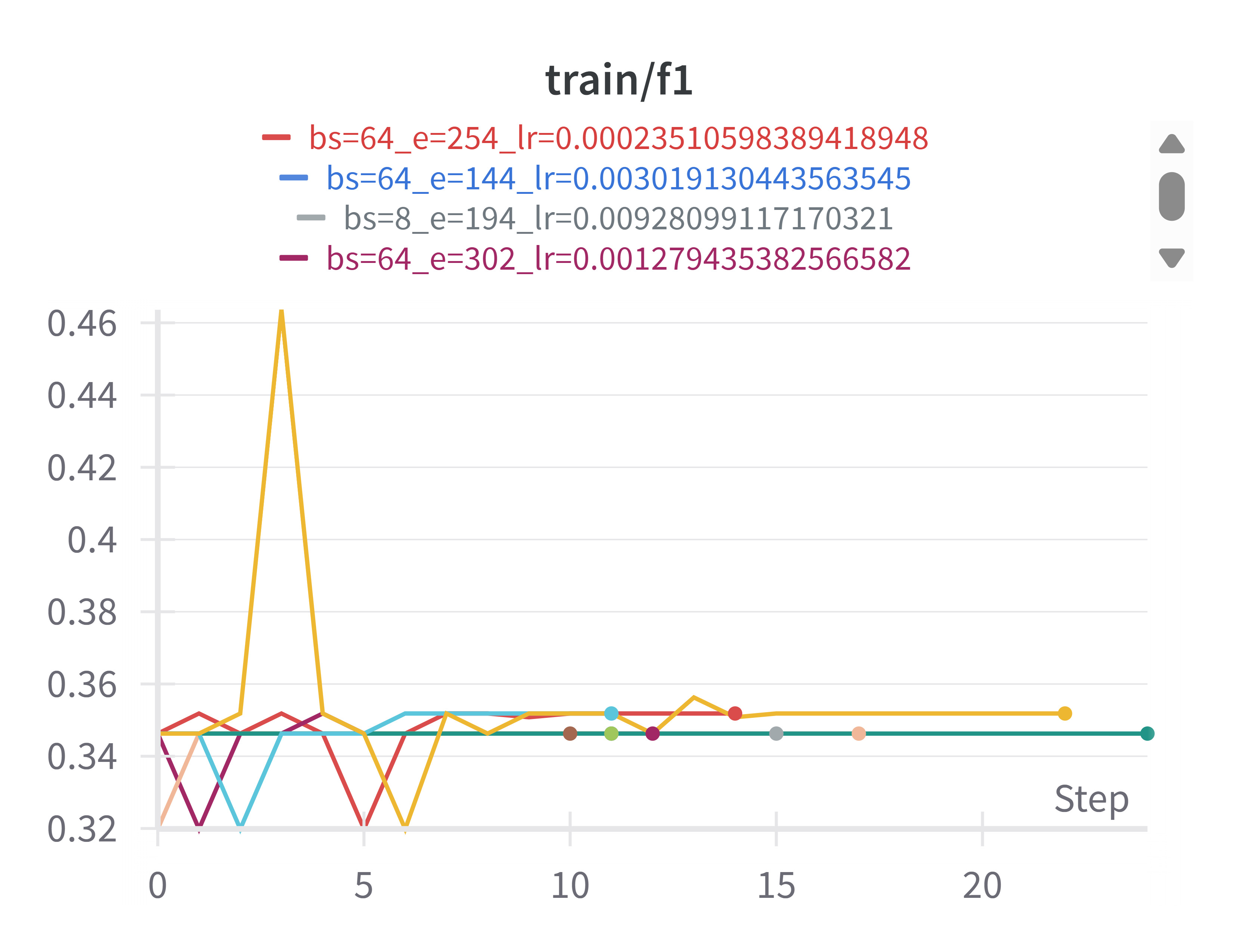}
        \caption{mil\_xlmr\_MaxMLP\_gender}
        \label{fig:img3}
    \end{subfigure}
    
    \vspace{0.5cm} 

    \begin{subfigure}[b]{0.3\textwidth}
        \centering
        \includegraphics[width=\textwidth]{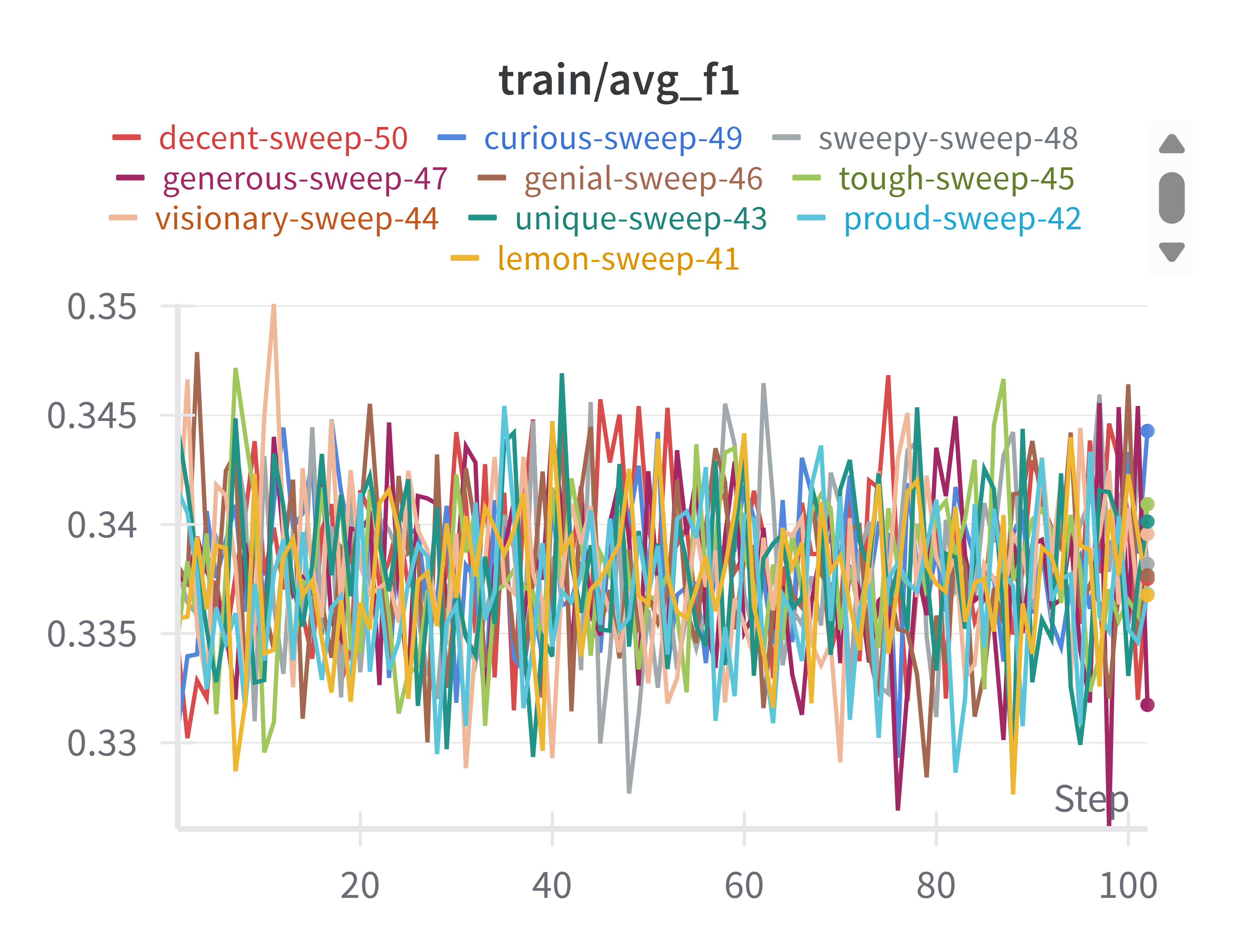}
        \caption{rlmil\_roberta\_MaxMLP\_gender}
        \label{fig:img4}
    \end{subfigure}
    \hfill
    \begin{subfigure}[b]{0.3\textwidth}
        \centering
        \includegraphics[width=\textwidth]{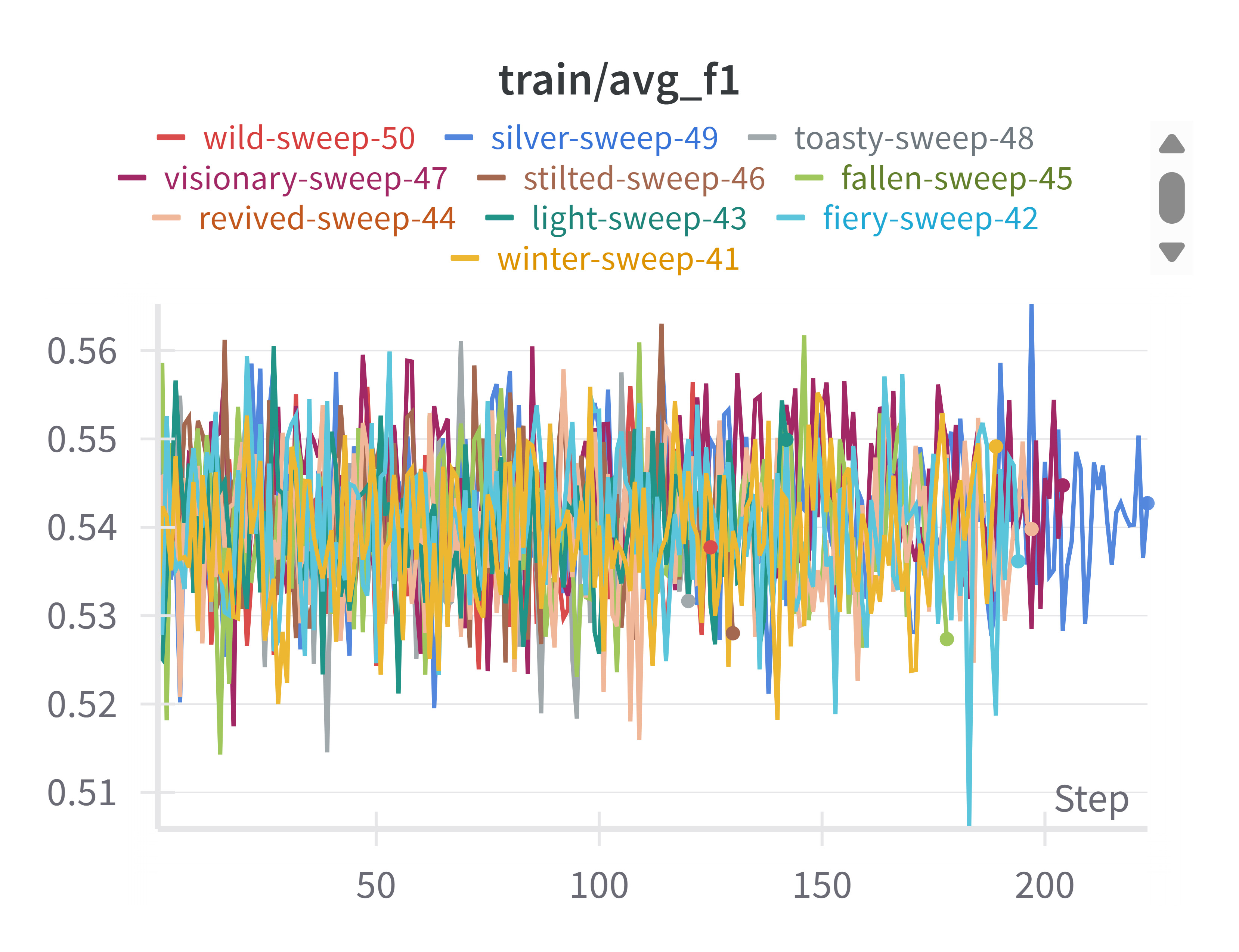}
        \caption{rlmil\_mbert\_MaxMLP\_gender}
        \label{fig:img5}
    \end{subfigure}
    \hfill
    \begin{subfigure}[b]{0.3\textwidth}
        \centering
        \includegraphics[width=\textwidth]{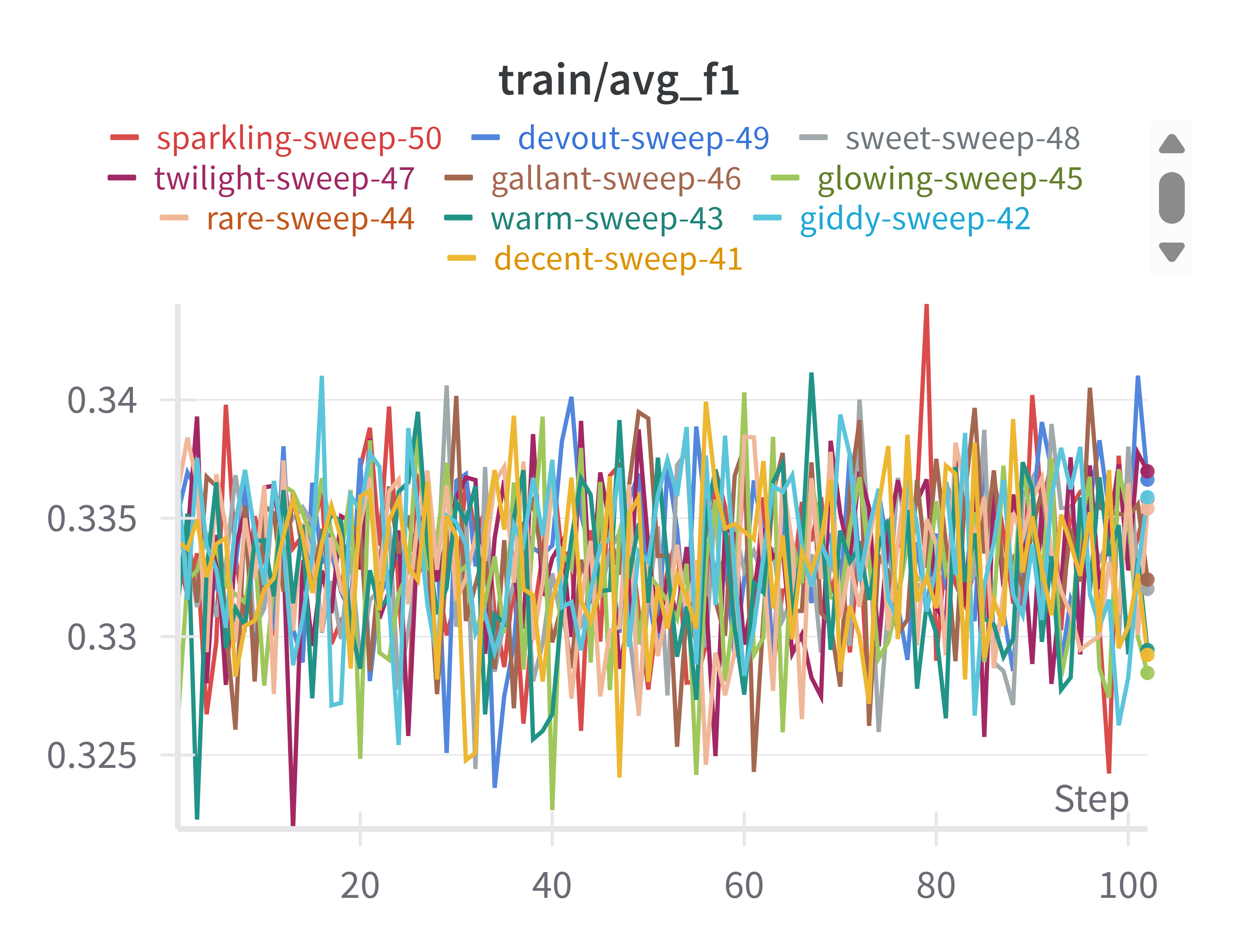}
        \caption{rlmil\_xlmr\_MaxMLP\_gender}
        \label{fig:img6}
    \end{subfigure}

    \vspace{0.5cm} 

    \begin{subfigure}[b]{0.3\textwidth}
        \centering
        \includegraphics[width=\textwidth]{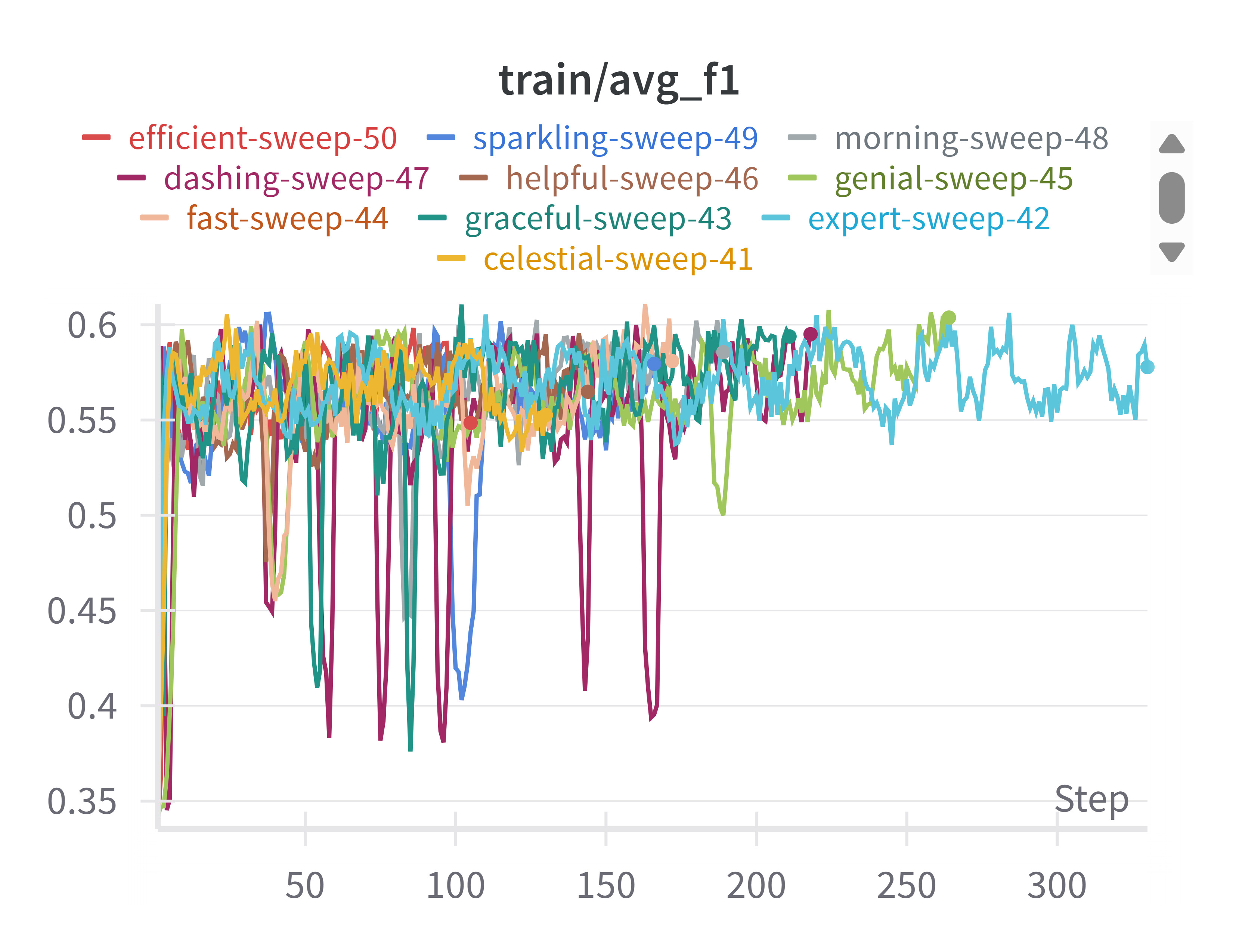}
        \caption{rlmil\_dat\_roberta\_MaxMLP\_gender}
        \label{fig:rlmil_dat_roberta_MaxMLP_gender}
    \end{subfigure}
    \hfill
    \begin{subfigure}[b]{0.3\textwidth}
        \centering
        \includegraphics[width=\textwidth]{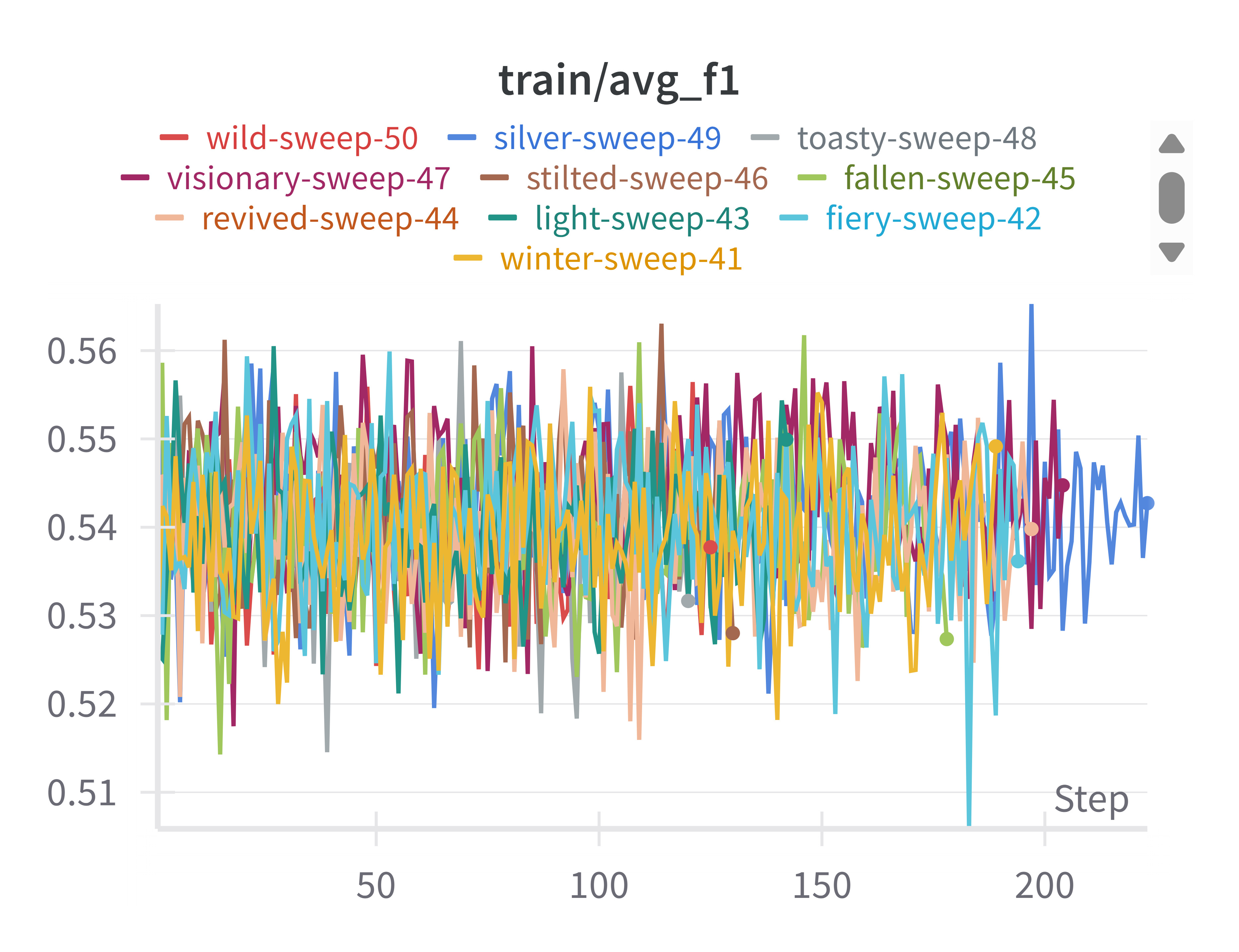}
        \caption{rlmil\_dat\_mbert\_MaxMLP\_gender}
        \label{fig:rlmil_dat_mbert_MaxMLP_gender}
    \end{subfigure}
    \hfill
    \begin{subfigure}[b]{0.3\textwidth}
        \centering
        \includegraphics[width=\textwidth]{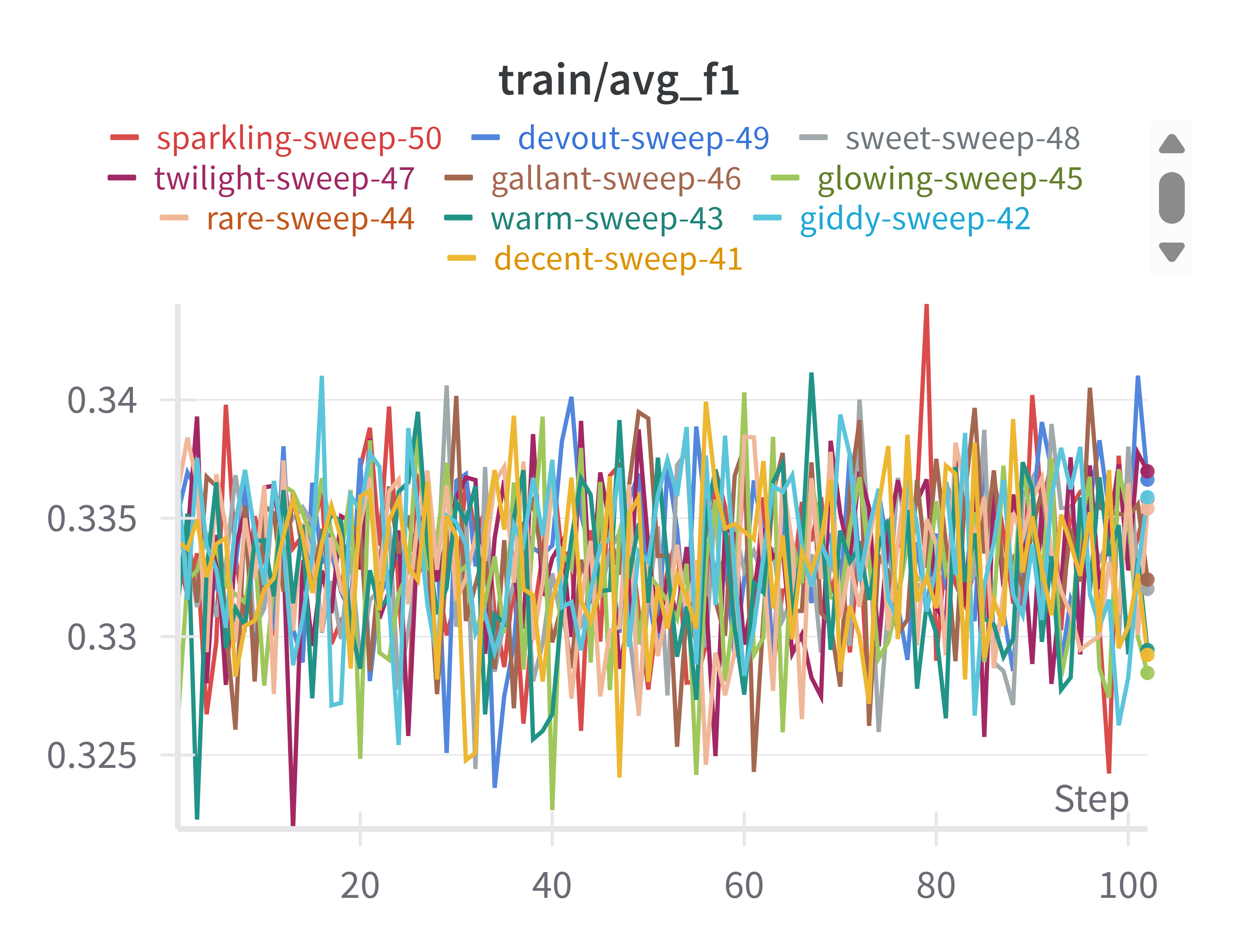}
        \caption{rlmil\_dat\_xlmr\_MaxMLP\_gender}
        \label{fig:rlmil_dat_xlmr_MaxMLP_gender}
    \end{subfigure}

    \caption{Training F1 Score Learning Curves for MaxMLP on Gender Prediction (Twitter - Seed 42). This figure displays the training F1 learning curves from Weights \& Biases for various model configurations predicting the gender attribute using the MaxMLP pooling head.}
    \label{fig:main_grid}
\end{figure}


\begin{figure}[h!]
    \centering

    \begin{subfigure}[b]{0.3\textwidth}
        \centering
        \includegraphics[width=\textwidth]{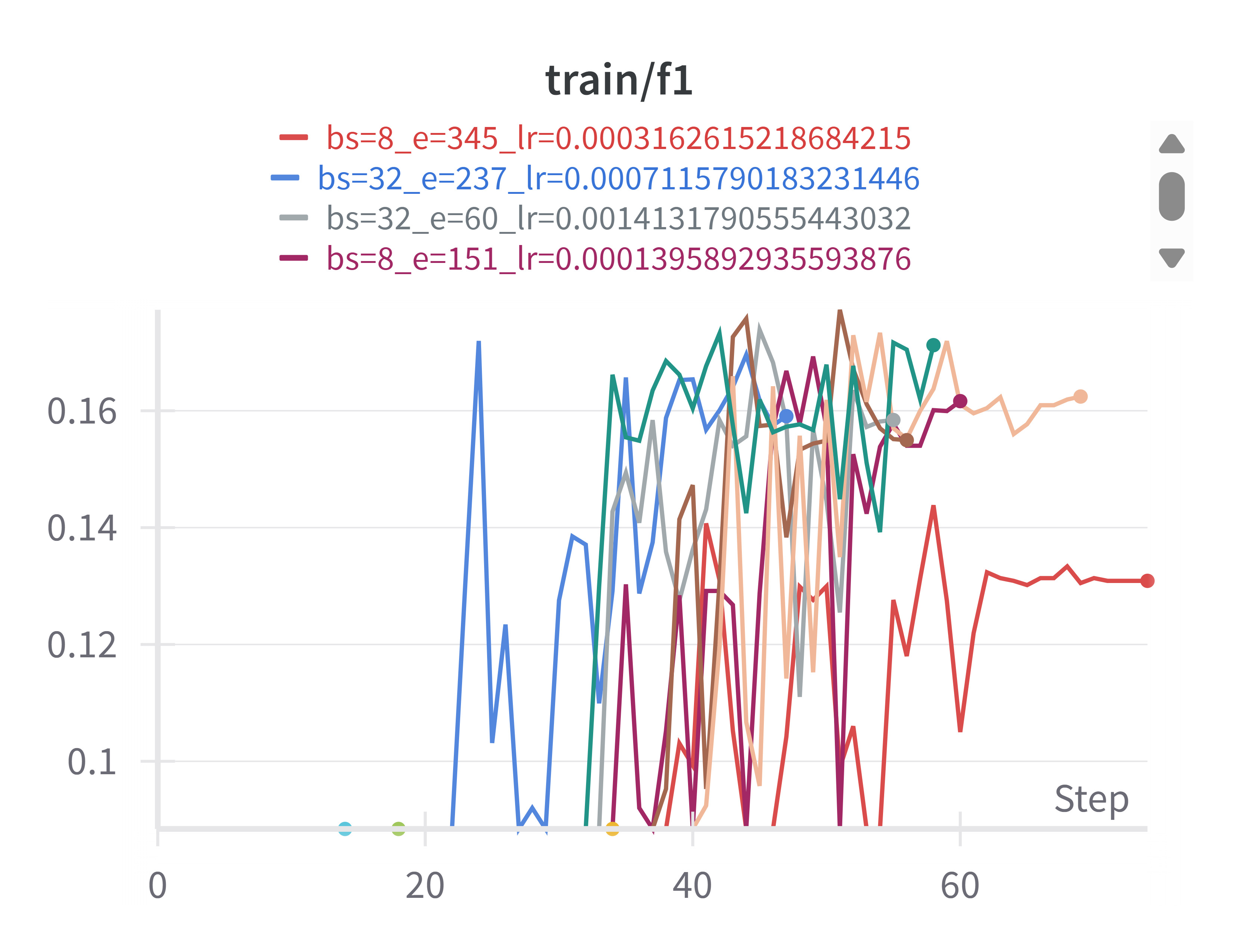}
        \caption{mil\_roberta\_MaxMLP\_age}
        \label{fig:img1}
    \end{subfigure}
    \hfill
    \begin{subfigure}[b]{0.3\textwidth}
        \centering
        \includegraphics[width=\textwidth]{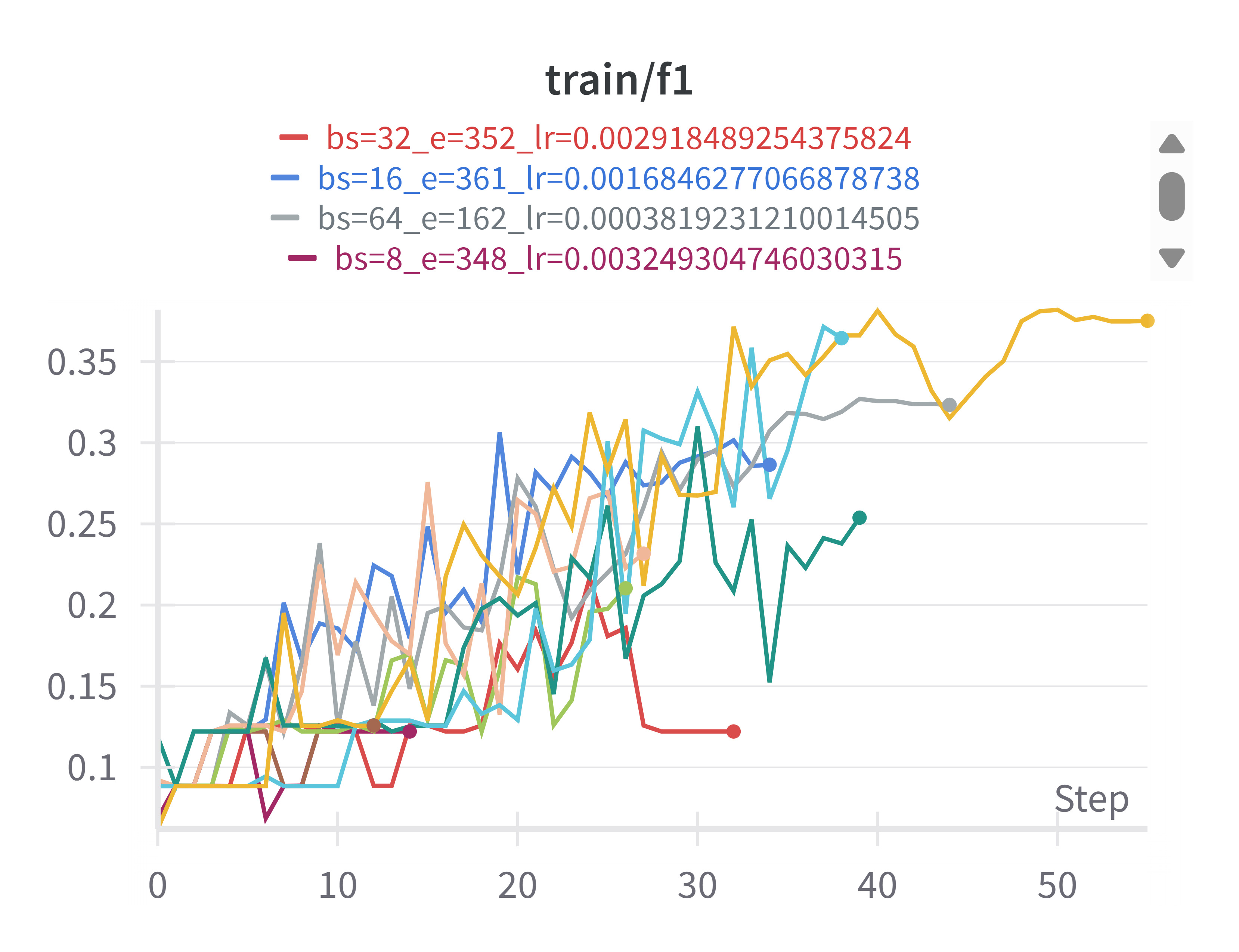}
        \caption{mil\_mbert\_MaxMLP\_age}
        \label{fig:img2}
    \end{subfigure}
    \hfill
    \begin{subfigure}[b]{0.3\textwidth}
        \centering
        \includegraphics[width=\textwidth]{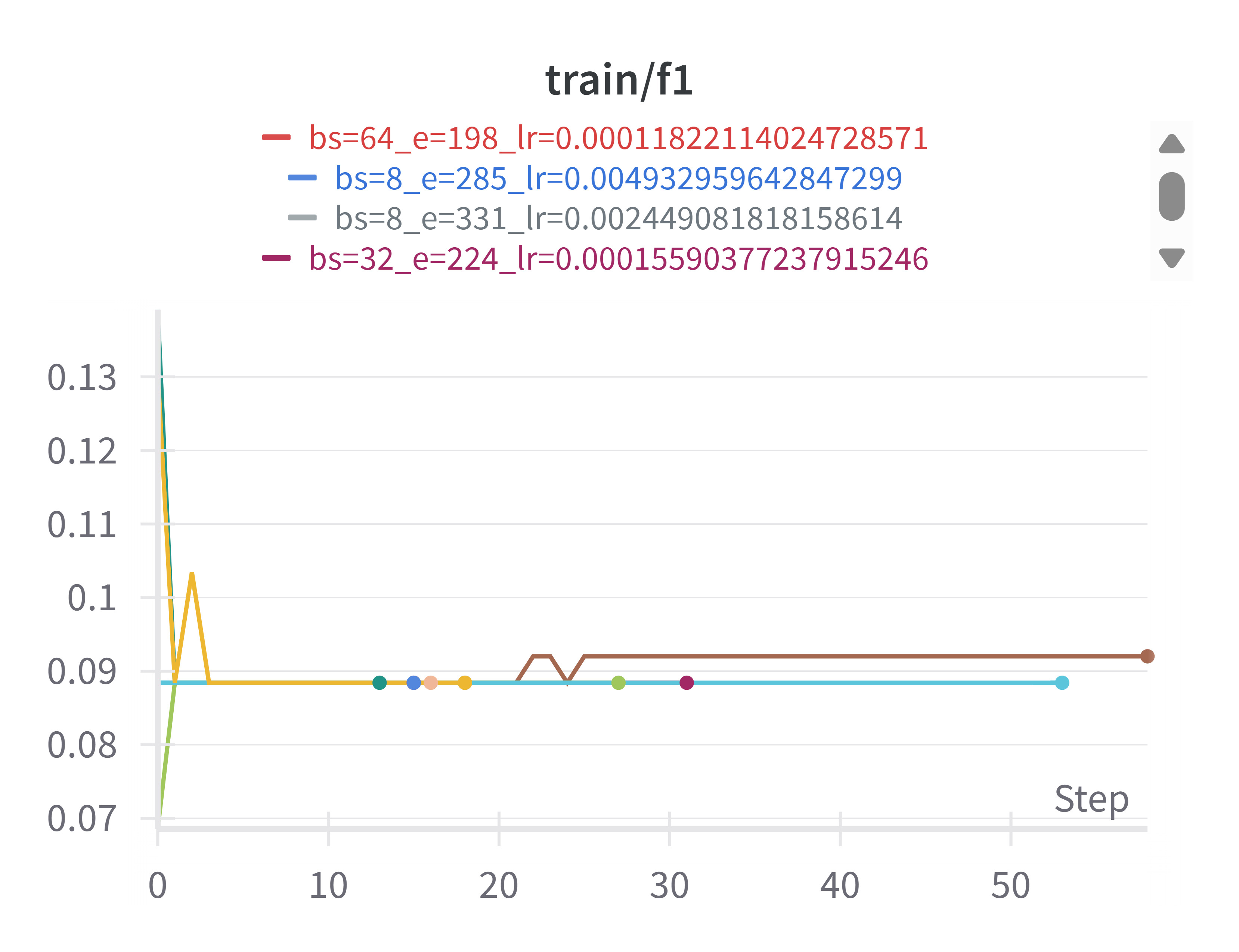}
        \caption{mil\_xlmr\_MaxMLP\_age}
        \label{fig:img3}
    \end{subfigure}
    
    \vspace{0.5cm} 

    \begin{subfigure}[b]{0.3\textwidth}
        \centering
        \includegraphics[width=\textwidth]{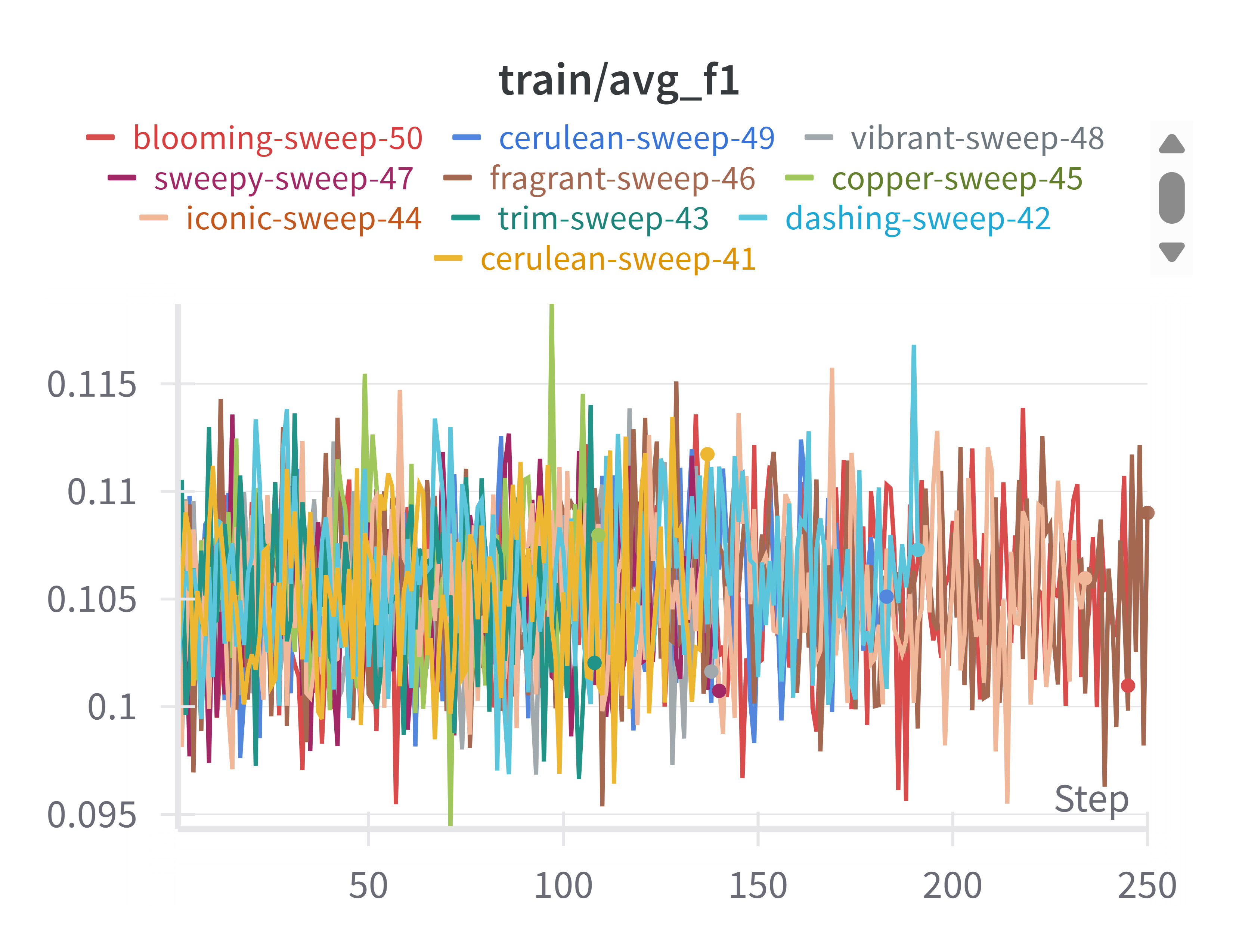}
        \caption{rlmil\_roberta\_MaxMLP\_age}
        \label{fig:img4}
    \end{subfigure}
    \hfill
    \begin{subfigure}[b]{0.3\textwidth}
        \centering
        \includegraphics[width=\textwidth]{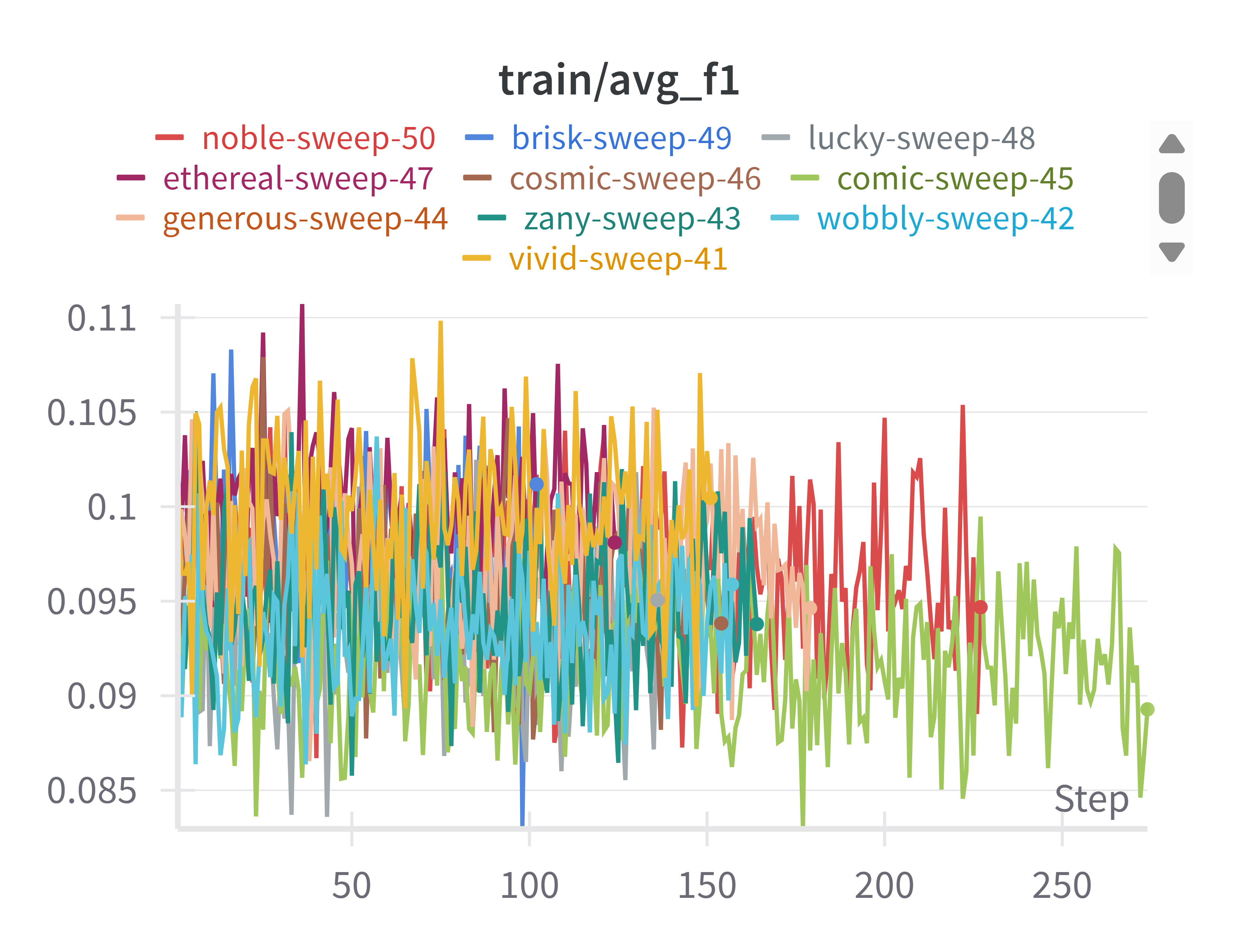}
        \caption{rlmil\_mbert\_MaxMLP\_age}
        \label{fig:img5}
    \end{subfigure}
    \hfill
    \begin{subfigure}[b]{0.3\textwidth}
        \centering
        \includegraphics[width=\textwidth]{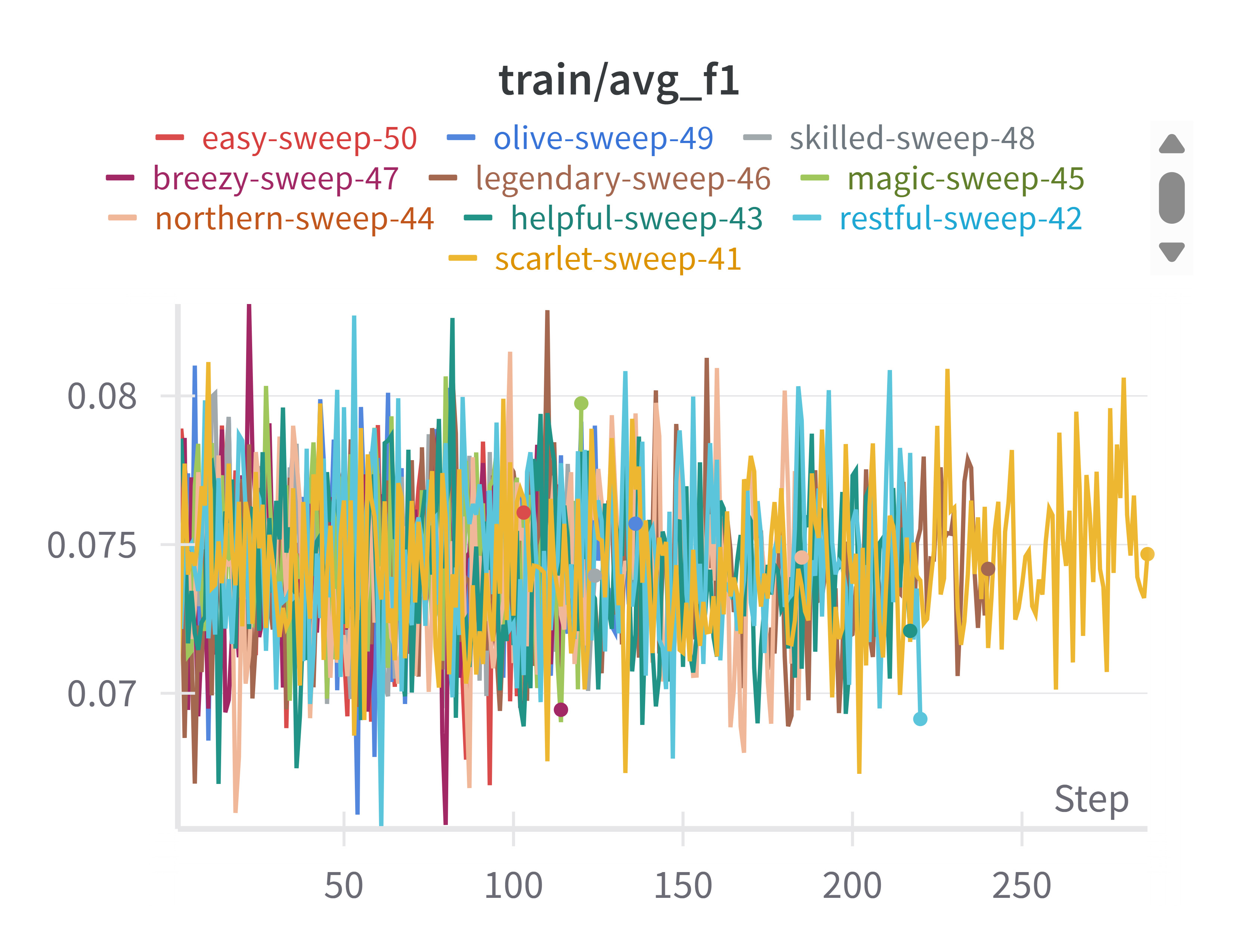}
        \caption{rlmil\_xlmr\_MaxMLP\_age}
        \label{fig:img6}
    \end{subfigure}

    \vspace{0.5cm} 

    \begin{subfigure}[b]{0.3\textwidth}
        \centering
        \includegraphics[width=\textwidth]{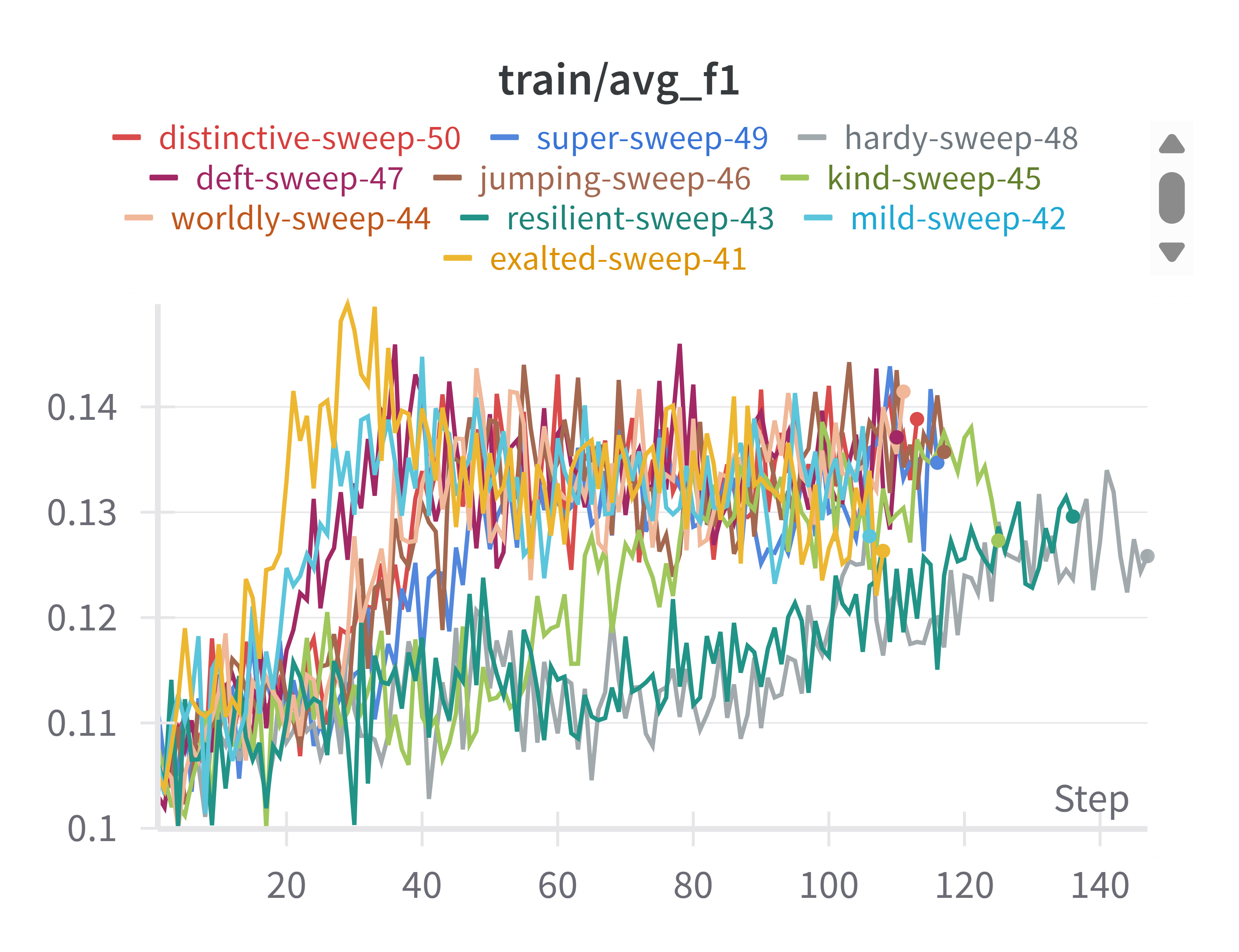}
        \caption{rlmil\_dat\_roberta\_MaxMLP\_age}
        \label{fig:img7}
    \end{subfigure}
    \hfill
    \begin{subfigure}[b]{0.3\textwidth}
        \centering
        \includegraphics[width=\textwidth]{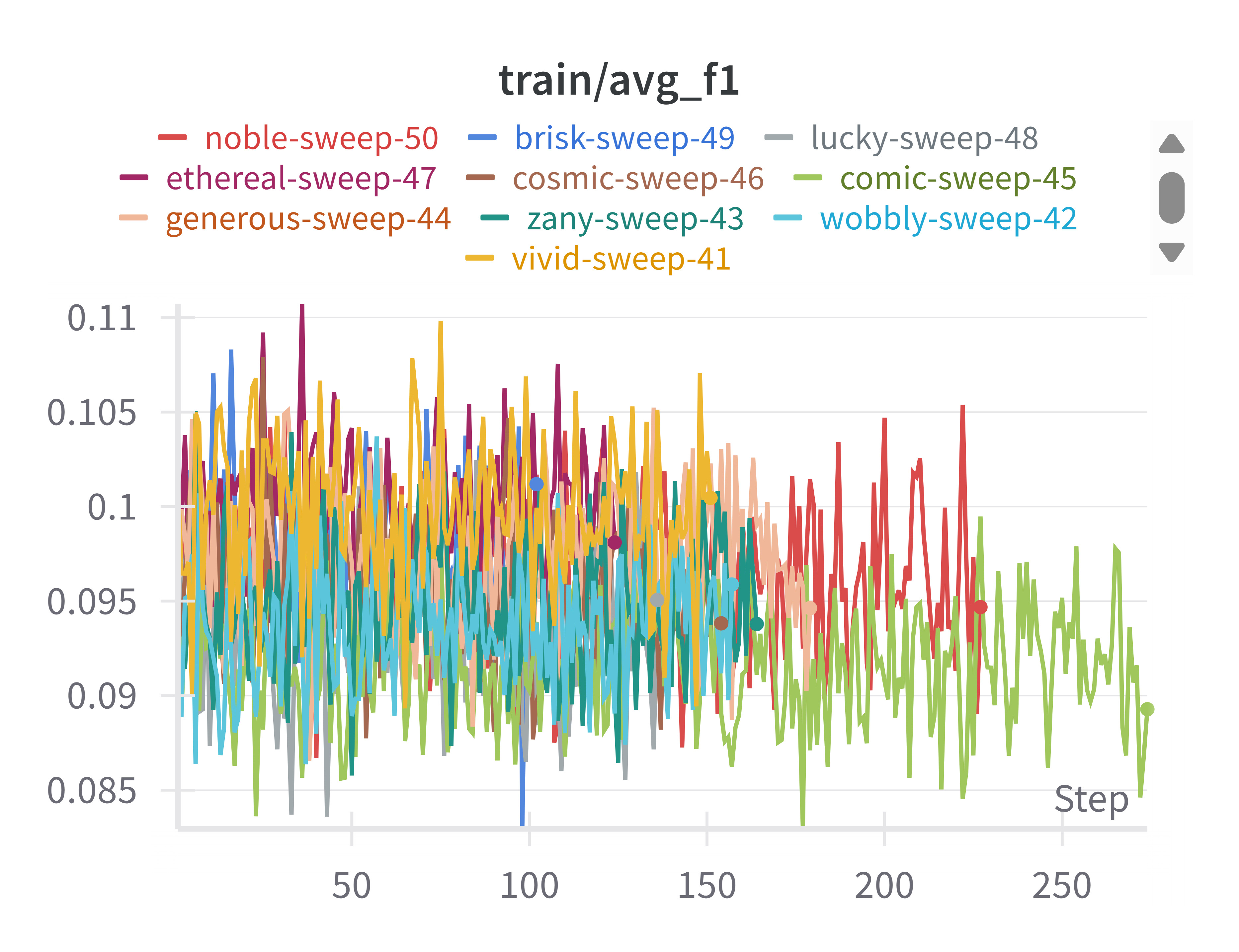}
        \caption{rlmil\_dat\_mbert\_MaxMLP\_age}
        \label{fig:img8}
    \end{subfigure}
    \hfill
    \begin{subfigure}[b]{0.3\textwidth}
        \centering
        \includegraphics[width=\textwidth]{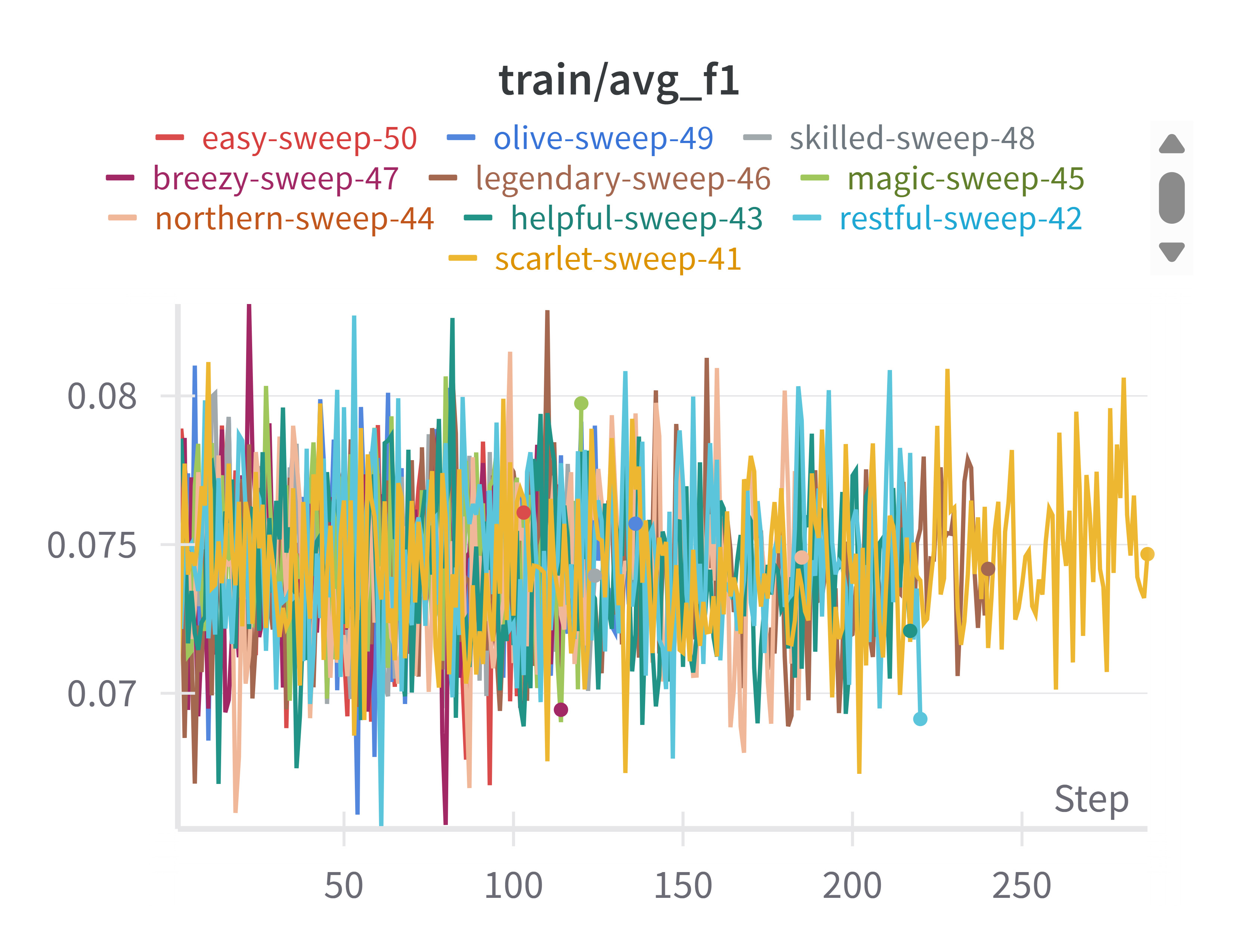}
        \caption{rlmil\_dat\_xlmr\_MaxMLP\_age}
        \label{fig:img9}
    \end{subfigure}

    \caption{Training F1 Score Learning Curves for MaxMLP on Age Prediction (Twitter - Seed 42). This figure displays the training F1 learning curves from Weights \& Biases for various model configurations predicting the age attribute using the MaxMLP pooling head.}
    \label{fig:main_grid}
\end{figure}


\begin{figure}[h!]
    \centering

    \begin{subfigure}[b]{0.3\textwidth}
        \centering
        \includegraphics[width=\textwidth]{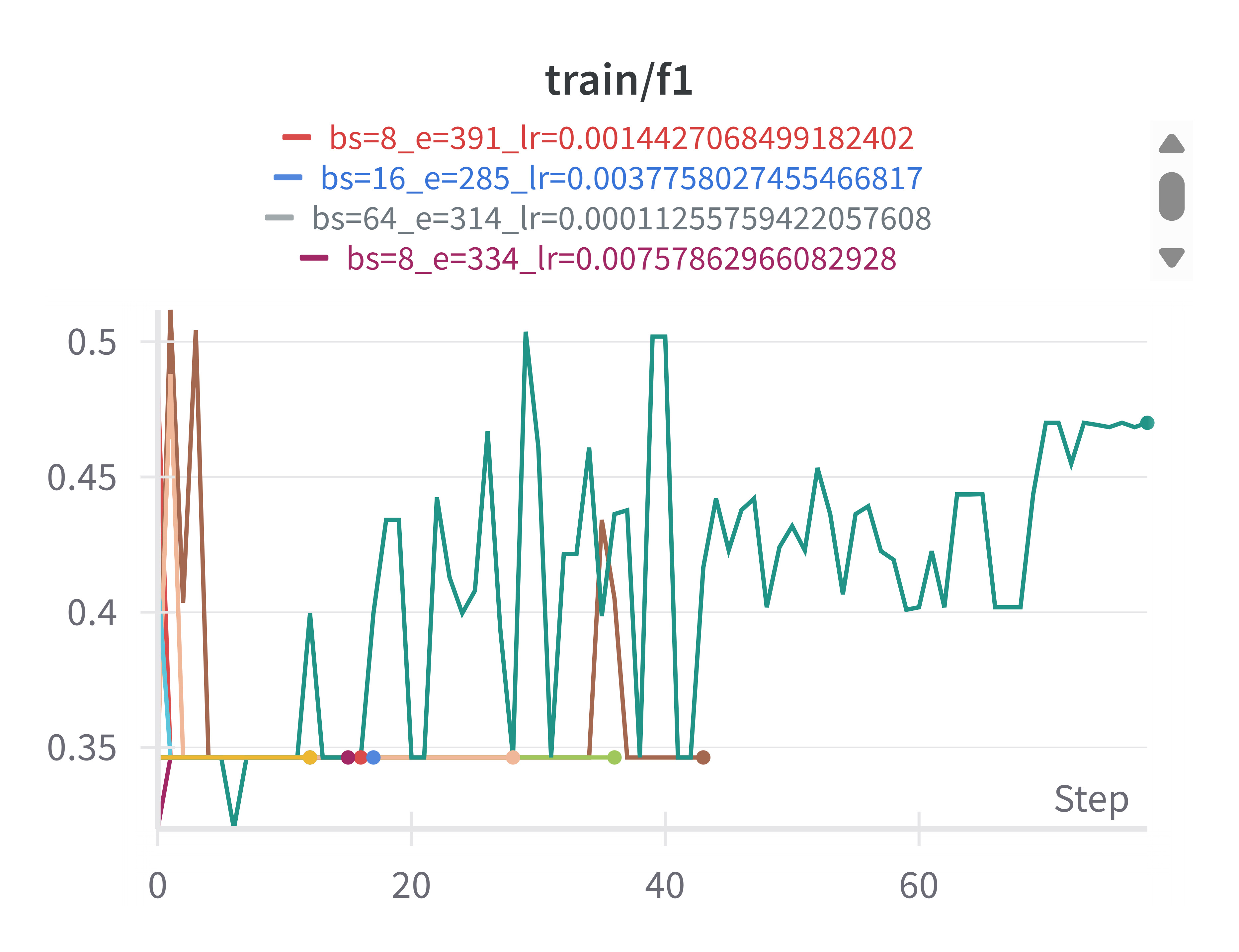}
        \caption{mil\_roberta\_AttentionMLP\_gender}
        \label{fig:img1}
    \end{subfigure}
    \hfill
    \begin{subfigure}[b]{0.3\textwidth}
        \centering
        \includegraphics[width=\textwidth]{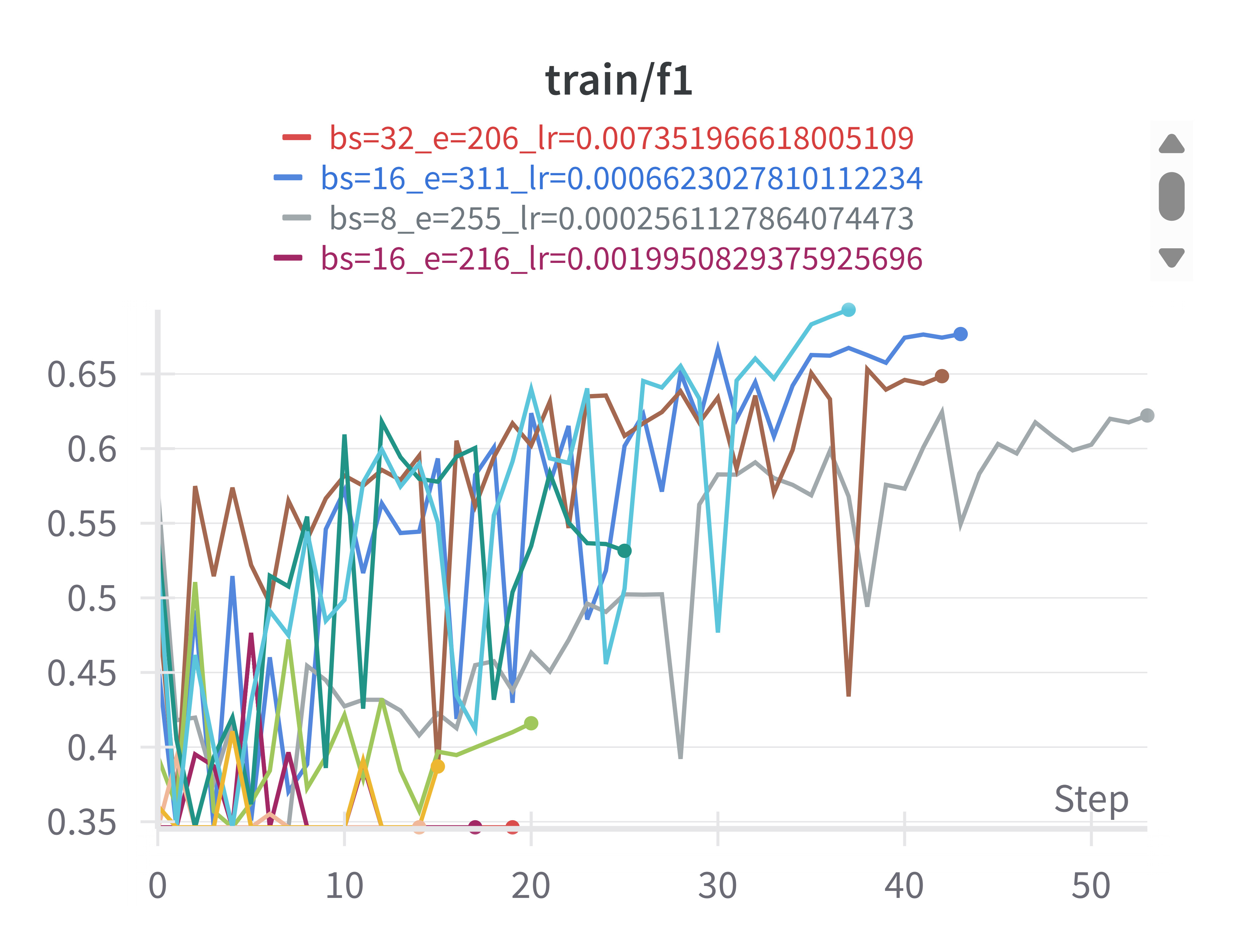}
        \caption{mil\_mbert\_AttentionMLP\_gender}
        \label{fig:img2}
    \end{subfigure}
    \hfill
    \begin{subfigure}[b]{0.3\textwidth}
        \centering
        \includegraphics[width=\textwidth]{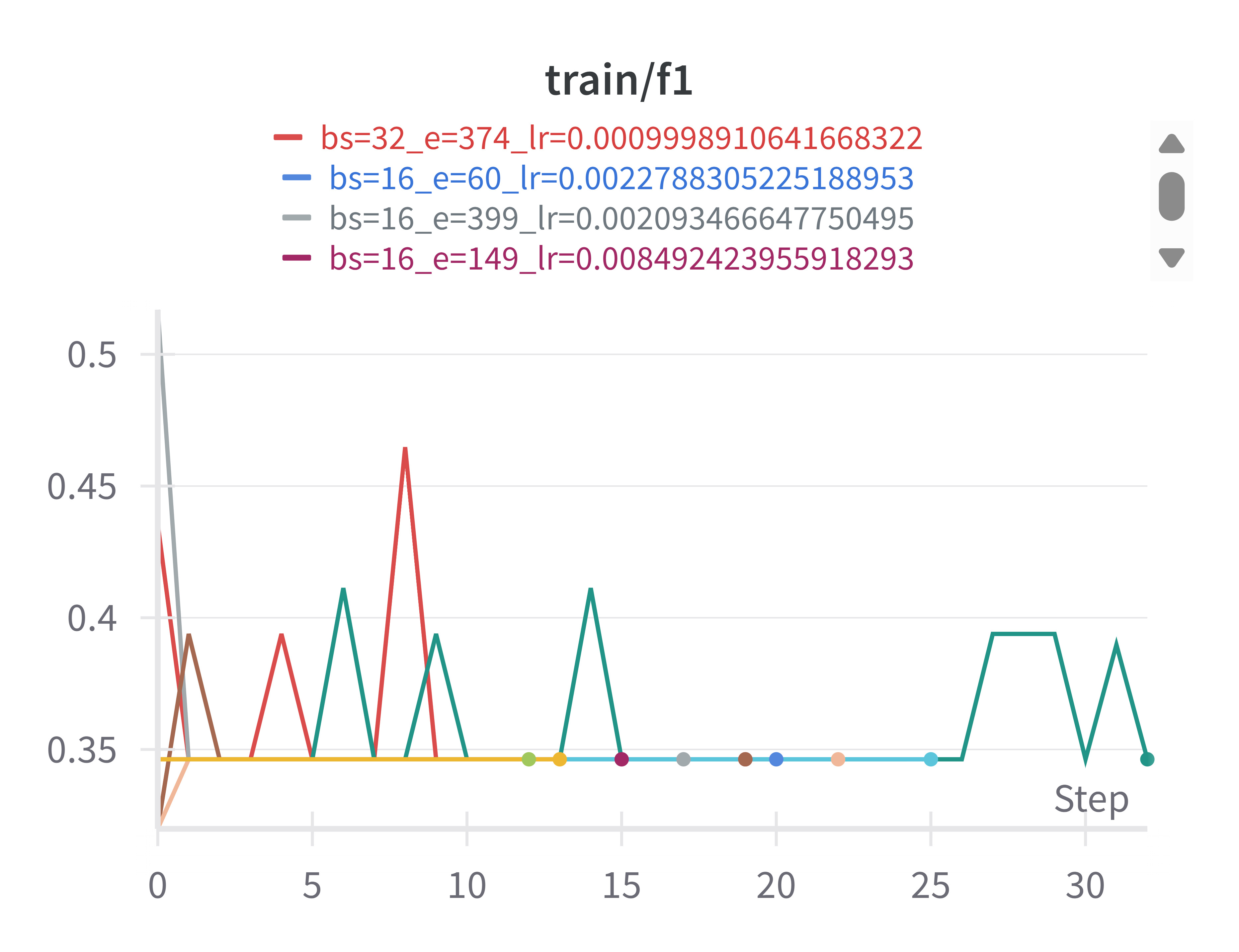}
        \caption{mil\_xlmr\_AttentionMLP\_gender}
        \label{fig:img3}
    \end{subfigure}
    
    \vspace{0.5cm} 

    \begin{subfigure}[b]{0.3\textwidth}
        \centering
        \includegraphics[width=\textwidth]{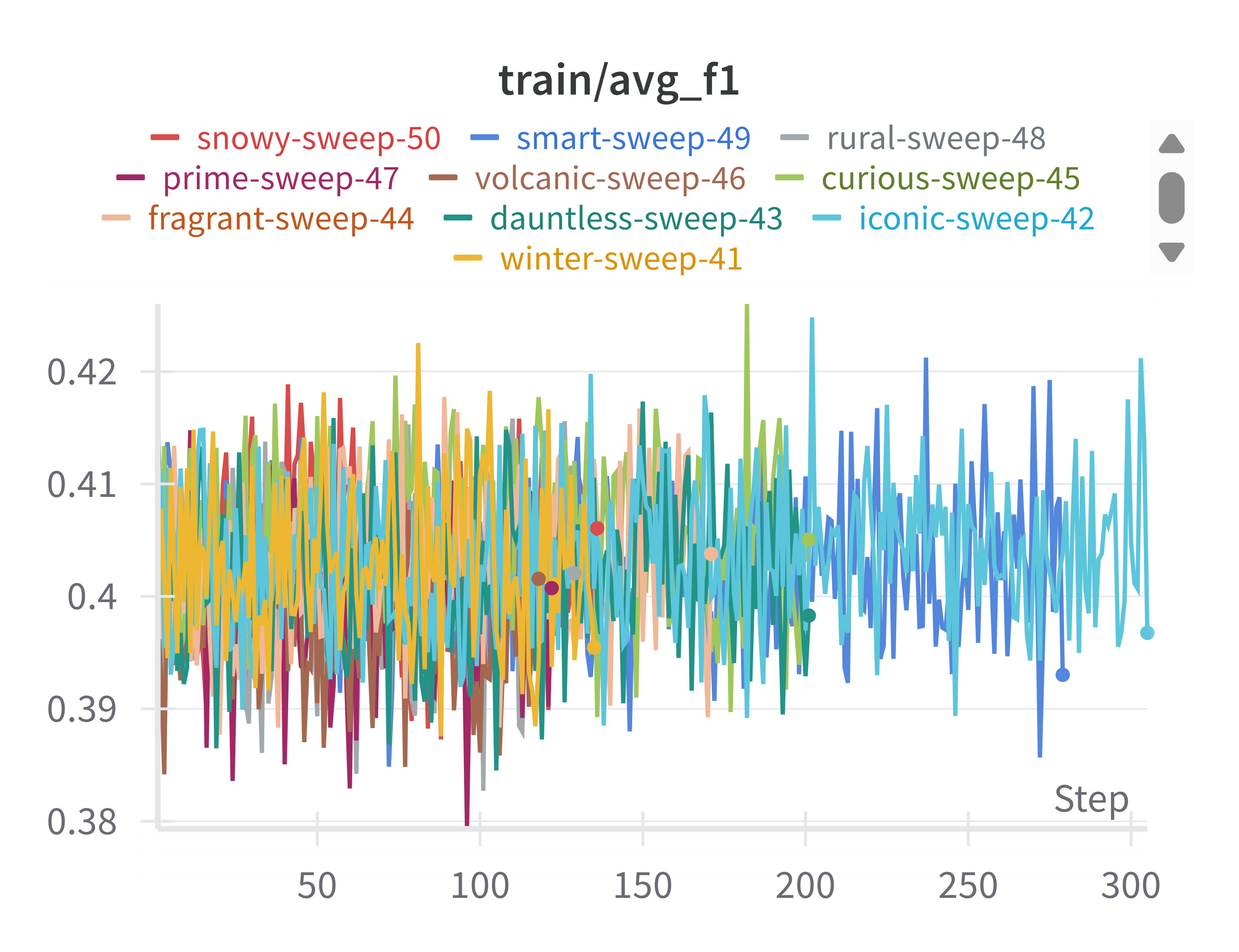}
        \caption{rlmil\_roberta\_AttentionMLP\_gender}
        \label{fig:img4}
    \end{subfigure}
    \hfill
    \begin{subfigure}[b]{0.3\textwidth}
        \centering
        \includegraphics[width=\textwidth]{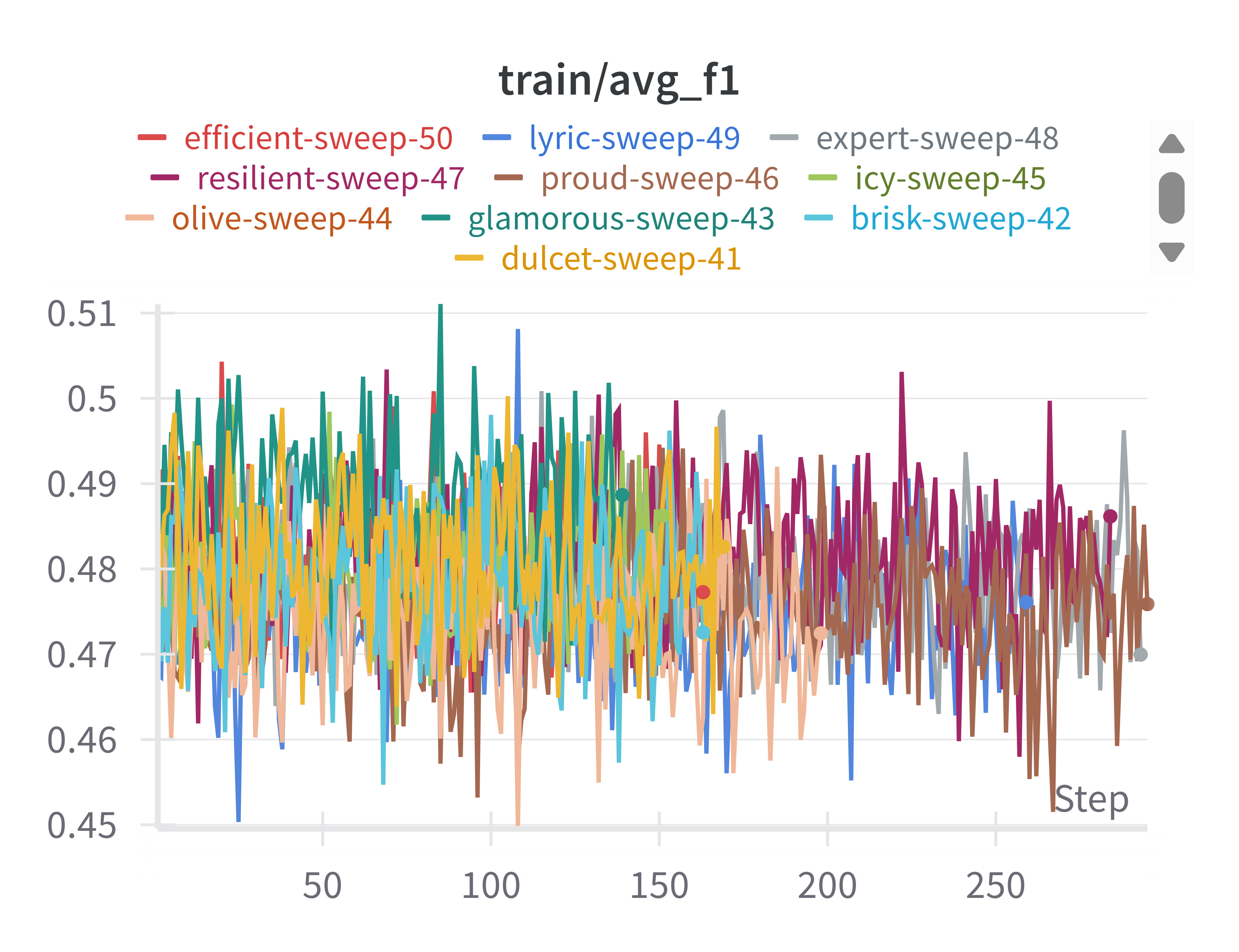}
        \caption{rlmil\_mbert\_AttentionMLP\_gender}
        \label{fig:img5}
    \end{subfigure}
    \hfill
    \begin{subfigure}[b]{0.3\textwidth}
        \centering
        \includegraphics[width=\textwidth]{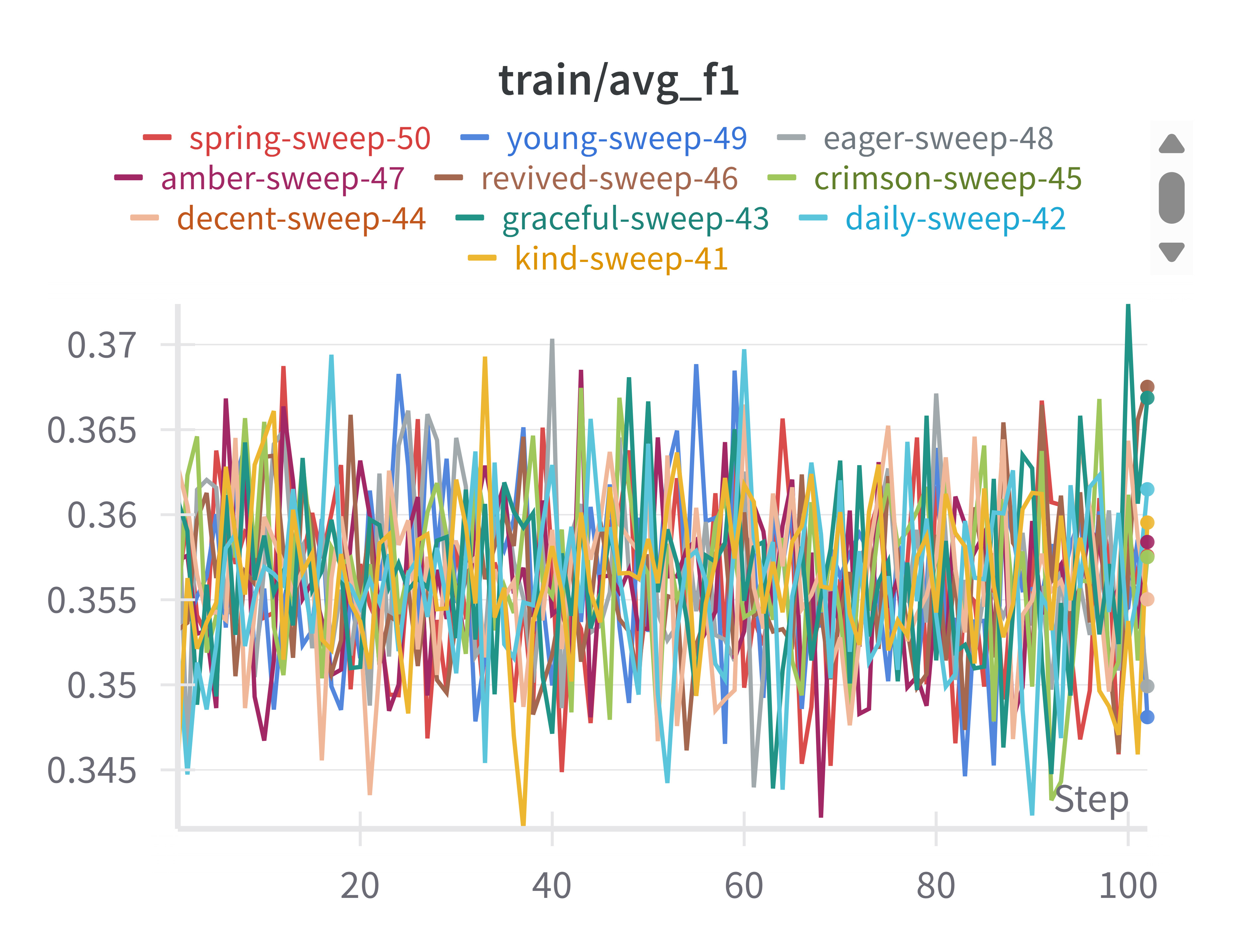}
        \caption{rlmil\_xlmr\_AttentionMLP\_gender}
        \label{fig:img6}
    \end{subfigure}

    \vspace{0.5cm} 

    \begin{subfigure}[b]{0.3\textwidth}
        \centering
        \includegraphics[width=\textwidth]{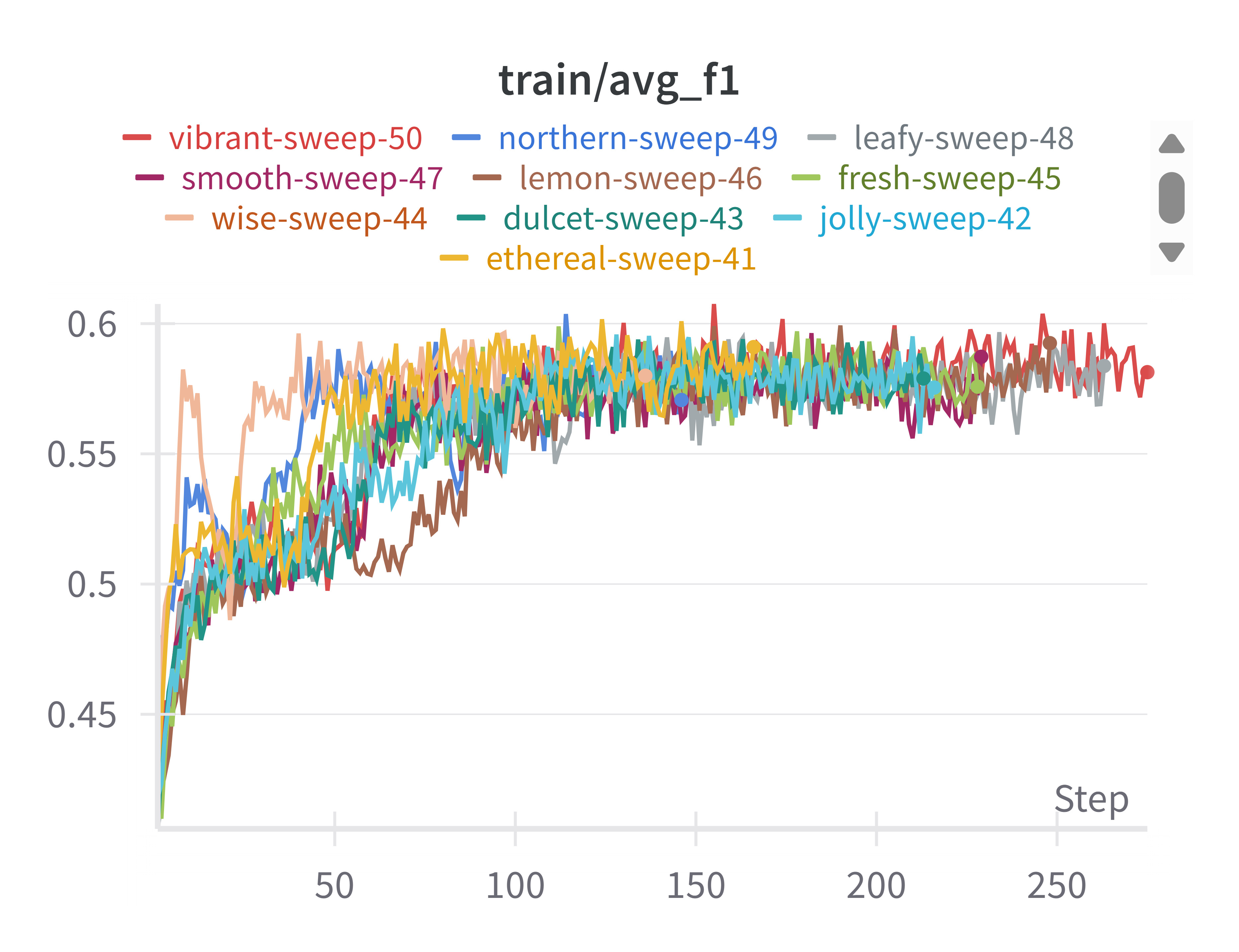}
        \caption{rlmil\_dat\_roberta\_AttentionMLP\_gender}
        \label{fig:rlmil_dat_roberta_AttentionMLP_gender}
    \end{subfigure}
    \hfill
    \begin{subfigure}[b]{0.3\textwidth}
        \centering
        \includegraphics[width=\textwidth]{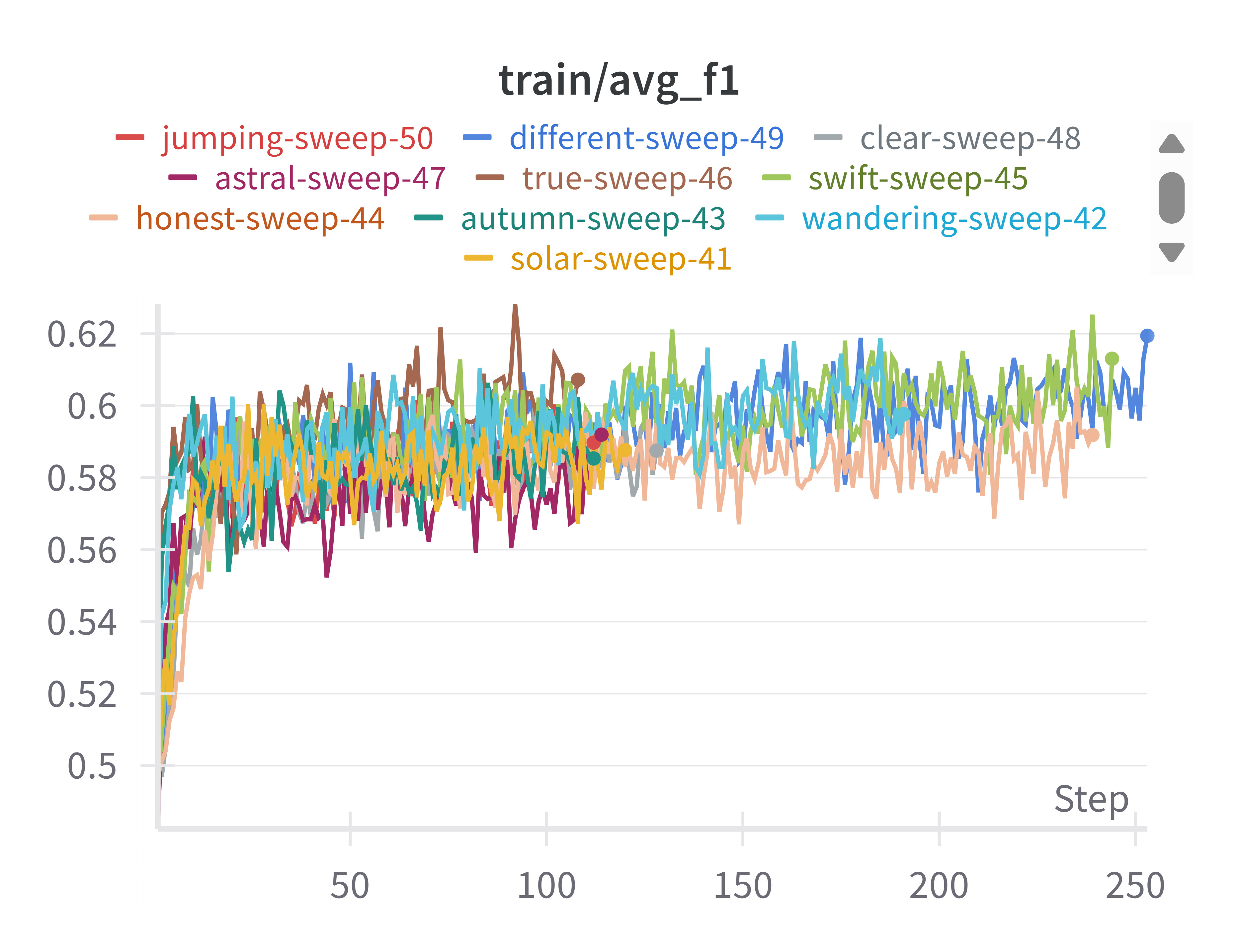}
        \caption{rlmil\_dat\_mbert\_AttentionMLP\_gender}
        \label{fig:rlmil_dat_mbert_AttentionMLP_gender}
    \end{subfigure}
    \hfill
    \begin{subfigure}[b]{0.3\textwidth}
        \centering
        \includegraphics[width=\textwidth]{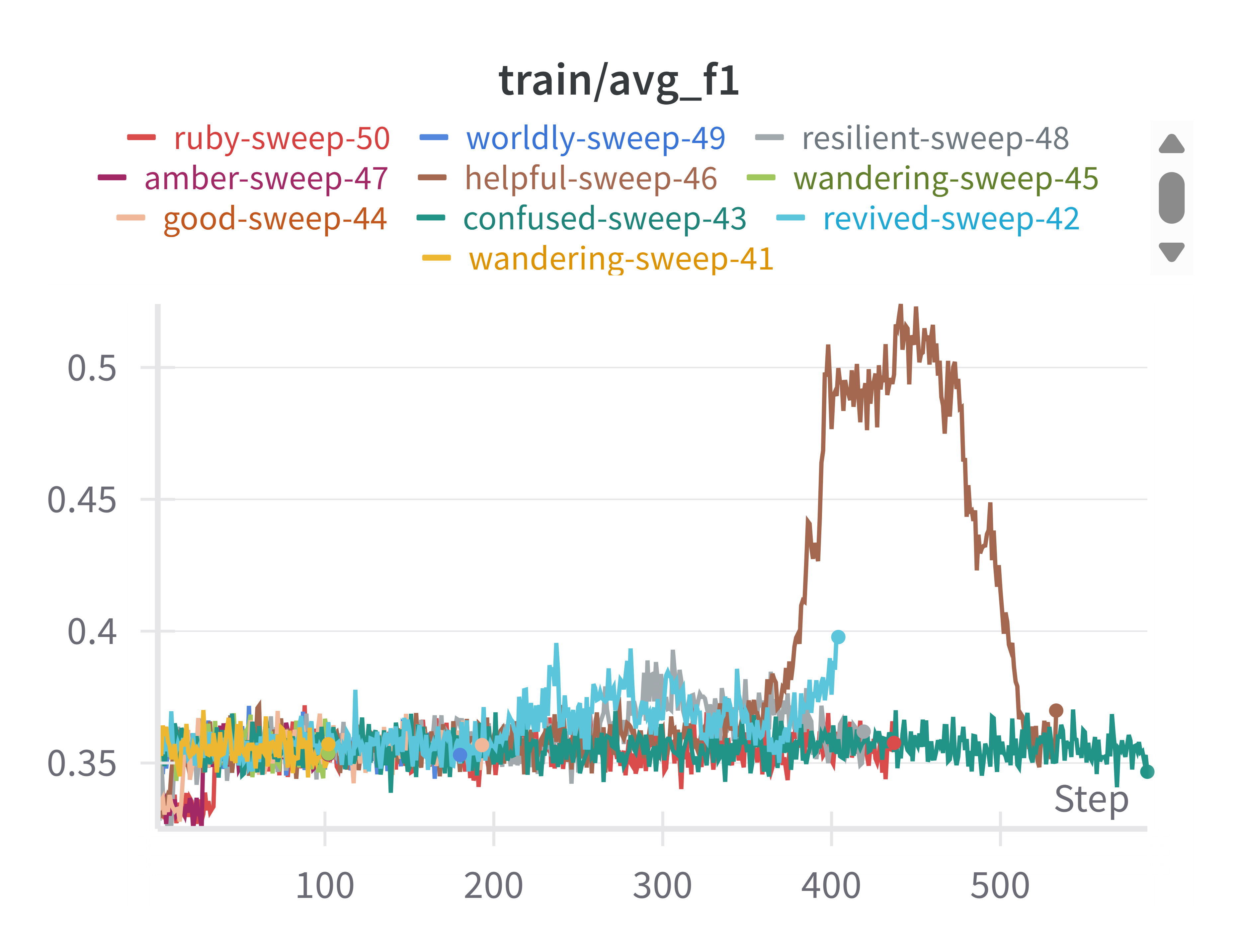}
        \caption{rlmil\_dat\_xlmr\_AttentionMLP\_gender}
        \label{fig:rlmil_dat_xlmr_AttentionMLP_gender}
    \end{subfigure}

    \caption{Training F1 Score Learning Curves for AttentionMLP on Gender Prediction (Twitter - Seed 42). This figure displays the training F1 learning curves from Weights \& Biases for various model configurations predicting the gender attribute using the AttentionMLP pooling head.}
    \label{fig:main_grid}
\end{figure}


\begin{figure}[h!]
    \centering

    \begin{subfigure}[b]{0.3\textwidth}
        \centering
        \includegraphics[width=\textwidth]{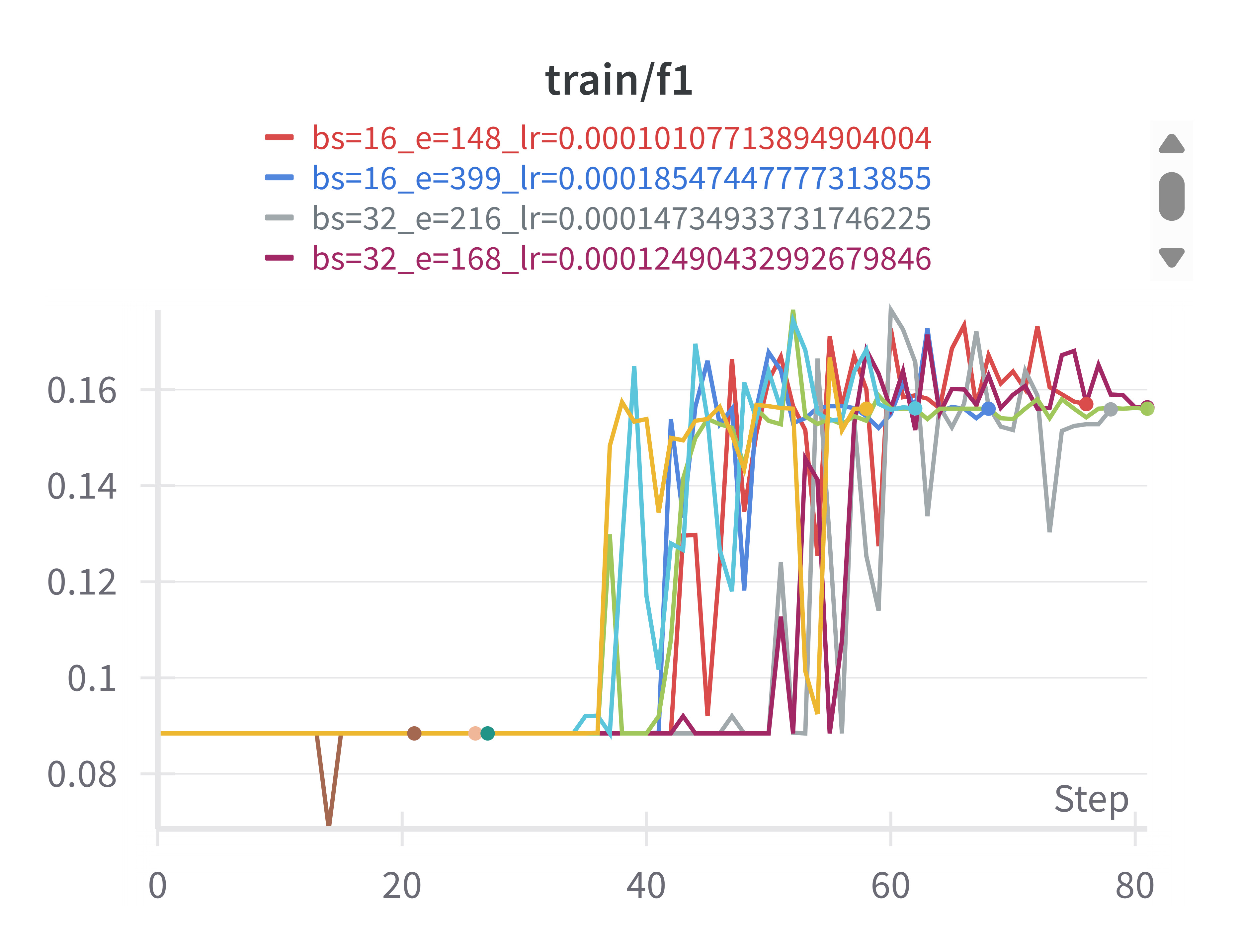}
        \caption{mil\_roberta\_AttentionMLP\_age}
        \label{fig:img1}
    \end{subfigure}
    \hfill
    \begin{subfigure}[b]{0.3\textwidth}
        \centering
        \includegraphics[width=\textwidth]{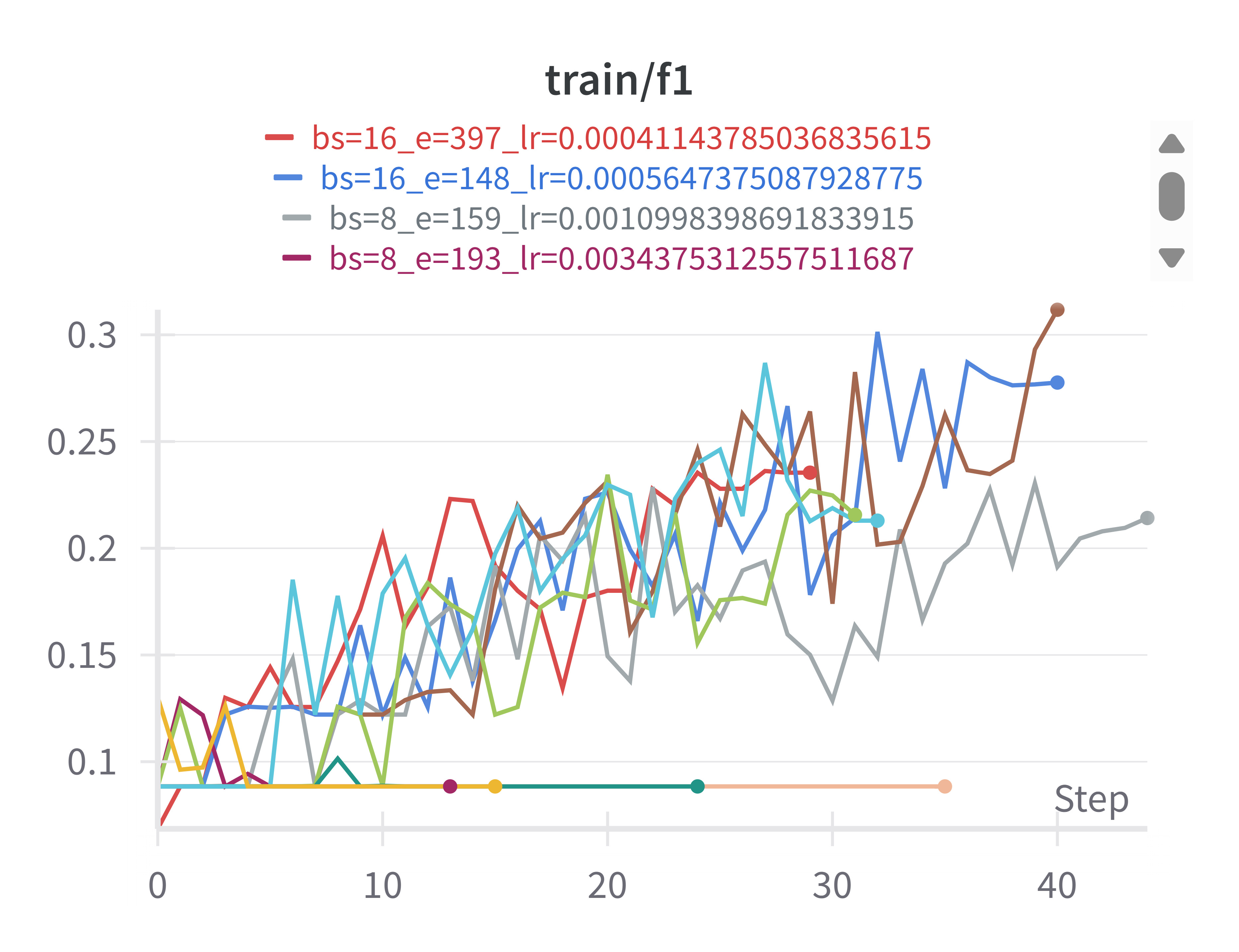}
        \caption{mil\_mbert\_AttentionMLP\_age}
        \label{fig:img2}
    \end{subfigure}
    \hfill
    \begin{subfigure}[b]{0.3\textwidth}
        \centering
        \includegraphics[width=\textwidth]{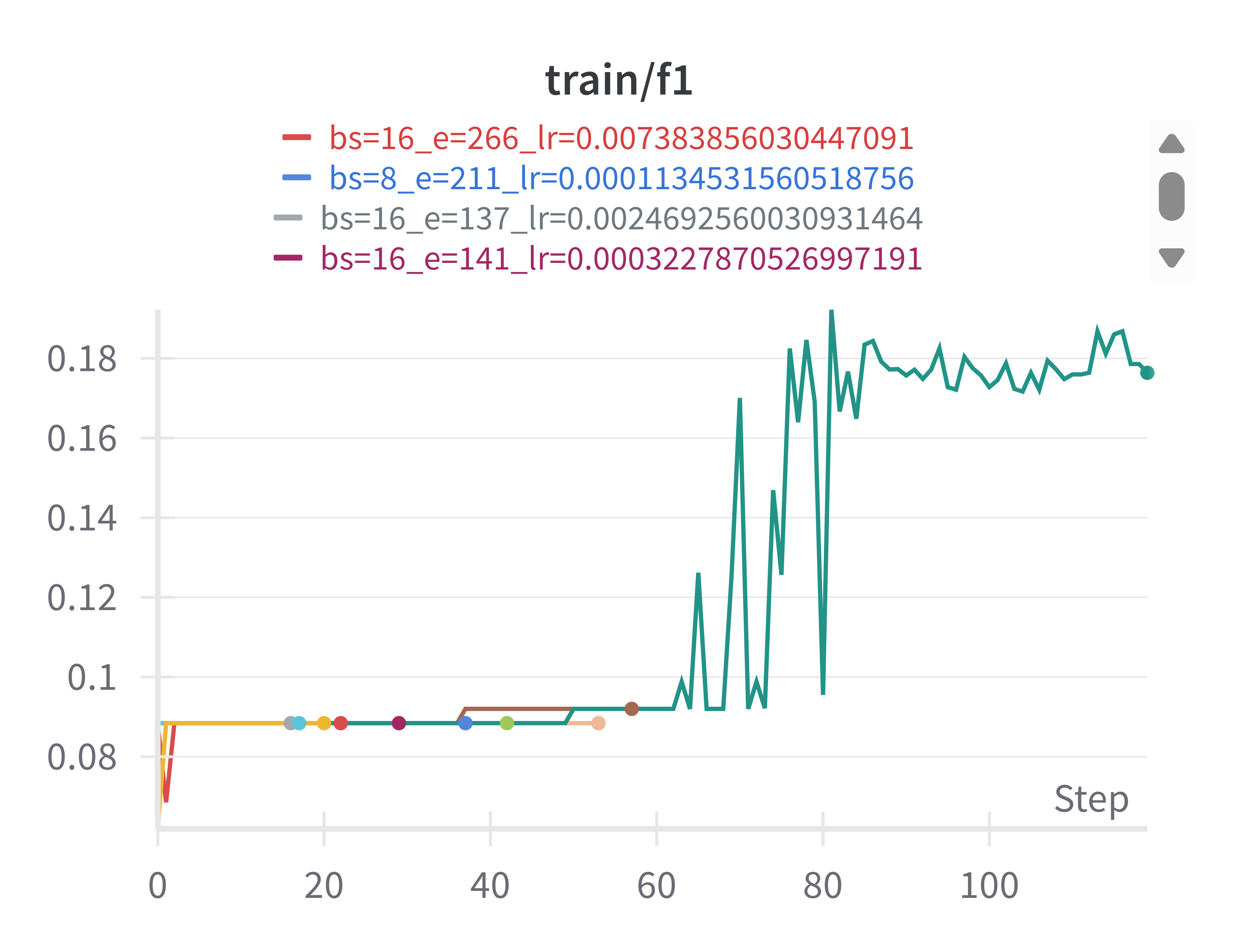}
        \caption{mil\_xlmr\_AttentionMLP\_age}
        \label{fig:img3}
    \end{subfigure}
    
    \vspace{0.5cm} 

    \begin{subfigure}[b]{0.3\textwidth}
        \centering
        \includegraphics[width=\textwidth]{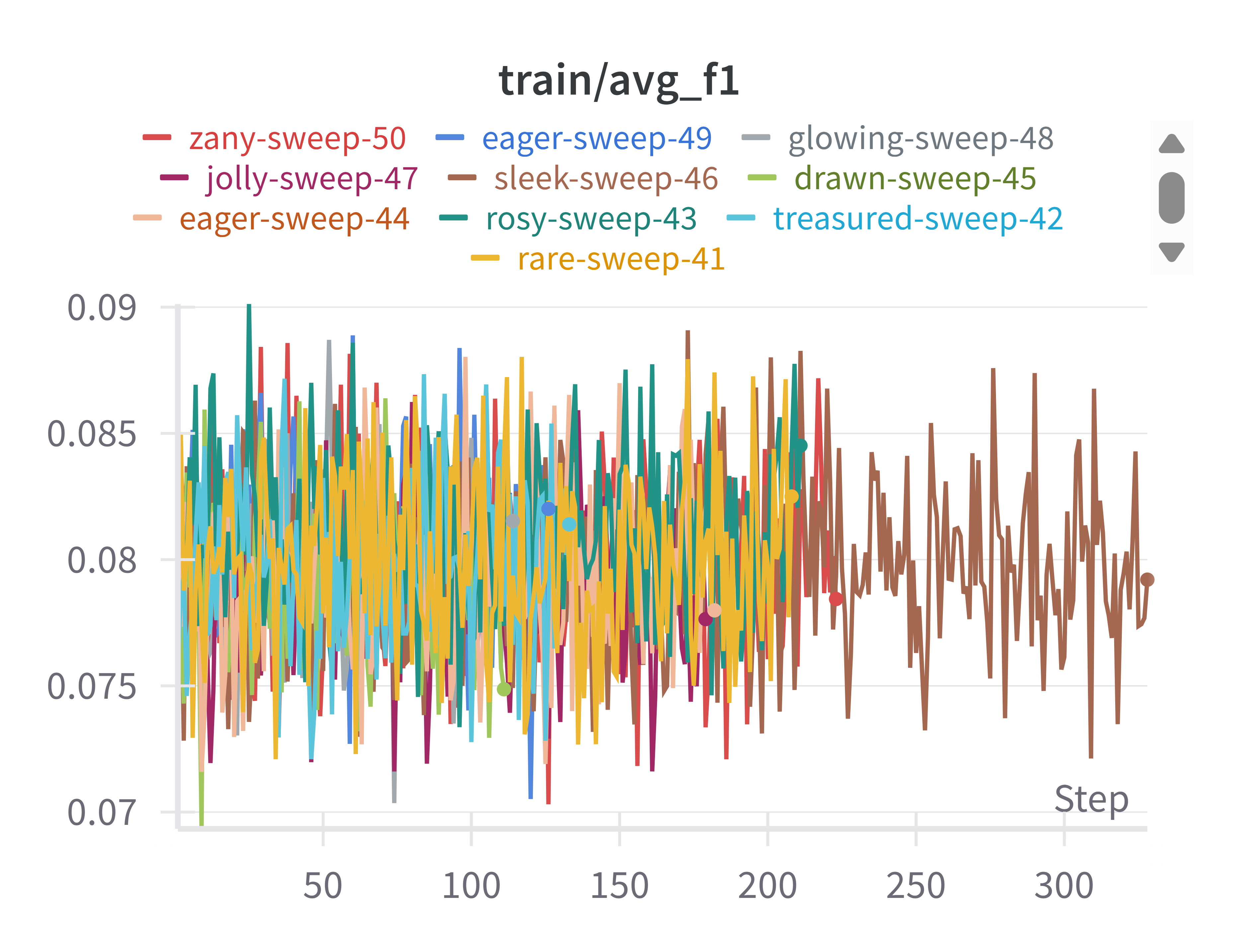}
        \caption{rlmil\_roberta\_AttentionMLP\_age}
        \label{fig:img4}
    \end{subfigure}
    \hfill
    \begin{subfigure}[b]{0.3\textwidth}
        \centering
        \includegraphics[width=\textwidth]{rlmil_roberta_AttentionMLP_age.jpg}
        \caption{rlmil\_mbert\_AttentionMLP\_age}
        \label{fig:img5}
    \end{subfigure}
    \hfill
    \begin{subfigure}[b]{0.3\textwidth}
        \centering
        \includegraphics[width=\textwidth]{rlmil_xlmr_AttentionMLP_gender.jpg}
        \caption{rlmil\_xlmr\_AttentionMLP\_age}
        \label{fig:img6}
    \end{subfigure}

    \vspace{0.5cm} 

    \begin{subfigure}[b]{0.3\textwidth}
        \centering
        \includegraphics[width=\textwidth]{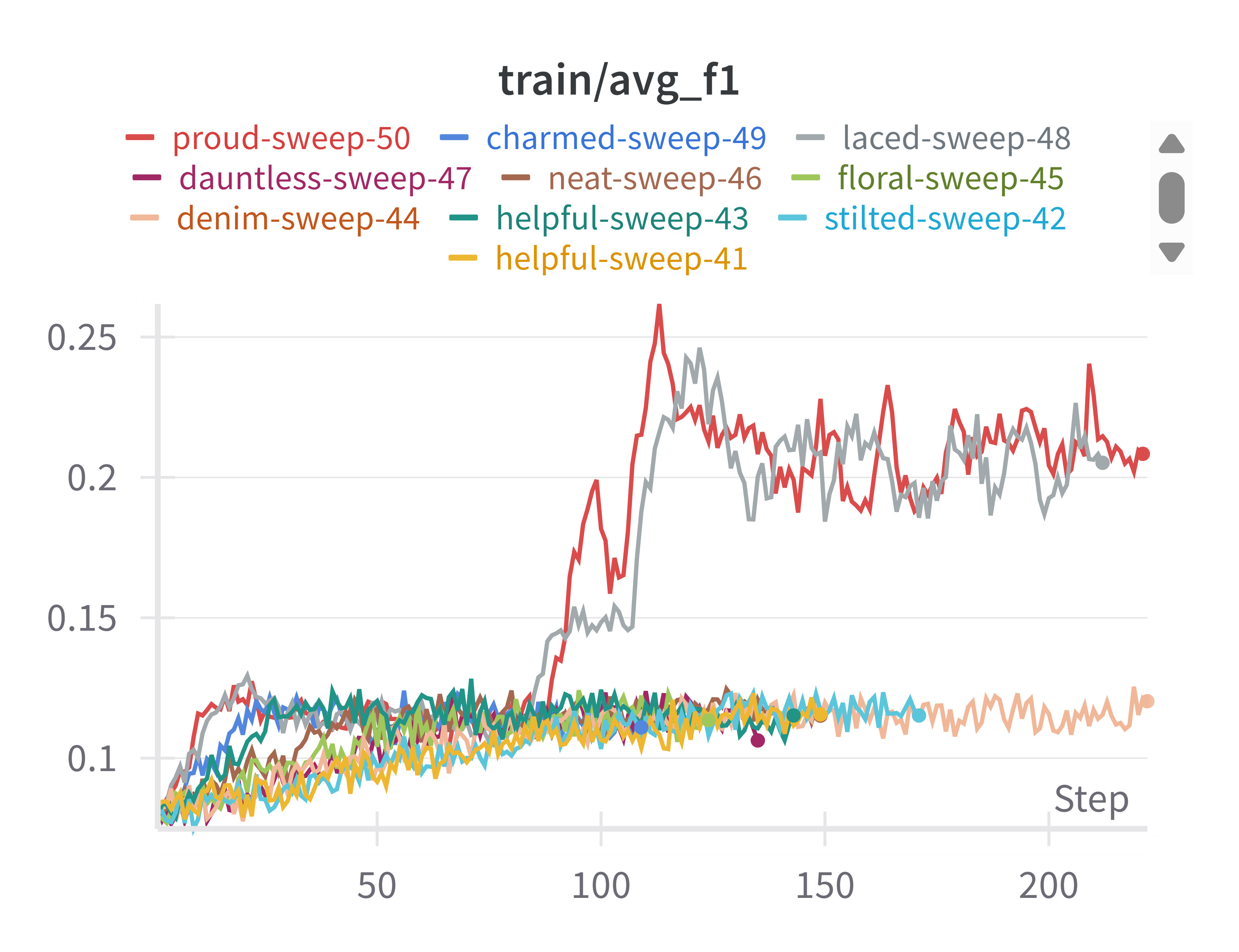}
        \caption{rlmil\_dat\_roberta\_AttentionMLP\_age}
        \label{fig:img7}
    \end{subfigure}
    \hfill
    \begin{subfigure}[b]{0.3\textwidth}
        \centering
        \includegraphics[width=\textwidth]{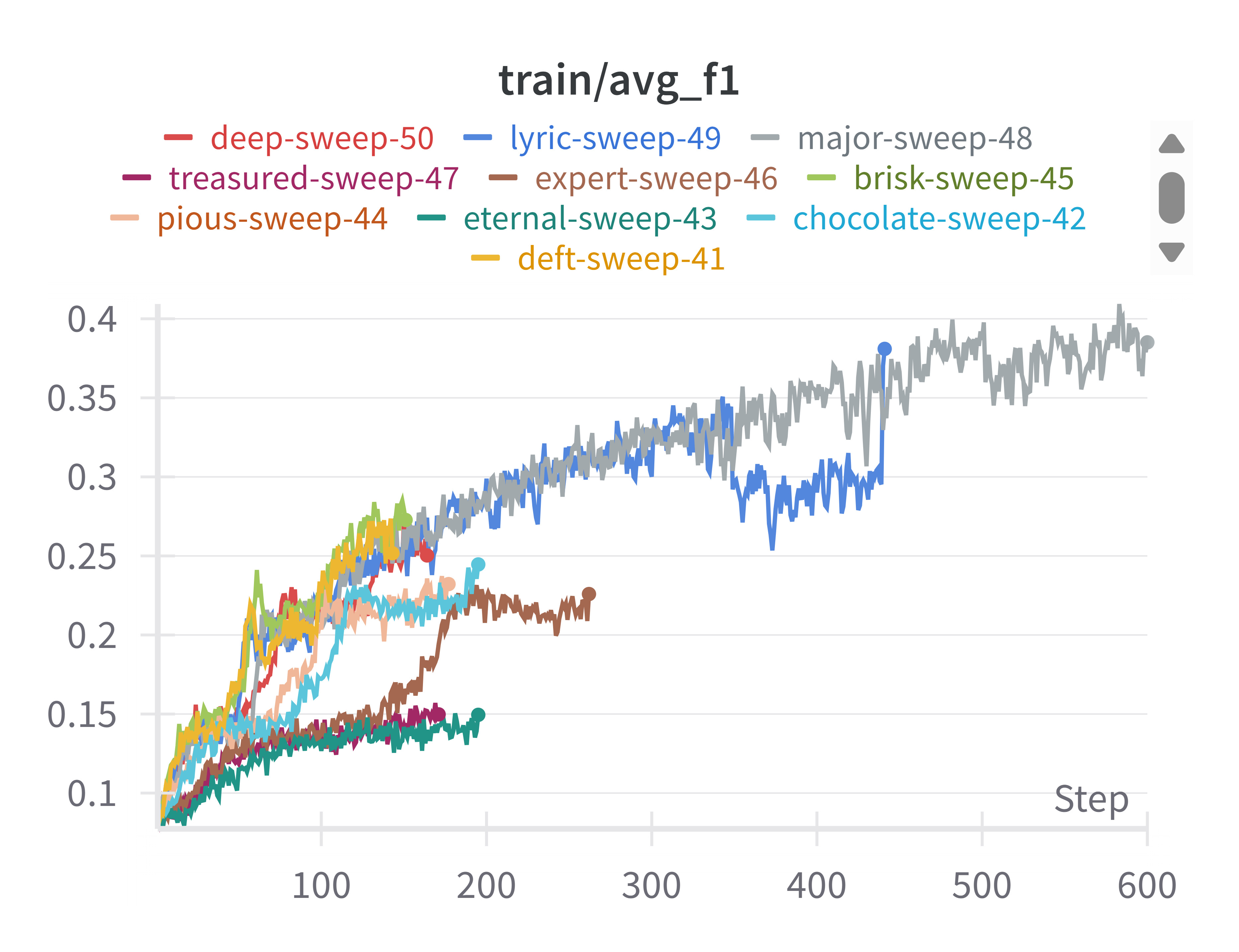}
        \caption{rlmil\_dat\_mbert\_AttentionMLP\_age}
        \label{fig:img8}
    \end{subfigure}
    \hfill
    \begin{subfigure}[b]{0.3\textwidth}
        \centering
        \includegraphics[width=\textwidth]{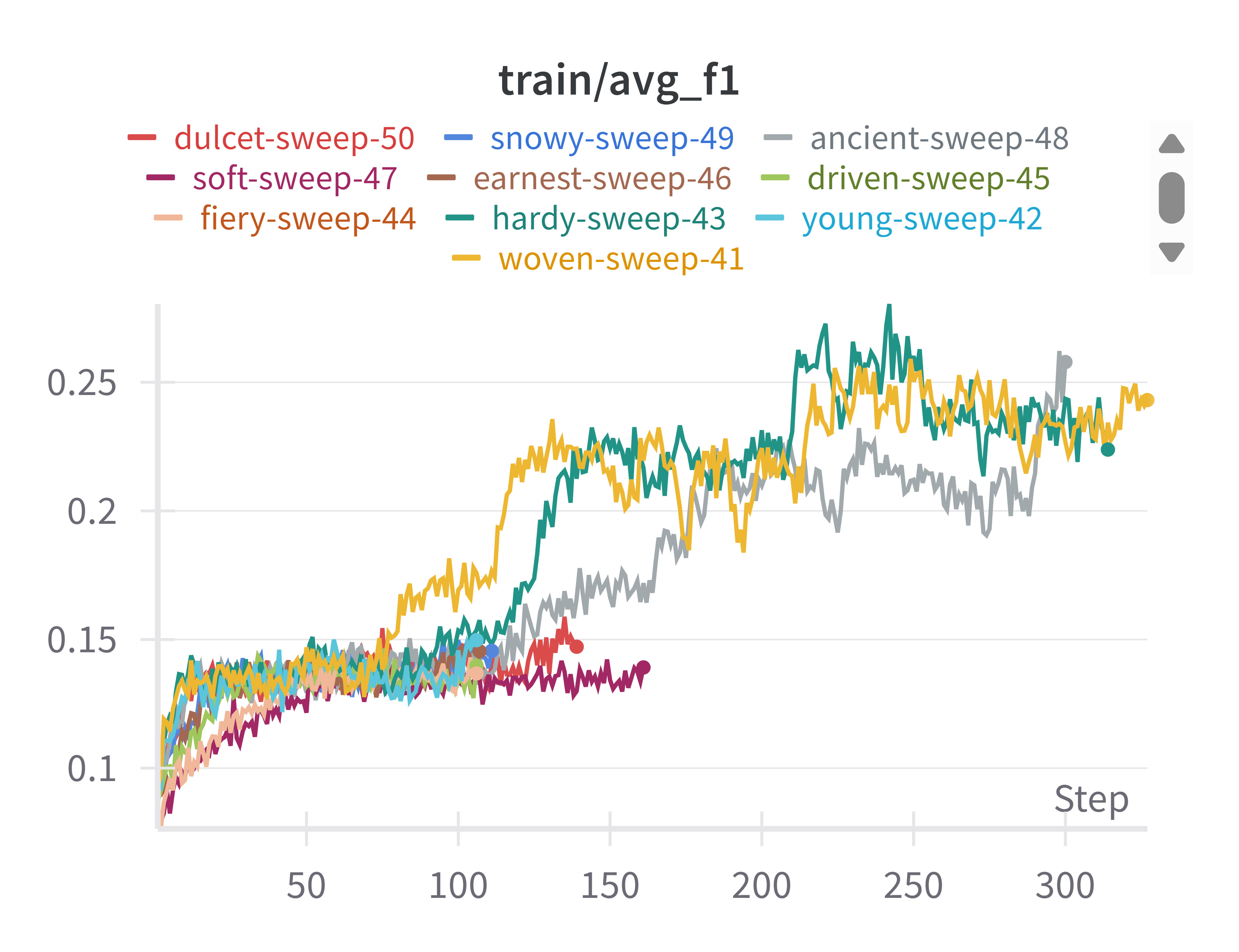}
        \caption{rlmil\_dat\_xlmr\_AttentionMLP\_age}
        \label{fig:img9}
    \end{subfigure}

    \caption{Training F1 Score Learning Curves for AttentionMLP on Age Prediction (Twitter - Seed 42). This figure displays the training F1 learning curves from Weights \& Biases for various model configurations predicting the age attribute using the AttentionMLP pooling head.}
    \label{fig:main_grid}
\end{figure}

\FloatBarrier
\clearpage
\subsection{Supplemental Figures}

\begin{figure}[h!]
    \centering

    \begin{subfigure}[b]{0.49\textwidth}
        \centering
        \includegraphics[width=\textwidth]{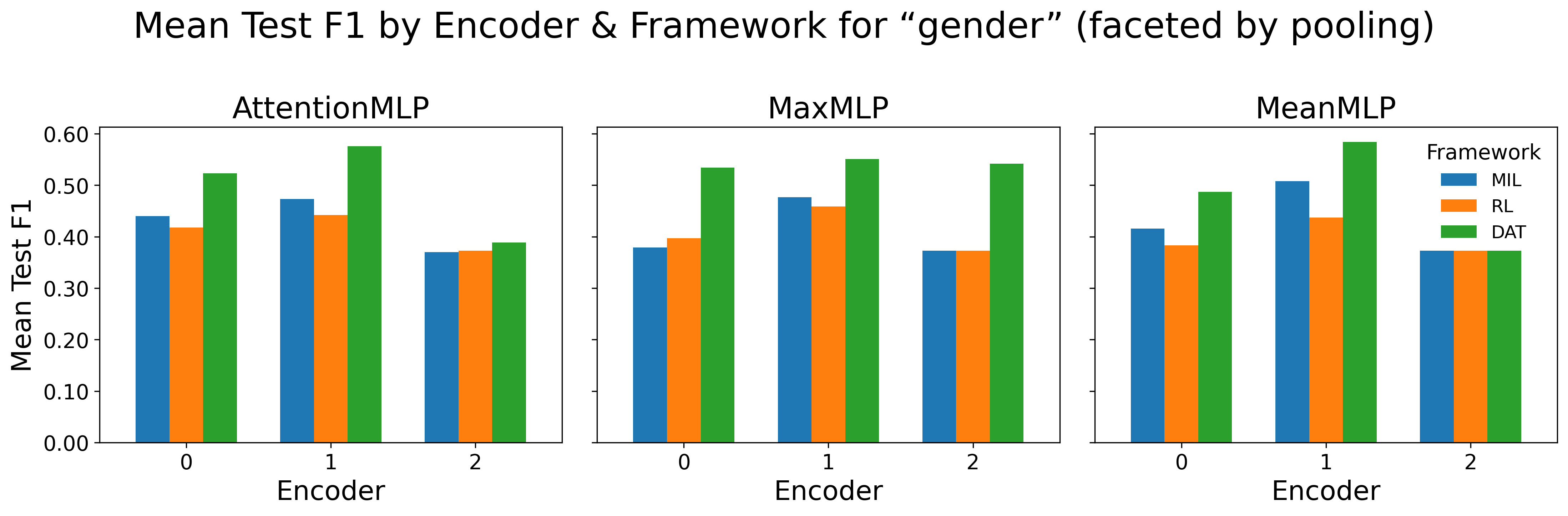}
        \label{fig:bar_gender}
    \end{subfigure}
    \hfill
    \begin{subfigure}[b]{0.49\textwidth}
        \centering
        \includegraphics[width=\textwidth]{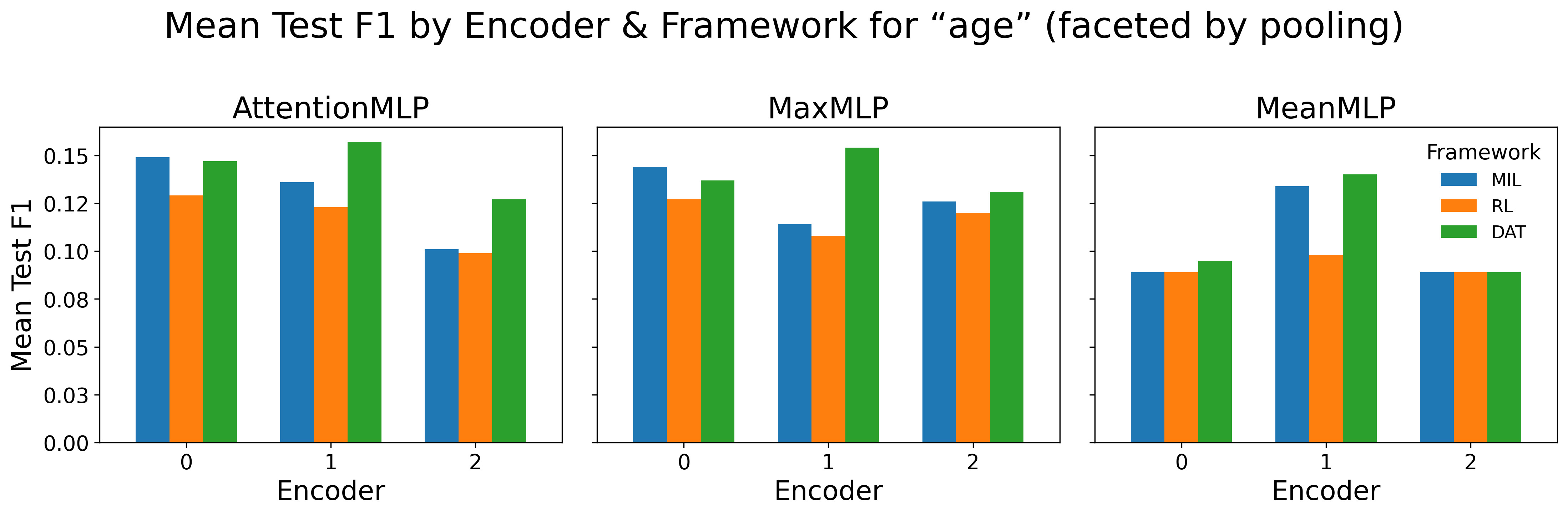}
        \label{fig:bar_age}
    \end{subfigure}
    
    \caption{Test F1 by Encoder \& Model, faceted by pooling (VoxCeleb2).}
    \label{fig:bar_vox}
\end{figure}

\end{appendices}

\end{document}